%% file: main.tex
\pdfoutput=1
\documentclass[letterpaper]{article}
\input{Preamble}
\title{\sysname: Table Explicitization, Neurosymbolically}
  \author{
    Nikita Mehrotra,
    Ayush Kumar,
    Sumit Gulwani,
    Arjun Radhakrishna,
    Ashish Tiwari
}
\affiliations{
    Microsoft, Redmond, WA 98052 \\
    nmehrotra@microsoft.com,
    t-aaykumar@microsoft.com,
    sumitg@microsoft.com, \\
    arradha@microsoft.com,
    astiwar@microsoft.com
}

\begin{document}

\maketitle
\begin{abstract}
We present a neurosymbolic approach, \sysname, for extracting tabular data from semistructured input text. This task is particularly challenging for text input that does not use special delimiters consistently to separate columns and rows. Purely neural approaches perform poorly due to hallucinations and their inability to enforce hard constraints. \sysname uses Structural Decomposition prompting -- a specialized chain-of-thought prompting approach -- on a large language model (LLM) to generate an initial table, and thereafter uses a symbolic checker to evaluate not only the well-formedness of that table, but also detect cases of hallucinations or forgetting. The output of the symbolic checker is processed by a critique-LLM to generate guidance for fixing the table, which is presented to the original LLM in a self-debug loop. Our extensive experiments demonstrate that \sysname significantly outperforms purely neural baselines across multiple datasets and metrics, achieving significantly higher exact match accuracy and substantially reduced hallucination rates. A 21-participant user study further confirms that \sysname's tables are rated significantly more accurate (mean score: 5.0 vs 4.3; p = 0.021), and are consistently preferred for ease of verification and correction, with participants favoring our method in over 60\% of the cases.
\end{abstract}

\input{Sections/Introduction}
\input{Sections/RelatedWork}

\input{Sections/TableExplicitizationProblem}
\input{Sections/Approach}

\input{Sections/Evaluation}

\input{Sections/Results}

\bibliography{main}
\newpage

\appendix
\setcounter{secnumdepth}{2}
\newpage
 \input{appendix/Alogrithim}

\input{appendix/app_evaluation}

\input{appendix/app_results}

\input{appendix/user_study}
 \input{appendix/StructuralDecompositionExample}
\input{appendix/Prompt_Details}
\input{appendix/Qualitative_Analysis_of_Ablation}
\input{appendix/FinReCon60DatasetDetails}

\end{document}

%% file: Preamble.tex
\newif\ifarxiv
\arxivtrue  

\ifarxiv
  \PassOptionsToPackage{frozencache,cachedir=minted-main}{minted}
\else
  \PassOptionsToPackage{finalizecache,cachedir=minted-main}{minted}
\fi
\usepackage{aaai2026}
\usepackage{times}  
\usepackage{helvet}  
\usepackage{courier}  
\usepackage[hyphens]{url}  
\usepackage{graphicx} 
\usepackage{natbib}  
\usepackage{caption} 
\usepackage{latexsym}
\usepackage{microtype}
\usepackage{longtable}
\usepackage{xcolor}
\usepackage{float}
\usepackage{xspace}
\usepackage{adjustbox}
\usepackage{subcaption}
\usepackage{minted}
\usepackage{booktabs}
\usepackage{multirow}
\usepackage{amsmath} 
\usepackage{listings}
\DeclareCaptionStyle{ruled}{labelfont=normalfont,labelsep=colon,strut=off} 
\floatstyle{ruled}
\newfloat{listing}{tb}{lst}{}
\floatname{listing}{Listing}
\usepackage{xcolor}
\usepackage[most]{tcolorbox}
\usepackage{tcolorbox}
\usepackage{algorithm}
\usepackage{algpseudocode}
\tcbuselibrary{breakable,skins}
\tcbuselibrary{listingsutf8}

\newtcolorbox[auto counter, number within=subsection]{promptrefbox}[2][]{%
  colback=green!10!white,
  colframe=green!50!black,
  fonttitle=\bfseries,
  title=Box~\thetcbcounter:~#2,
  label=#1,
  enhanced,
}
\usepackage{enumitem}

\usepackage{fancyvrb}

\definecolor{punct}{RGB}{0,128,128}
\lstdefinelanguage{json}{
    basicstyle=\ttfamily\small,
    showstringspaces=false,
    breaklines=true,
    literate=
     *{0}{{{\color{blue}0}}}{1}
      {1}{{{\color{blue}1}}}{1}
      {2}{{{\color{blue}2}}}{1}
      {3}{{{\color{blue}3}}}{1}
      {4}{{{\color{blue}4}}}{1}
      {5}{{{\color{blue}5}}}{1}
      {6}{{{\color{blue}6}}}{1}
      {7}{{{\color{blue}7}}}{1}
      {8}{{{\color{blue}8}}}{1}
      {9}{{{\color{blue}9}}}{1}
      {:}{{{\color{punct}{:}}}}{1}
      {,}{{{\color{punct}{,}}}}{1}
      {"}{{{\color{red}{"}}}}{1},
}
\newcommand{\sysname}{{\fontfamily{ptm}\selectfont\textsc{Ten}}\xspace}

\newcommand\ignore[1]{{}}
\newcommand\userstudysys{{\fontfamily{ptm}\selectfont\textsc{Ten}}\xspace}

%% file: Sections/Introduction.tex
\section{Introduction}
\label{sec:intro}

\par LLMs have significantly advanced the state-of-the-art on numerous Natural Language Processing (NLP) tasks, particularly in text generation \cite{samuel2024bertsgenerativeincontextlearners, DBLP:journals/corr/abs-2005-14165}. Recent works leveraging LLMs have shown promising capabilities in transforming natural language into structured formats (\textit{text-to-table}) \cite{text2table, wu-etal-2022-text-table, deng2024texttupletable, mukul}, generating descriptive narratives from structured data (\textit{table-to-text}) \cite{andrejczuk2022tabletotextgenerationpretrainingtabt5, parikh-etal-2020-totto,wiseman-etal-2017-challenges}, and performing question-answering tasks on tabular datasets \cite{herzig-etal-2020-tapas, yin-etal-2020-tabert}. 

\par Despite these advances, extracting structured data directly from unstructured documents such as PDFs, emails, and reports remains a critical challenge. Unlike \textit{text-to-table} or \textit{table-to-text} tasks, this problem is complicated by the complexity of the input data, including ambiguous delimiters, unpredictable formatting, merged cells, and inconsistent line breaks. Prior work has often focused on extracting tables from clean textual descriptions, such as medical discharge summaries or news reports, where structured relationships are clear. However, our setting involves flattened text, often produced by copy-paste operations or OCR, where the original table structure is lost. While layout-based tools like Tesseract \cite{tesseract} and Tabula \cite{tabula} leverage visual structure and coordinate metadata from PDFs, we assume no such layout information. These methods perform poorly when visual structure is missing or distorted, such as in copy-pasted text or noisy OCR outputs. In contrast, \sysname, operates solely on linearized text and focuses on reconstructing structured tables from noisy, token-only inputs making \sysname a complementary component to OCR pipelines.

\par Our motivation comes from the challenges faced by users when copy-pasting data. Research indicates that manual data entry error rates range from 1\% to 5\%, with some complex document processing tasks reporting extraction errors as high as 50\% \cite{BARCHARD20111834}. In clinical data management, transcription errors vary from 2.3\% to 26.9\% \cite{Mathes2017DataExtractionErrors}, and database errors can span 2 to 2,784 per 10,000 fields, depending on the processing method \cite{Garza2025ErrorRates}. These error rates highlight a significant productivity bottleneck: knowledge workers devote substantial time to manually transfer tabular data from PDFs, reports, and other unstructured documents into spreadsheets, introducing human errors and incurring high time costs. The fragility of copy-paste operations exacerbates the issue, as formatting is often lost and data integrity compromised. Furthermore, extraction errors, including AI-induced hallucinations, can lead to significant financial losses, regulatory violations, and flawed decision-making, underscoring the need for robust solutions~\cite{Crisanto2024RegulatingAI,10.1093/polsoc/puaf001}.

\par Existing methods for table extraction fall into two categories: symbolic approaches, which rely on rule-based techniques, and neural approaches, which use deep learning to infer table structures. While symbolic methods offer interpretability and determinism, they rely on predefined patterns and struggle with irregular formatting. Conversely, neural models, particularly LLMs, excel at learning diverse table layouts, but lack explicit structural constraints. A major limitation of neural approaches is hallucination, where models generate plausible but incorrect table structures, fabricate data points, or misalign extracted information.
\begin{figure}[t] 
  \centering
  \captionsetup{skip=1pt}        
  \ignore{
  \begin{subfigure}{\linewidth}
    \centering
\includegraphics[width=\linewidth]{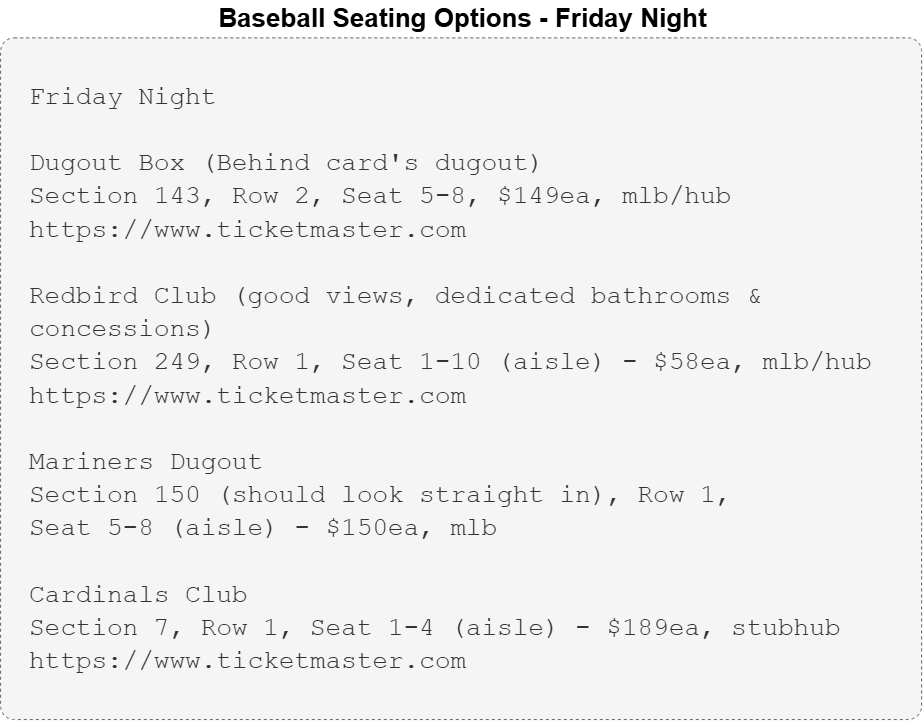}
    \caption*{\small Ground‑truth seating description}
    \end{subfigure}
    \endignore}
    \begin{subfigure}{\linewidth}
\begin{Verbatim}[fontsize=\scriptsize,frame=single]
Friday Night
   
Dugout Box (Behind card's dugout)
Section 143, Row 2, Seat 5-8, $149ea, mlb/hub
https://www.ticketmaster.com

Redbird Club (good views, dedicated bathrooms &
concessions)
Section 249, Row 1, Seat 1-10 (aisle) - $58ea, mlb/hub
https://www.ticketmaster.com

Mariners Dugout
Section 150 (should look straight in), Row 1, 
Seat 5-8 (aisle) - $150ea, mlb

Cardinals Club
Section 7, Row 1, Seat 1-4 (aisle) - $189ea, stubhub
https://www.ticketmaster.com
\end{Verbatim}
  \end{subfigure}
 {\small{Input text containing seating description}}
 
  \vspace{1em} 
  \begin{subfigure}{\linewidth}
    \centering
    \includegraphics[width=\linewidth,
                     trim=0 1 0 1,clip]
                     {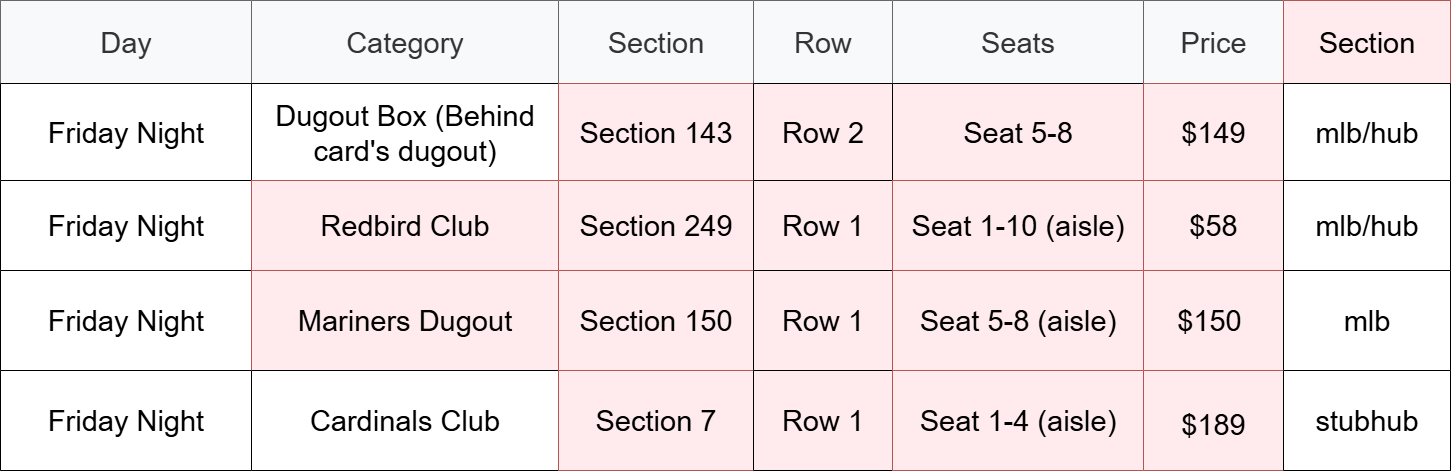}
  \end{subfigure}
{\small{Table predicted by LLM Run \#1}}

  \vspace{1em}
  \begin{subfigure}{\linewidth}
    \centering
    \includegraphics[width=\linewidth,
                     trim=0 1 0 1,clip]
                     {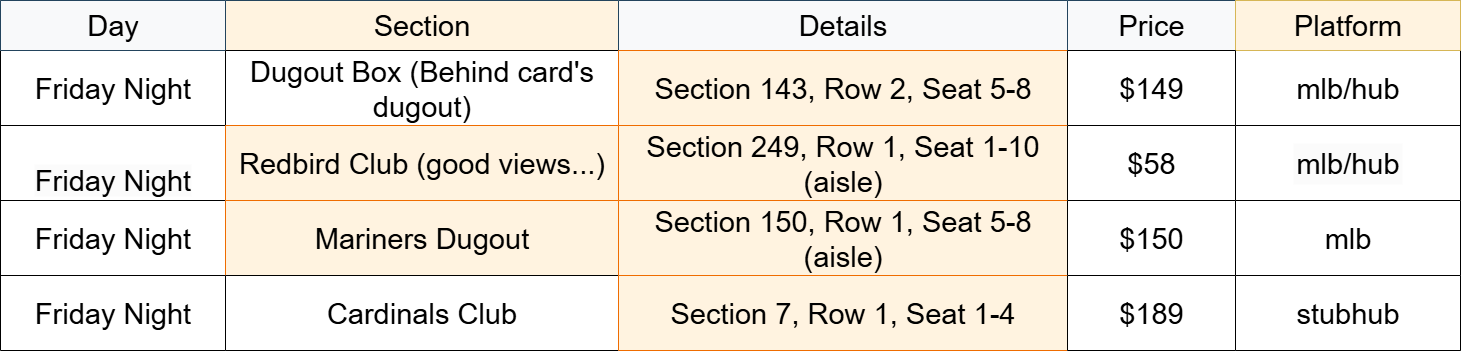}
  \end{subfigure}
{\small{Table predicted by LLM Run \#2}}
  \caption{Comparison of tables from two LLM runs on identical input: Run \#1 extracts detailed fields, with red shading marking cells split from composite strings. Run \#2 creates a coarser structure by combining attributes into broader fields. Color coding shows distinct interpretations and restructuring of the same unstructured input.}
  \label{fig:comparison}
\end{figure}

\par Figure~\ref{fig:comparison} illustrates the challenges of using purely neural solutions for structured table extraction. The top panel contains text showing some tabular seating data and the middle and bottom panels show outputs from two separate LLM invocations on this text input, producing inconsistent tables. Run \#1 generates a seven-column table, separating section, row, and seat information into distinct columns. Run \#2, however, uses five columns, merging section, row, and seat information into a single ``Details'' column. This highlights inconsistencies in data modeling and parsing strategies when LLM output lacks guided constraints. 

\par Moreover, the two runs reveal discrepancies in semantic interpretation and text handling. For instance, Run \#1 misinterprets the venue name as ``Category,'' while Run \#2 labels it as ``Section.'' Additionally, Run \#2 truncates descriptive text, reducing ``good views, dedicated bathrooms \& concessions'' to ``good views\ldots.'' These issues reflect underlying model hallucinations and the absence of strict structural constraints, leading to non-deterministic outputs that compromise reproducibility and require costly downstream corrections.

To address the shortcomings of both purely symbolic and purely neural methods, we propose \textbf{\sysname}, an \emph{iterative neuro‑symbolic feedback loop} that unites LLM‑driven inference with rule‑based validation. The process begins with an LLM that drafts a candidate table from raw text. This table is evaluated by a suite of symbolic sanity-checking rules that check for structural violations. The output of these symbolic checks is then presented to a secondary \textit{Critique LLM} tasked specifically with analyzing these feedbacks and translating them into actionable feedback. This critique explicitly guides targeted corrections for subsequent table regeneration. Thus, by orchestrating a continuous dialogue between deterministic rules for integrity and neural inference for layout flexibility, \sysname delivers tables that are both structurally sound and semantically faithful, outperforming one‑shot LLM outputs and static symbolic systems alike.

%

This paper makes the following key contributions:
(1) We show that domain-specialized chain-of-thought (CoT) prompting, which takes the form of Structural Decomposition prompting here, significantly outperforms generic CoT prompting and naive task prompting. 
(2) We show that generating feedback using a  symbolic validator coupled with a critique LLM outperforms other strategies for generating feedback.
(3) We show that our neurosymbolic approach, \sysname, produces tables that users prefer over tables generated by baseline approaches.


%% file: Sections/RelatedWork.tex
\section{Related Work}
\label{sec:related}

\subsubsection{Table Extraction from Unstructured Sources.}

Early approaches in table extraction leveraged visual layout-based heuristics and spatial positional clustering to recover structure; for example, see
\textsc{Tabula}~\cite{tabula}, 
PDFMiner~\cite{pdfminer},
TableFormer~\cite{tableformer}, 
TableNet~\cite{tablenet}, and 
layout-aware transformers~\cite{layoutlm, layoutlmv2}, and also
datasets such as~\cite{pubtabnet,tablebank,scitsr}.
While these approaches excel at extracting structured tables from well-formatted document images, they are inherently designed for image-based inputs and do not directly apply to scenarios involving unstructured or noisy text.

Recent advances have explored the direct generation of tables from unstructured textual data. Prominent works include the sequence-to-sequence framework, Text-to-table~\cite{wu-etal-2022-text-table},
an approach based on extracting tuples and then organizing them into tables,
Text-Tuple-Table~\cite{deng2024texttupletable},
and
Tabularis Revilio~\cite{mukul} that uses a neurosymbolic system 
that first detects headers and an initial table sketch using an LLM, and 
then uses an enumerate-and-test strategy for table reconstruction. 

In contrast, \sysname is an iterative neuro-symbolic pipeline. It employs LLMs to generate initial table hypotheses and a symbolic layer to validate structure. Unlike prior approaches that often produce tables in a single pass, \sysname emphasizes a feedback-driven mechanism where symbolic validation informs subsequent LLM generations. 

\subsubsection{Table Interpretation and Structure Recovery.}

Beyond basic extraction, table understanding involves inferring the semantic and structural roles of table components such as identifying headers, detecting units or entities, and aligning cells correctly. Prior work has addressed these through both heuristic and learned approaches.

Rule-based systems use patterns like casing, punctuation, or lexical cues to detect headers and types. Meanwhile, contextual models \cite{dong2019semantic,tabert,turl,tapas} learn joint representations over tables and surrounding text to support classification, question answering, or schema understanding. More recent models treat structure recovery as a sequence labeling or pointer-based generation task \cite{khang2025tfloptablestructurerecognition}

Our work explicitly addresses malformed or noisy table extractions. It introduces a feedback-driven loop wherein structure understanding is not assumed but incrementally induced through critique and regeneration. The symbolic sanity checker plays a central role in identifying inconsistencies, while the LLM-based critique proposes semantically coherent adjustments to improve structure over iterations.

\subsubsection{Self-Refinement and Feedback-Driven LLMs.}
Recent research has shown that LLMs can improve their outputs through iterative feedback mechanisms, mirroring human revision cycles. This includes both self-evaluation and externalized critique processes. \textsc{Self-Refine} \cite{madaan2023selfrefine} and \textsc{Reflexion} \cite{shinn2023reflexion} demonstrate that LLMs can iteratively critique and revise their responses, achieving better accuracy over successive attempts. Similarly, reasoning frameworks such as \textsc{Chain-of-Thought} \cite{wei2022chain} and \textsc{Tree-of-Thought} \cite{yao2023tree} introduce intermediate reasoning steps to support answer verification and path exploration.

External feedback mechanisms have also gained traction. \textsc{CRITIC} \cite{gou2024critic} integrates symbolic tools with LLMs for output assessment and refinement. \textsc{LLMRefine} \cite{liu2024llmrefine} generates fine-grained feedback on partially incorrect outputs using LLMs that are used to revise the original responses.

\sysname builds on these ideas but tailors them to the unique challenges of table extraction. Using LLMs only to generate feedback produces generic feedback and tends to accentuate LLM biases. 
Using symbolic tools only to generate feedback results in nonactionable feedback.
By using both a symbolic checker and a Critique LLM, we ensure that feedback is targeted and contextualized.
This design avoids over-correction and improves stability during iteration.

\subsubsection{Symbolic Evaluation and Sanity Checking of Tabular Outputs.}
Symbolic techniques are integral to the analysis and repair of tabular data, particularly when robustness and interpretability are required~\cite{wrangler}. 
This symbolic lens becomes even more critical in the context of table generation by LLMs. Prior works have noted the brittleness of LLM-generated tables \cite{10.1145/3571730}, particularly under noisy inputs or ambiguous formatting. Moreover, widely used evaluation metrics such as BLEU or string overlap \cite{parikh-etal-2020-totto} often fail to capture deep structural fidelity.

\sysname introduces a symbolic sanity checker specifically optimized for structural anomalies in generated tables. Beyond producing binary validity judgments, it outputs detailed signals (e.g., mismatched signatures, broken parenthesis, entity inconsistency) that highlight local violations. These signals are fed directly into the feedback loop via the Critique LLM, rather than being used post hoc or discarded. This tight coupling enables dynamic, interpretable refinement and supports robust convergence to high-fidelity table outputs.

%% file: Sections/TableExplicitizationProblem.tex
\section{The Table Explicitization Problem}
\par The goal of \emph{table explicitization}  is to transform an unstructured or semi-structured input string \(T\) into a structured format \(T'\). In \(T\), tables are \textbf{implicitly} represented, lacking clear row and column delimiters, making automated extraction difficult and error-prone. 

For instance, the input below presents medical data as a block of text without explicit structure:

\noindent \textbf{Input text \(T\) (Unstructured PDF-style content):}

\begin{minted}[frame=lines,fontsize=\small,breaklines=true,bgcolor=gray!5]{text}
Parameter Time (relative to transplantation time) NODAT NoNODAT p value
Before 95±15.97 88.36±21.97 0.043
Fasting plasma glucose (mg/dL) 1 month after 164.26±64.68 86.78±14.59 0.001
6 months after 135.57±66.32 88.44±12.63 0.001
Before 100.78±57.81 94.27±51.85 0.324
Triglycerides (mmol/L) 1 month after 203.78±88.21 167.16±76.71 0.015
6 months after 166.78±95.93 129.43±64.03 0.014
Before 136.35±47.80 146.32±104.2 0.475
Total cholesterol (mmol/L) 1 month after 185.67±57.73 163.84±48.52 0.02
6 months after 184.36±103.13 150.09±50.62 0.002
\end{minted}

\par This unstructured data lacks consistent boundaries for rows, columns, and hierarchical relationships (e.g., time points under parameters like \textit{Fasting plasma glucose}). Traditional extractors fail to parse such irregular layouts accurately.

\par The explicitized output \(T'\) restructures this into a well-defined structured format:

\noindent \textbf{Output \(T'\) (Explicit table in JSON):}
\begin{minted}[frame=lines, fontsize=\small, breaklines=true, bgcolor=gray!5]{json}
{
  "Fasting plasma glucose (mg/dL)": [
    {"Time": "Before", "NODAT": "95±15.97", "NoNODAT": "88.36±21.97", "p": 0.043},
    {"Time": "1 month after", "NODAT": "164.26±64.68", "NoNODAT": "86.78±14.59", "p": 0.001},
    {"Time": "6 months after", "NODAT": "135.57±66.32", "NoNODAT": "88.44±12.63", "p": 0.001}
  ],
...
}
\end{minted}

\par In \(T'\), each field is clearly defined, making further automated processing or analysis much more reliable. The transformation from \(T\) to \(T'\) illustrates the essence of the \emph{table explicitization} problem, which aims to automate the conversion of messy, unstructured text into reliable, structured data. The challenge lies in handling noisy, irregular inputs and correctly inferring the underlying table structure.

%% file: Sections/Approach.tex
\section{The \sysname\ Approach}

\par
\begin{figure}[t]
  \includegraphics[width=0.5\textwidth]{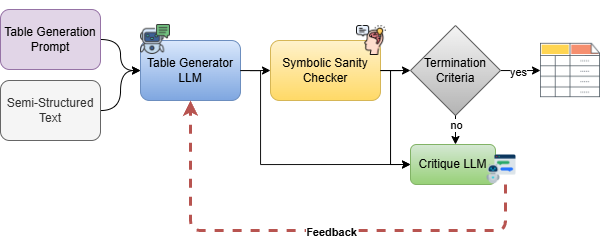}
  \caption{{\small{Overview of \sysname: The 3 main components are the table generator that uses structural decomposition prompting, the sanity checker that evaluates generations, and the critique LLM that produces feedback.}}}
  \label{fig:overview}
\end{figure}

 The overall high-level approach of \sysname is shown in Figure \ref{fig:overview}. The main idea is to utilize a pretrained large language model (Table Generator LLM) to simultaneously perform logical partitioning and initial generation of explicit tabular data from the unstructured input text. To ensure the structural integrity and correctness, we then apply ``Symbolic Sanity Checker'' designed to detect common extraction issues. The detected symbolic violations, along with the initial tables, are subsequently passed to a secondary ``Critique LLM''. This LLM analyzes the symbolic feedback and generates precise, actionable critiques, suggesting targeted improvements. These critiques guide further refinements by prompting the table-generating LLM to produce improved table representations iteratively. This iterative neuro-symbolic feedback loop continues until predefined convergence criteria are satisfied. For a detailed description of the algorithm used in \sysname, please refer to Appendix \ref{app:algo}.

\subsection{Structural Decomposition Prompting}
\textit{Structural Decomposition Prompting} is a specialized chain-of-thought (COT) approach that explicitly 
asks the LLM to first identify table-like regions in the input text and generate table delimiters before generating a structured representation for each table. This mirrors human table extraction by segmenting content based on layout and semantics before reconstructing it into structured form. The model is also asked to think about the row delimiters it will use. See Appendix~\ref{app:structural-decomposition} for an example applied to a PDF table.
\par This prompt design eliminates the need for brittle layout heuristics by leveraging the LLM's ability to use contextual cues holistically in detecting natural table boundaries and preserving semantic coherence across rows. The LLM is instructed to prioritize recall in this phase, capturing all visible content plausibly associated with tabular structure without introducing hallucinated entries. The outputs of this stage are a set of partial tables, which are then concatenated to form a unified tabular structured representation. The prompt design is detailed in Appendix~\ref{app:structural-decomposition_prompt}.

\subsection{Symbolic Sanity Checker}


The symbolic sanity checker applies a set of symbolic rules to assess the quality of a given table. The rules are handcrafted and domain-agnostic. The input to the sanity checker is the predicted table and the input text. The sanity checker returns a measure of coverage, hallucination rate, structure goodness/badness and a list of failing rules.

\textbf{Entity Consistency Rules}: 
For each row and column, this rule checks if all data values in that row/column match a entity regular expression. We use a fixed collection of entities, such as date, time, email, url, and word.
    
\textbf{Signature-Based Column Analysis}: When a column does not contain a predefined entity, 
we use syntactic features, such as presence or absence of special characters (digits, letters, punctuation, whitespace) to detect inconsistencies, such as text in a numeric column or unexpected delimiters.
    
\textbf{Merged Cell Detection}: 
A common and subtle table extraction error is the merging of adjacent cells, where column boundaries are lost and values collapse into one cell. Though no data is lost, column alignment is distorted, hindering structural recovery. We detect such cases by flagging cells with multiple numeric tokens (e.g., “102, 205”) or mixed formats (e.g., “Revenue 750”), which suggest improper merges. 


\textbf{Delimiter-Induced Errors}: 
Another frequent table extraction error arises from using the wrong column separator; for example, the number “12,345” splitting into “12” and “345”. Such subtle issues often evade column-wise checks, so we detect them by analyzing adjacent row cells for numeric fragmentation patterns.

\textbf{Parenthesis and Bracket Matching}: Mismatched brackets are indicators of incomplete or corrupted cell content. These issues often arise when the extraction process truncates cells, skips tokens, or fragments lines across row boundaries. To catch such anomalies, we scan and flag all cells containing unbalanced opening and closing symbols.
    
 \textbf{Empty Row Detection}: Rows with only whitespace or empty strings are identified and marked, as they usually indicate a malformed table region.
\par The sanity checker aggregates the findings from the above rules into two metrics: \textbf{goodness} score is the proportion of cells in the full table that have some desirable property (entity or signature consistency), and \textbf{badness} score is the maximum proportion of cells in a column that have some undesirable property (possible merged cell, incorrect delimiter, etc.). The rule violations are noted and returned, and subsequently forwarded to the critique LLM.
In addition to goodness and badness scores, the sanity checker also computes \textbf{coverage}: how much of the original input text appears in the predicted table, and \textbf{hallucination rate}: how much of the predicted table cannot be traced to the input text. The 4 scores are used to decide if the feedback loop should terminate.

\subsection{Critique LLM}

\begin{figure}
  {\includegraphics[width=\linewidth]{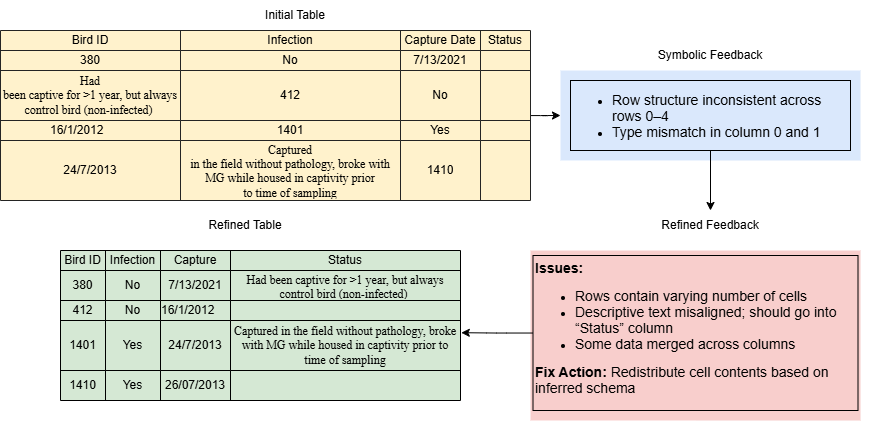}}
  \caption{\small{
  Example of table generation and refinement in \sysname. The initial table contains structural inconsistencies, such as uneven row lengths and merged data cells. Symbolic feedback identifies these issues and infers a corrected structure, triggering a refinement step. Cells are then reorganized to produce a normalized output table with semantically coherent columns.
  }}
  \label{fig:feedback_loop}
\end{figure}

\par While the symbolic checker is effective in identifying structural inconsistencies, as seen in the example in Figure \ref{fig:feedback_loop} where uneven row lengths and merged cells are flagged, its reliance on surface-level patterns can lead to false positives, flagging correct tables due to formatting deviations. If such unfiltered symbolic feedback were passed directly to the table regeneration step, it could lead to over-correction or cause the system to enter an unproductive loop of unnecessary transformations.

\par To address this, we introduce a Critique LLM that acts as a flexible, reasoning-aware intermediary. It interprets symbolic feedback contextually, evaluating the table 
and feedback holistically to produce a natural language critique with actionable improvements if needed. 

\par The Critique LLM assesses structural soundness based on semantic and visual coherence, decides whether to accept, refine, or disregard feedback, and suggests targeted fixes like header realignment or cell separation while preserving data. 
%
Thus, we complete one iteration of \sysname's neuro-symbolic refinement loop:
\emph{initial table generation → symbolic checking → LLM-based critique → table regeneration}. The loop continues until the sanity checker accepts the table or we reach the maximum number of allowed iterations. 

%% file: Sections/Evaluation.tex
\section{Evaluation}
\newcommand{\added}[1]{\textcolor{blue}{#1}}  
\newcommand{\userstudybaseline}{{\fontfamily{ptm}\selectfont\textsc{TenCOT}}\xspace}

\par The goal of our evaluation is to assess the effectiveness of \sysname, and its components, in producing high-fidelity, structurally coherent tables from noisy and unstructured text. 

\begin{table}[t]
\centering
\small
\begin{tabular}{lrrrrrr}
\toprule
\textbf{Dataset} & \textbf{\# Tasks} & \textbf{\# Tokens} & \textbf{\# Rows} & \textbf{\# Columns} \\
\midrule
FinTabNet   & 1000  & 111.41 & 13.54 & 2.59 \\
PubTabNet   & 1000  & 120.3 & 14.34 & 4.09 \\
BrokenCSV   & 1000  & 128734.3 & 14871.65 & 148.65  \\
WikiTables  & 1000  & 219.81 & 16.06  & 6.2  \\
FinReCon20 & 20 & 217.55 & 52.1 & 4.7 \\
\bottomrule
\end{tabular}
    \caption{\small{Statistics of datasets used in evaluation: the number of tasks, and average number of tokens, rows, and columns per table.}}
\label{tab:dataset-summary}
\end{table}

{\textbf{Datasets}}.
To assess \sysname, we chose benchmarks with varied structural complexity, formatting noise, and semantic richness, summarized in Table \ref{tab:dataset-summary}.
Specifically, we obtained our benchmarks from 
{\bf{PubTabNet}}~\cite{pubtabnet} (using the raw OCR text as the input),
{\bf{FinTabNet}}~\cite{fintabnet} (using OCR text as the input),
{\textbf{WikiTables}}~\cite{wikitables},
{\bf{BrokenCSV}}~\cite{clevercsv},
and \textbf{FinRecon20}. FinRecon20 is a dataset of 20 tables manually curated by us from real-world financial documents that is used exclusively for the user study.

{\textbf{Baselines}}.
We evaluate \sysname using six representative language models: {\bf{GPT-4}}, {\bf{DeepSeek-R1}}, {\bf{o3-mini}}, {\bf{DeepSeek-V3-0324}}, {\bf{Phi-4}}, and {\bf{Mistral-Small-2503}}. 
We also compare against two baselines designed for structured table reconstruction: a neurosymbolic method, namely Tabularis {\bf{Revilio}}~\cite{mukul}, 
and a symbolic method, namely {\bf{SplitText}}~\cite{raza-17,raza2020web}. 
%
While \sysname may seem related to prior models such as \cite{deng2024texttupletable, wu-etal-2022-text-table}, their inputs are narrative-style paragraph descriptions that contain some information that can be extracted as a table, which is
substantially different from our inputs, and hence they are not included as baselines.

{\textbf{Evaluation Metrics}}.
We adopt both structural and semantic metrics to comprehensively evaluate the quality of the generated tables.
\textbf{Coverage} is the fraction of alphanumeric characters from the input that are preserved in the output table: \(
\text{\textbf{Coverage}} = 1 - \frac{\lvert U \rvert}{\lvert S \rvert}
\) where $\lvert S \rvert$ is the number of alphanumeric characters in the source input, and $\lvert U \rvert$ is the number of uncovered characters based on string matching.
\textbf{Hallucination Rate} is the average fraction of unmatched content across all cells in the generated table. For a table with $R$ rows, each containing $C_i$ cells, it is computed as: \(
\text{\textbf{HallucinationRate}} = \frac{1}{R} \sum_{i=1}^{R} \left( \frac{1}{C_i} \sum_{j=1}^{C_i}(1 - \text{cellCoverage}_{ij}) \right)
\), where $cellCoverage_{ij} \in [0,1]$ denotes the degree to which cell(i,j) matches the input source(including partial matches).
\textbf{Exact Match(EM)} is a binary indicator that evaluates to 1 if the predicted table matches the ground-truth table exactly in both structure and content, and 0 otherwise. \textbf{Tree Edit Distance (TED)} is the normalized cell-level edit distance between the predicted and ground-truth tables \cite{ted}. \textbf{Column Match (Col.V.M.)} is defined as the average percentage of columns that are exactly matched between the predicted and reference tables, micro-averaged per table and then averaged across all tables. \textbf{Value Match (C.V.M.)} reports the average percentage of individual cell values that are correctly reconstructed across all tables.

{\textbf{User Study}}. We recruited 21 active Spreadsheet users for 2-4 sessions. In each session, the user was shown a ground-truth table and two predicted tables extracted from textual representations of the ground-truth table using 2 tools. The users were asked to (partially) edit the predicted tables to make them look like the ground-truth table. This process was video recorded. The users were then asked to fill a questionnaire on the (1) perceived accuracy of the predicted table, (2) effort required to verify the accuracy, and (3) effort required to fix the table. See Appendix~\ref{sec:app_eval} and \ref{sec:questionnaire} for more details.

\ignore{ 

\begin{table}[t]  
\centering\scriptsize  
\setlength{\tabcolsep}{4pt}  
\begin{tabular}{@{}lp{7cm}@{}}  
\toprule 
\ \textbf{Code} & \textbf{Description} \\  
\midrule
\multicolumn{2}{c}{\textbf{Action Type}} \\  
\midrule  
Add & Adding values to the replication that were present in the source table but missing from the replication \\  
Modify & Fixing incomplete or incorrect values within individual cells in the replication \\
Delete & Removing values from the replication that were not present in the source table \\ 
    Realign & Shifting cells that were incorrectly aligned in the replication \\
\midrule  
\multicolumn{2}{c}{\textbf{Data Type}} \\  
\midrule  
Column Heading & Heading for a column \\  
Row Heading & Heading for a row \\  
Subheading & Heading not for a particular row or column but for a subsection of the spreadsheet \\   
Note & Values that are not associated with a particular row or column, but describe the data in the spreadsheet (e.g., 'values in thousands') \\  
Data & All non-header, non-note values\\
\bottomrule  
\end{tabular}  
\caption{Actions performed by user study participants to fix table regenerations. Each action consisted of an action type as well as the data type upon which the action was performed}  
\label{tab:annotation_labels}  
\end{table}  

\endignore}


%% file: Sections/Results.tex

\section{Results and Analysis}
\begin{figure*}
    \centering
    \includegraphics[width=1\linewidth]{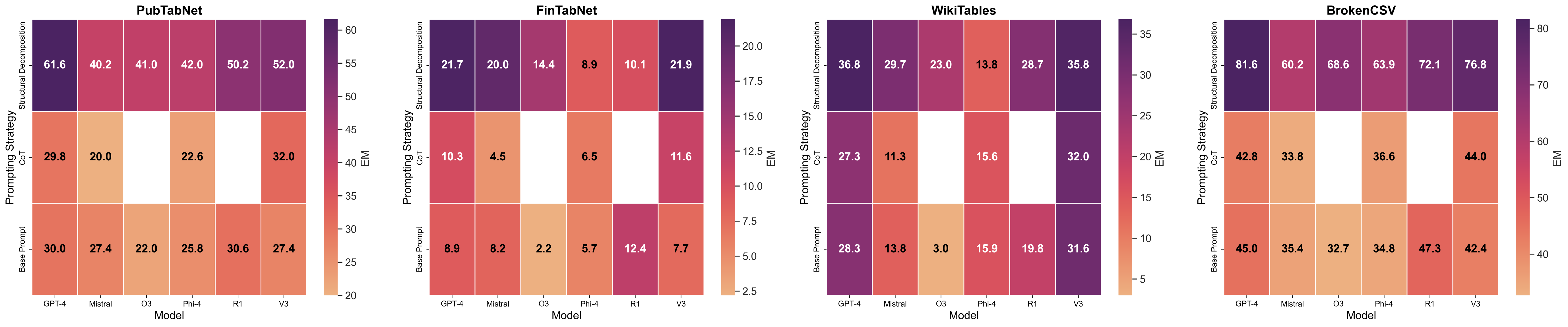}
    \caption{{\small{Exact Match (EM) scores (darker is better) across three prompting strategies: Base Prompt (bottom row), CoT (middle row), and Structural Decomposition (top row) and six LLMs: GPT-4, Mistral-Small-2503, o3-mini, Phi-4, DeepSeek-R1, DeepSeek-V3-0324 (left to right on x-axis) on four datasets: PubTabNet, FinTabNet, WikiTables, and BrokenCSV (left to right).}}}
    \label{fig:em_heatmap}
\end{figure*}
\begin{figure}
    \centering
    \includegraphics[width=\linewidth]{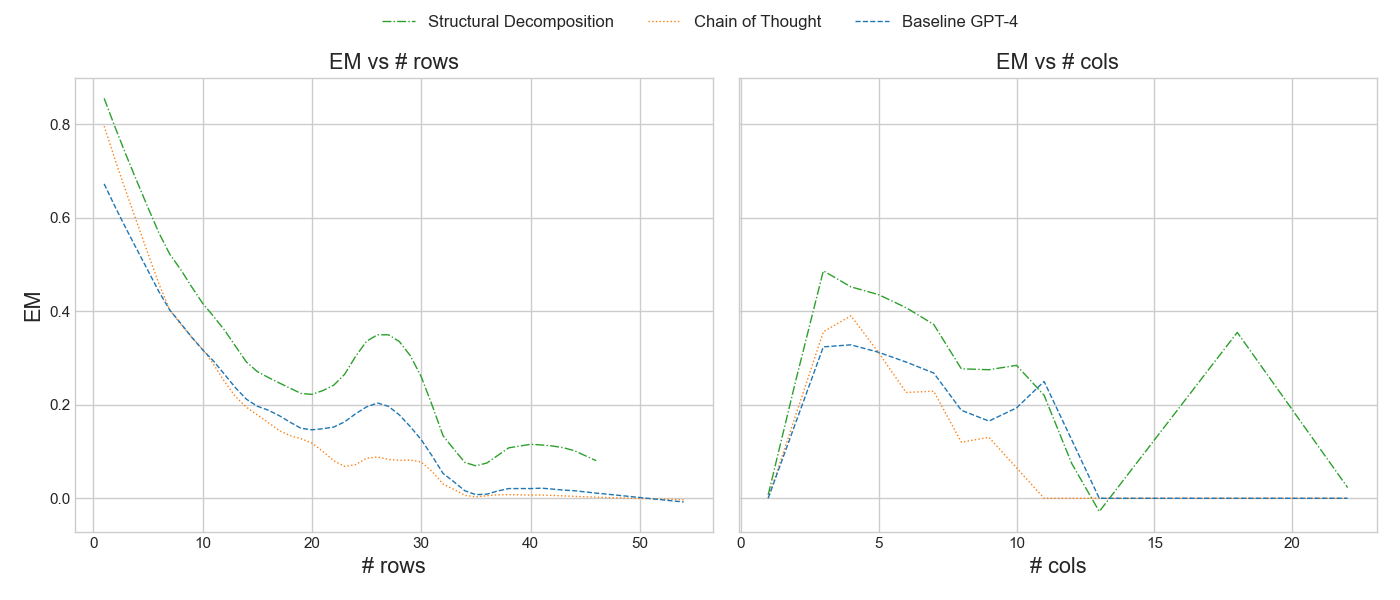}
    \caption{{\small{Effect of table size (x-axis) on EM accuracy (y-axis) across prompting strategies. We report mean EM scores grouped by table row count (left) and column count (right).}}}
    \label{fig:size_em_analysis}
\end{figure}
\textbf{RQ1}: \textit{How does the prompting strategy and model choice influence performance?}\\
\textbf{``Structure first'' beats ``reason more''}: We evaluated the Structural Decomposition prompt (SD) across four datasets and six models, comparing it against a baseline prompt (instructs the model to construct a table) and a Chain-of-Thought (CoT) prompt (instructs the model to also think step by step). Results in Figure~\ref{fig:em_heatmap} demonstrate that SD consistently outperforms both baseline and CoT, with improvements of 10–40 EM points and 8-15 points in ColVM/CVM metrics. CoT offers marginal gains over the baseline.
Complete results are in Appendix~\ref{sec:app_results}.\\
\textbf{Model Capacity Matters}:
Larger models consistently outperform smaller models across all prompting strategies. However, SD substantially narrows this performance gap. For instance, o3-mini achieves only 2.2\% EM on FinTabNet using the Base Prompt, but rises to 14.4\% with SD (6x), while GPT-4 improves from 8.9\% to 21.7\% (2.4×), indicating that SD disproportionately benefits lower-capacity models. This trend is consistent across datasets. 
\\
\textbf{Large tables are difficult:}
Figure~\ref{fig:size_em_analysis} shows EM scores declining noticeably as the number of rows increases. Similarly, performance peaks at moderate column widths (about 6–8 columns) and deteriorates for wider tables, likely due to increased ambiguity in column alignment and value placement. SD mitigates these effects to some extent. Larger models exhibit a more gradual degradation in accuracy with size. 
\\
\textbf{Dataset Complexity and Performance Characteristics}:
Based on LLM performance, the datasets can be ordered in difficulty level as
BrokenCSV $<$ PubTabNet $<$ WikiTables $<$ FinTabNet. We observe that the ``easier'' datasets have contextual information or descriptive entries that make it easier to recover structure even when spatial and visual layout information is not present. \\
\textbf{Iterative Feedback Loop Performance Analysis}: 
We observe the largest gains in EM, CVM and ColVM metrics in the first two to three iterations - consistent with several past works; see 
Figure~\ref{fig:metric_convergence} in Appendix~\ref{sec:app_results}.\\
\textbf{Robustness \& Statistical Significance}: We conducted statistical analyses to confirm that our observations above are statistically significant; see
details in Appendix~\ref{sec:app_results}.

\begin{table}[t]
\centering
\scriptsize
\setlength{\tabcolsep}{4pt}
\begin{tabular}{@{}llcccc@{}}
\toprule
\textbf{Method} & \textbf{Metric} & \textbf{PubTabNet} & \textbf{FinTabNet} & \textbf{WikiTables} & \textbf{BrokenCSV} \\
\midrule
\multirow{4}{*}{\sysname(GPT-4o)} 
& TED      & 0.67 & 0.47 & 0.65 & 0.22  \\
& EM        & 61.60 & 21.7 & 36.80 & 81.6  \\
& C.V.M.    & 71 & 33 & 64 & 81  \\
& Col.V.M. & 75 & 29 & 56 & 69  \\

\midrule
\multirow{4}{*}{Revilio} 
& TED       & 0.60 & 0.53 & 0.59 & 0.17 \\
& EM        & 52.34 & 15.65 & 33.43 & 60.96 \\
& C.V.M.    & 66 & 29 & 59 & 76 \\
& Col.V.M.  & 60 & 23 & 51 & 63 \\
\midrule
\multirow{4}{*}{SplitText} 
& TED       & 0.55 & 0.58 & -- & 0.43 \\
& EM        & 26 & 11 & 1.1 & 53.8 \\
& C.V.M.    & 37 & 22 & 1 & 46 \\
& Col.V.M.  & 35 & 19 & 8 & 40 \\
\bottomrule
\end{tabular}
\caption{\small{Comparison of \sysname with existing baselines on table reconstruction accuracy.}}
\label{tab:baseline_comparison}
\end{table}

\smallskip\noindent
\textbf{RQ2}: \textit{How does \sysname compare to existing baselines?}\\
Table~\ref{tab:baseline_comparison}
shows that \sysname consistently outperforms both Revilio and SplitText across four benchmarks 
on content-fidelity metrics (EM, CVM, ColVM). On PubTabNet, \sysname improves EM by +9.26 pts over Revilio, together with higher CVM (+5 pt) and ColVM (+15 pt). On the more challenging FinTabNet, absolute EM is lower for all methods, but \sysname still yields a +6.05 pt EM gain and higher CVM/ColVM. On WikiTables, \sysname delivers smaller yet consistent improvements (+3.37 EM; +5 CVM; +5 ColVM). The largest gains appear on BrokenCSV, where \sysname attains 81.6 EM (vs. 60.96 for Revilio and 53.8 for SplitText), alongside higher CVM/ColVM.

Structural alignment, measured by TED, shows mixed results. 
This suggests that \sysname
can sometimes normalize header layouts or handle spans in ways that improve content agreement (EM/CVM/ColVM) yet diverge modestly from the reference tree, reflecting granularity differences rather than content errors. 
%
The TED metric rewards structural isomorphism and penalizes benign layout normalization. 
When \sysname consolidates multi-row headers, propagates or flattens spans, removes empty scaffold rows/columns, or relocates footnotes and units, the values are preserved but the parse tree changes. Consequently, TED can worsen even as EM, CVM, and ColVM improve, so it should be interpreted alongside content-fidelity metrics.

\begin{figure}
    \centering
    \includegraphics[width=1\linewidth]{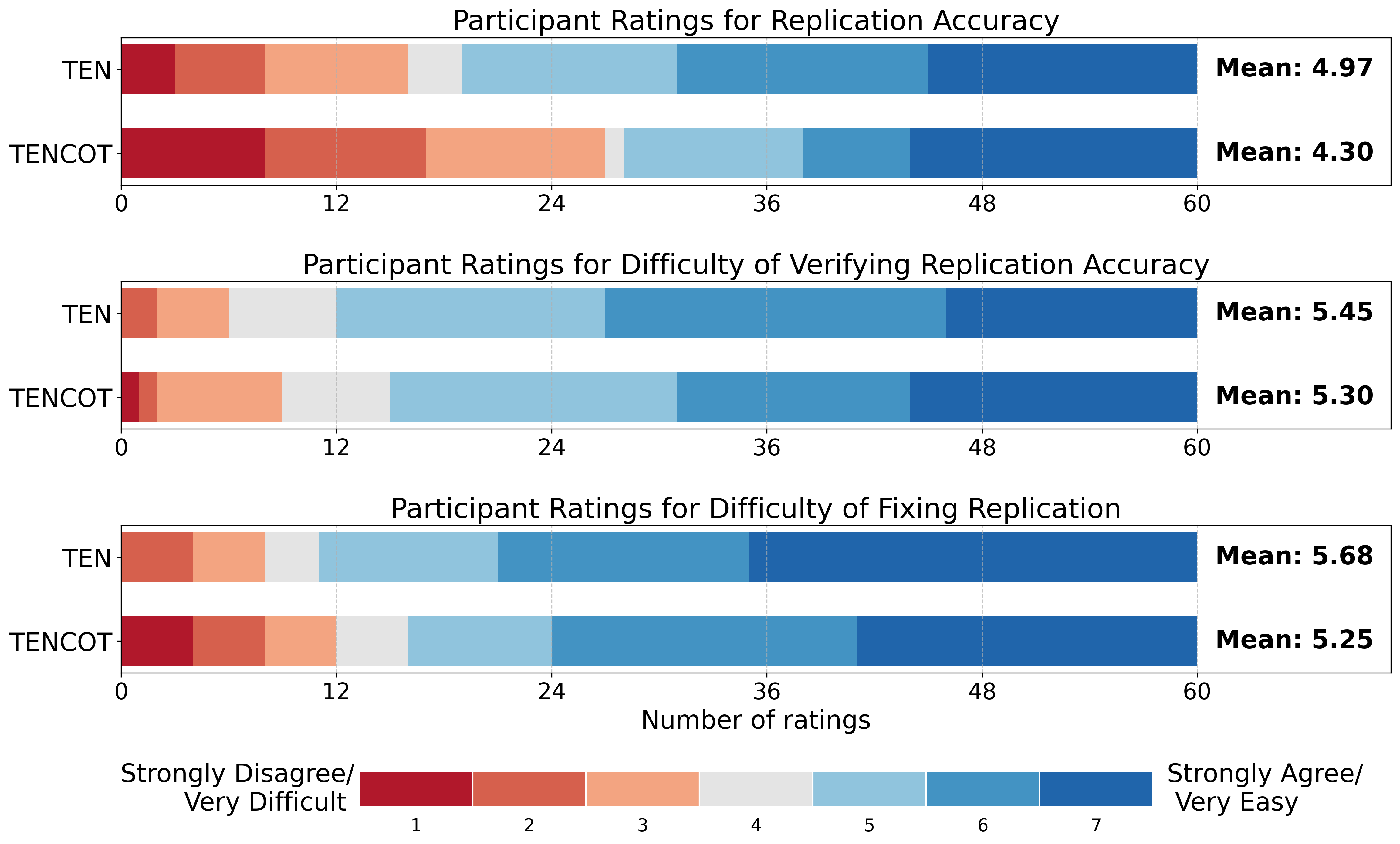}
    \caption{Questionnaire responses in user study}
    \label{fig:annotations_final}
\end{figure}

\smallskip\noindent
\textbf{RQ3}: \textit{Is \sysname useful in practice?}\\
We analyzed data from a user study where participants were given tables generated by two variants of \sysname: original \sysname and a slightly weaker version of \sysname where we use CoT prompt (instead of SD) that we call \userstudybaseline.
%
%
Figure \ref{fig:annotations_final} summarizes participants' responses to the different questions in the questionnaire provided to them. 
We found that on the Likert scale from 1-7, participants reported that the accuracy of tables generated by \sysname was significantly higher than of those generated by \userstudybaseline  (Wilcoxon signed-rank test, $z = 284.0, p = .02$) with a moderate effect size ($r = -0.30)$. 
Further, in 7 out the 20 tables used for the study, all three reviewers independently rated \sysname to be more accurate than \userstudybaseline, while all participants preferred \userstudybaseline for only 1 table. 
%

Despite this difference in responses on the tables' accuracy, we did not observe any significant differences in participants' responses on the effort required to verify ($z = 260.0, p = .51$) or fix ($z = 266.0, p = .08$) these tables. This might be as while \sysname required users to often fix alignment issues, \userstudybaseline often required users to add missing data, both of which led to increased effort in fixing the replication; see details in Appendix~\ref{sec:app_results}.

\ignore{
\par \textbf{Participant Actions:} Based on the codes in Table \ref{tab:annotation_labels}, we coded a total of 385 actions performed by participants (179 with tables generated by \sysname vs 206 with tables generated by \userstudybaseline). To assess the effect of each type of action and type of data (Table \ref{tab:annotation_labels}) on the increase in effort by participants to verify and fix the table, we perform statistical tests on those samples for which participants performed the action for one tool and did not perform the action for the other tool.
Thus, for each test, one sample contains instances where participants performed the action on a particular table and one sample contains instances where participants did not perform the action for the same table (but a different tool). We do not control for which tool participants worked with in these samples to avoid any biases.
While creating such pairs leads to a reduction in our sample size, this ensures that our data is paired, and allows us to continue to use the Wilcoxon signed-rank test to test for this effect. 
The results of these tests across different types of actions and types of data operated upon are reported in the Appendix (Section \ref{sec:wil_stats}). From these tests, we can infer that participants reported that it was significantly more difficult to verify the accuracy of the replication when they were required to modify values or fix misaligned data in the replication. Further, in addition to modifying values and fixing misaligned data, participants reported that it was significantly more difficult to fix replications when adding new data.


Similarly, we found that participants reported that it was significantly more difficult to fix tables when performing operations on subheadings, row headings, and data values, and reported that it  was significantly more difficult to verify tables when performing operations on row headings and data values. This may be due to the fact that these types of data tend to be embedded within the table and thus may be difficult to locate and to operate on.

To analyze the occurence of the different types of actions and data that participants worked on across the two conditions, we performed McNemar's exact test \cite{McNemar1947}. The results of these tests are reported in the Appendix (Section \ref{sec:mcnemar}). We found that the number of times participants added new data to the replication was significantly higher when working on tables generated by \userstudybaseline as compared to those generated by \userstudysys. Further, participants operated on subheadings a significantly higher number of times when working on tables generated by \userstudybaseline. On the other hand, the number of times participants worked on fixing misaligned data was significantly higher when working on tables generated by \userstudysys. Upon performing the Wilcoxon signed-rank test, we found that participants rated the accuracy of replications to be significantly higher when the fixes to these replications did not require adding any data ($N = 35,z = 52.5, p = .002$) with a moderate effect size ($r = -0.53$). However, we did not find significant differences in the accuracy ratings based on fixing misaligned data in the replication ($N = 16, z  = 18.0, p = .051$). \footnote{$N$ is lesser than 60 for these tests as we only consider those tables for which one replication required performing the action, and the other one did not.} This suggests that participants may have reported that the tables generated by \userstudybaseline were less accurate due to missing data in these tables. However, since both adding missing data and fixing misaligned data significantly increased the effort required to fix the replication, each tool had its own drawbacks that increased the effort for users to verify and fix the replications generated by that tool.


TODO (Nikita): explain technically why structural decomposition (\userstudysys) was stricter in its generation, resulting in more alignment issues, while providing flexibility in \userstudybaseline led to more issues in missing data.

(Aayush to Nikita: please change the following paragraph to make it more accurate in terms of the technical approaches)
Our results thus indicate that these different approaches to reconstructing tables might be useful in different scenarios. When accuracy is critical, applying a stricter approach that prioritizes maximum coverage (possibly at the cost of alignment issues) might ensure that the replication is useful for users. On the other hand, to ensure that the replication is easy for users to verify, we should apply a more flexible approach that ensures that individual values are accurate and correctly aligned.

\endignore}

\begin{table}[t]
\centering
\scriptsize
\setlength{\tabcolsep}{4pt}
\begin{tabular}{@{}llcccc@{}}
\toprule
\textbf{\sysname Feedback} & \textbf{Metric} & \textbf{Wikipedia} & \textbf{BrokenCSV} & \textbf{PubTabNet} & \textbf{FinTabNet} \\
\midrule
\multirow{6}{*}{No feedback} 
& Coverage   & 0.92  & 0.94  & 0.93  & 0.90 \\
& Halluc.    & 0.13  & 0.17  & 0.18  & 0.17 \\
& TED        & 0.71  & 0.36  & 0.76  & 0.50 \\
& EM         & 15.18 & 60.31 & 17.16  & 12.91 \\
& C.V.M.     & 42.31 & 65.34 & 54.98 & 22.96 \\
& Col.V.M.   & 45.49 & 61.01 & 48.12 & 19.26 \\
\midrule
\multirow{6}{*}{Symbolic} 
& Coverage   & 0.94  & 0.96  & 0.93  & 0.91 \\
& Halluc.    & 0.11  & 0.14  & 0.15  & 0.15 \\
& TED        & 0.68  & 0.30  & 0.75  & 0.48 \\
& EM         & 20.36 & 68.93 & 21.28  & 20.34 \\
& C.V.M.     & 55.84 & 68.56 & 59.96 & 25.36 \\
& Col.V.M.   & 57.69 & 63.35 & 53.35 & 21.12 \\
\midrule
\multirow{6}{*}{LLM} 
& Coverage   & 0.94  & 0.96  & 0.94 & 0.92 \\
& Halluc.    & 0.11  & 0.15  & 0.11 & 0.11 \\
& TED        & 0.66  & 0.31  & 0.73 & 0.48  \\
& EM         & 27.93 & 72.45 & 22.83 & 18.69 \\
& C.V.M.     & 59.01 & 74.93  & 55.69 & 29.12 \\
& Col.V.M.   & 54.69 & 70.89  & 52.87 & 24.95 \\
\midrule
\multirow{6}{*}{Both} 
& Coverage   & 0.94  & 0.97 & 0.93   & 0.91 \\
& Halluc.    & 0.11  & 0.15 & 0.14   & 0.15 \\
& TED        & 0.65  & 0.28 & 0.67   & 0.47 \\
& EM         & 36.80 & 81.6 & 61.60  & 21.73 \\
& C.V.M.     & 64.36 & 81.03 & 71.54 & 33.92 \\
& Col.V.M.   & 59.59 & 69.56 & 75.69 & 29.36 \\
\bottomrule
\end{tabular}
 \caption{\small{Ablation Study for Variants of \sysname, EM, C.V.M., and Col.V.M. are reported as percentages.}}
\label{tab:ablationstudy}
\end{table}

\smallskip\noindent
\textbf{RQ4}: {\textit{What is the impact of different components of \sysname?}}\\
To assess the contributions of \sysname’s feedback mechanisms, we ablated its components and evaluated four variants under identical prompts, decoding, and evaluation setup: (i) No feedback disables all feedback; (ii) Symbolic feedback enables rule-based checks; (iii) LLM feedback uses critique-and-rewrite without symbolic validation; (iv) {\bf{Both}} combines both as in \sysname. Table~\ref{tab:ablationstudy}
shows that
compared to No feedback, Symbolic feedback consistently improves structural and content metrics, with EM gains of +5.18 (Wikipedia) and +8.62 (BrokenCSV). LLM feedback provides even larger gains on simpler or noisier datasets, e.g., +12.75 EM (Wikipedia). However, on complex tables, it is less stable and may underperform Symbolic feedback on alignment metrics.
The full system, {\bf{Both}}, achieves the best performance across all metrics. It offers the highest content fidelity (EM, CVM), lowest TED, and near-saturated coverage (0.90–0.97). In some cases, LLM feedback slightly reduces hallucination compared to the full system, but this comes at the cost of structural fidelity.

\textbf{Case Study Examples.} In one clinical table example from PubTabNet dataset (Appendix \ref{app:qual_analysis_1}), LLM-only output hallucinated a data column due to header misplacement, which NeuroSymbolic feedback corrected by promoting the label and aligning columns—improving EM and CVM while marginally affecting TED. In another example (Appendix \ref{app:qual_analysis_2}), LLM-only flattened group headers into categorical columns, which improved readability but increased TED and reduced ColVM. NeuroSymbolic feedback preserved the hierarchical structure, achieving better fidelity and lower error.

\section{Conclusion}
We presented a neurosymbolic approach for extracting tables from text that uses specialized
CoT prompting and self-debug loop with feedback generated by a critique LLM conditioned on results generated by a symbolic validity checker. The effectiveness of the approach was established through extensive evaluation and a user study. Ablation studies showed that the prompting strategy, the symbolic feedback component, and the critique LLM are all essential to the success of the overall approach.

%% file: appendix/Alogrithim.tex
\pagebreak
\section{\sysname's Algorithm}
\label{app:algo}
\par Algorithm \ref{alg:ten} outlines the main pipeline, which iteratively generates tables using Structural Decomposition prompting, validates them with symbolic rules, and refines them via critique feedback. The process initializes an empty table, generates an initial structure, and iterates until convergence or a maximum number of iterations is reached, guided by metrics such as coverage and hallucination rate. The \texttt{SymbolicSanityChecker} (Algorithm \ref{alg:symbolic_sanity_checker}) is a critical component, applying domain-agnostic rules to detect well formedness of a table. It computes metrics (e.g., goodness and badness scores) to inform refinement, ensuring robust table structures. Detailed prompts for table generation and critique are provided in Appendices \ref{app:structural-decomposition_prompt}, \ref{app:regeneration_prompt} and \ref{app:critique_generation}, respectively. 

\algrenewcommand\alglinenumber[1]{} 

\begin{algorithm*}
\caption{\sysname: Table 
Explicitization, Neurosymbolically}
\label{alg:ten}
\small{
    \begin{algorithmic}
    \Require Unstructured text $T$, max iterations $N$, convergence threshold $C$
    \Ensure Structured table $T'$
    \State Initialize $iteration\_{count} \gets 0$
    \State Initialize $table\_{candidate} \gets \emptyset$ \Comment{Empty table structure}
    \State Initialize $critique\_{feedback} \gets \emptyset$ \Comment{No feedback for first iteration}
    \Repeat
        \State $iteration\_{count} \gets iteration\_{count} + 1$
        \If{$iteration\_count = 1$}
            \State $table\_{candidate} \gets \Call{StructuralDecompositionPrompting}{T}$ \Comment{Identifies table-like regions, Appendix \ref{app:structural-decomposition_prompt}}
        \Else
            \State $table\_{candidate} \gets \Call{RefineTable}{T, table\_candidate, critique\_feedback}$ \Comment{Refines table based on feedback, Appendix \ref{app:regeneration_prompt}}
        \EndIf
        \State $validation\_{results} \gets \Call{SymbolicSanityChecker}{table\_{candidate}, T}$ 
        \If{\Call{IsConverged}{validation\_{results}, C}} 
            \State \textbf{break}
        \EndIf
        \State $critique\_{feedback} \gets \Call{CritiqueLLM}{table\_[candidate], validation\_{results}, T}$ \Comment{Generates actionable feedback, Appendix \ref{app:critique_generation},}
    \Until{$iteration\_{count} \geq N$}
    \State \Return $table\_candidate$
\end{algorithmic}
}
\end{algorithm*}

\begin{algorithm*}
\caption{SymbolicSanityChecker}
\label{alg:symbolic_sanity_checker}
\small
\begin{algorithmic}
\Procedure{SymbolicSanityChecker}{\texttt{table\_candidate}, $T$}
    \State Initialize $\texttt{violations} \gets \emptyset$, $\texttt{goodness\_score} \gets 0$, $\texttt{badness\_score} \gets 0$
    \State Initialize $\texttt{coverage} \gets 0$, $\texttt{hallucination\_rate} \gets 0$
    \State $\texttt{total\_cells} \gets \Call{CountCells}{\texttt{table\_candidate}}$
    \State $\texttt{consistent\_cells} \gets 0$, $\texttt{violating\_cells} \gets 0$

    \Statex $\quad\rhd$ \textbf{Empty Row Detection}
    \For{each row $r$ in $\texttt{table\_candidate}$}
        \If{\Call{IsEmptyRow}{r}} \Comment{Whitespace or empty strings}
            \State Add ``Empty Row'' to $\texttt{violations}$
            \State $\texttt{violating\_cells} \gets \texttt{violating\_cells} + \Call{CountCells}{r}$
        \EndIf
    \EndFor

    \Statex $\quad\rhd$ \textbf{Entity Consistency and Column Analysis}
    \For{each column $c$ in $\texttt{table\_candidate}$}
        \State $\texttt{entity\_type} \gets \Call{DetectEntityType}{c}$ 
        \If{\Call{MatchesEntityType}{c, \texttt{entity\_type}}}
            \State $\texttt{consistent\_cells} \gets \texttt{consistent\_cells} + \Call{CountCells}{c}$
        \Else
            \State $\texttt{inconsistencies} \gets \Call{AnalyzeColumnSyntax}{c}$
            \If{$\texttt{inconsistencies} \neq \emptyset$}
                \State Add ``Inconsistent Column'' to $\texttt{violations}$
                \State $\texttt{violating\_cells} \gets \texttt{violating\_cells} + \Call{CountInconsistentCells}{c}$
            \EndIf
        \EndIf
    \EndFor

    \Statex $\quad\rhd$ \textbf{Merged Cell, Delimiter, and Bracket Checks}
    \For{each cell $cell$ in $\texttt{table\_candidate}$}
        \If{\Call{HasMultipleEntityTokens}{cell}} 
            \State Add ``Merged Cell'' to $\texttt{violations}$
            \State $\texttt{violating\_cells} \gets \texttt{violating\_cells} + 1$
        \EndIf
        \If{\Call{HasUnbalancedBrackets}{cell}} 
            \State Add ``Unbalanced Brackets'' to $\texttt{violations}$
            \State $\texttt{violating\_cells} \gets \texttt{violating\_cells} + 1$
        \EndIf
        \If{\Call{HasDelimiterErrors}{cell}} 
            \State Add ``Delimiter Error'' to $\texttt{violations}$
            \State $\texttt{violating\_cells} \gets \texttt{violating\_cells} + 1$
        \EndIf
    \EndFor

    \Statex $\quad\rhd$ \textbf{Compute and Return Metrics}
    \State $\texttt{coverage} \gets \frac{\Call{CountTokensInTable}{T, \texttt{table\_candidate}}}{\Call{CountTokens}{T}}$
    \State $non\textunderscore input\_tokens \gets \Call{CountTokensNotInInput}{T, \texttt{table\_candidate}}$
    \State $total\textunderscore table\_tokens \gets \Call{CountTokens}{\texttt{table\_candidate}}$
    \State $\texttt{hallucination\_rate} \gets \frac{non\textunderscore input\_tokens}{total\textunderscore table\_tokens}$
    \State $\texttt{goodness\_score} \gets \frac{\texttt{consistent\_cells}}{\texttt{total\_cells}}$
    \State $\texttt{badness\_score} \gets \frac{\texttt{violating\_cells}}{\texttt{total\_cells}}$
    \State \Return $\{$
\State \quad $\texttt{violations},$
\State \quad $\texttt{coverage},$
\State \quad $\texttt{hallucination\textunderscore rate},$
\State \quad $\texttt{goodness\textunderscore score},$
\State \quad $\texttt{badness\textunderscore score}$
\State $\}$
\EndProcedure
\end{algorithmic}
\end{algorithm*}

%% file: appendix/app_evaluation.tex
\section{Evaluation}
\label{sec:app_eval}

\par The goal of our evaluation is to assess the effectiveness of \sysname in producing high-fidelity, structurally coherent tables from noisy and unstructured text. Specifically, our evaluation investigates two key objectives. \textbf{First}, we examine whether the integration of symbolic sanity checks and critique-guided regeneration leads to improved table quality by reducing hallucinations, and correcting structural inconsistencies. \textbf{Second}, we compare \sysname against symbolic-only pipelines and neural-only(LLM-based) baselines to highlight the advantages of combining rule-based structural validation with neural generation in an iterative refinement framework. \textbf{Third}, we conduct a mixed-methods empirical study with spreadsheet users to understand the practical usefulness and drawbacks of tables generated by \sysname.
We aim to answer the following research questions:\\
\textbf{RQ1}: To what degree do prompting strategies and model choice influence table-extraction performance when evaluated on datasets of escalating complexity?\\
\textbf{RQ2}: How accurately does \sysname reconstruct tables compared to existing baselines?\\
\textbf{RQ3}: How useful is \sysname at reconstructing tables from real-world financial documents?\\
\textbf{RQ4}: What is the impact of different components of \sysname?

\subsection{Datasets}

\par To assess \sysname, we chose benchmarks with varied structural complexity, formatting noise, and semantic richness, summarized in Table \ref{tab:app:dataset-summary}.\\
\textbf{PubTabNet:} The PubTabNet dataset \cite{pubtabnet} contain tables from open-source scientific articles with OCR annotations. We generate benchmark tasks by using the raw OCR output as unstructured table text to simulate real-world reconstruction challenges.\\
\textbf{FinTabNet:} The FinTabNet dataset \cite{fintabnet} consists of tables extracted from annual reports of S\&P 500 companies. Each table is annotated with fine-grained cell structure information obtained via token alignment between PDF and HTML versions of the documents. Similar to PubTabNet, we use the raw OCR text of each table to generate unstructured inputs for our benchmark.\\
\textbf{Wikipedia Tables:} \textit{WikiTables} dataset~\cite{wikitables} comprises structured tables from Wikipedia  with clear formatting and rich semantic content. Despite their structural simplicity, these tables often rely heavily on positional context, making them challenging for unstructured text processing. We convert each table into unstructured OCR-style text, following the same procedure used for PubTabNet.\\
\textbf{BrokenCSV:} The BrokenCSV dataset \cite{clevercsv}, consists of noisy and irregular CSV-formatted tables with inconsistent delimiters, and missing entries.\\
\textbf{FinRecon20}: While prior benchmarks focus on structure extraction from images, our task centers on a different challenge: \textit{reconstructing structured tables from free-form text}. To evaluate this, we introduce FinReCon20, a curated dataset of 20 tables manually extracted from real-world financial documents, including balance sheets, earnings reports, and disclosures. These tables were selected to reflect common copy-paste issues seen in document workflows, such as inconsistent delimiters, fragmented rows, multi-line cells, and token-level segmentation errors. Dataset details appear in Table \ref{tab:app:dataset-summary}. 

\begin{table}[t]
\centering
\small
\begin{tabular}{lrrrrrr}
\toprule
\textbf{Dataset} & \textbf{\# Tasks} & \textbf{\# Tokens} & \textbf{\# Rows} & \textbf{\# Columns} \\
\midrule
FinTabNet   & 1000  & 111.41 & 13.54 & 2.59 \\
PubTabNet   & 1000  & 120.3 & 14.34 & 4.09 \\
BrokenCSV   & 1000  & 128734.3 & 14871.65 & 148.65  \\
Wikipedia Tables  & 1000  & 219.81 & 16.06  & 6.2  \\
FinReCon20 & 20 & 217.55 & 52.1 & 4.7 \\
\bottomrule
\end{tabular}
\caption{Dataset statistics used in our experiments. We report the number of tasks, and average number of tokens, rows, and columns per table.}
\label{tab:app:dataset-summary}
\end{table}

\subsection{Baselines}
We evaluate \sysname using six representative language models: GPT-4, DeepSeek-R1, o3-mini, DeepSeek-V3-0324, Phi-4, and Mistral-Small-2503. Each model is tested under three prompting conditions to assess the robustness of \sysname structural decomposition prompting. The first condition employs a baseline prompt that simply instructs the model to construct a table from unstructured text. The second condition uses Chain-of-Thought(CoT) prompting \cite{wei2022chain}, introducing intermediate reasoning by appending “let’s think step by step” to the instruction. The third condition applies our proposed structural decomposition prompting strategy, which explicitly segments the task into logical subcomponents to guide model behavior.

\par In addition to these prompting variants, we compare against two baselines designed for structured table reconstruction. The first is Tabularis Revilio \cite{mukul}, a neurosymbolic pipeline developed to convert flattened tabular text into accurate table representations. 
The second baseline is SplitText, a method for table reconstruction that segments flattened tabular content into columns by leveraging linguistic and statistical cues~\cite{raza-17,raza2020web}. 


\par While \sysname may seem related to prior models such as \cite{deng2024texttupletable, wu-etal-2022-text-table}, their input settings and goals differ substantially from ours. 
\sysname assumes that the input is a semistructured textual representation of some table.
In contrast, the input for these other systems is a narrative-style paragraph description that contains some information that can be extracted as a table. 
Since the input type is different,
we do not include these models as baselines.

\subsection{Evaluation Metrics}
We adopt both structural and semantic metrics to comprehensively evaluate the quality of the generated tables.
\textbf{Coverage} is defined as the fraction of alphanumeric characters from the input that are preserved in the output table: \(
\text{\textbf{Coverage}} = 1 - \frac{\lvert U \rvert}{\lvert S \rvert}
\) where $\lvert S \rvert$ is the number of alphanumeric characters in the source input, and $\lvert U \rvert$ is the number of uncovered characters based on string matching.
\textbf{Hallucination Rate} is defined as the average fraction of unmatched content across all cells in the generated table. For a table with $R$ rows, each containing $C_i$ cells, it is computed as: \(
\text{\textbf{HallucinationRate}} = \frac{1}{R} \sum_{i=1}^{R} \left( \frac{1}{C_i} \sum_{j=1}^{C_i}(1 - \text{cellCoverage}_{ij}) \right)
\), where $cellCoverage_{ij} \in [0,1]$ denotes the degree to which cell(i,j) matches the input source(including partial matches).
\textbf{Exact Match(EM)} is a binary indicator that evaluates to 1 if the predicted table matches the ground-truth table exactly in both structure and content, and 0 otherwise. \textbf{Tree Edit Distance (TED)} is computed as the normalized cell-level edit distance between the predicted and ground-truth tables \cite{ted}. \textbf{Column Match (Col.V.M.)} is defined as the average percentage of columns that are exactly matched between the predicted and reference tables, micro-averaged per table and then averaged across all tables. \textbf{Value Match (C.V.M.)} reports the average percentage of individual cell values that are correctly reconstructed across all tables.

\subsection{User Study}
\par We conducted a within-subject empirical study with 21 spreadsheet users in which each participant evaluated the accuracy of and corrected errors in tables generated by both \sysname (Structural Decomposition), hereafter referred to as \userstudysys, and those generated by \sysname (COT), hereafter referred to as \userstudybaseline. This design allowed us to assess the practical strengths and limitations of these different approaches to table extraction and to identify which types of errors users found most difficult to fix and verify.
\par\textbf{Tasks:} Each participant worked on 2–4 tables from the FinReCon20 dataset\footnote{\url{https://anonymous.4open.science/r/FinRecon20-Dataset-1B11/}}, using Microsoft Excel to assess and fix outputs produced by both \userstudysys and \userstudybaseline in a within-subject design. To control for individual biases, each table was independently reviewed by three different participants. The order of tool outputs was randomized to mitigate ordering effects.
\par In cases where table fixes involved repetitive or tedious edits (e.g., adding multiple rows of similar data), participants were instructed to perform only a representative sample of the changes. This approach helped us gauge the perceived difficulty and effort without requiring exhaustive correction. 
After completing each task, participants filled out a questionnaire addressing (1) perceived accuracy of the table, (2) effort required to verify its correctness, and (3) effort required to make corrections.
\par{\textbf{Participants:}} We recruited 21 active spreadsheet users through an online user study recruitment platform. To ensure sufficient proficiency, we only included participants who reported using Microsoft Excel on at least 8 or more days in the past month. Given the financial nature of our dataset, we also required participants to have some experience working with financial tables in Excel (all participants reported working with financial spreadsheets at least occasionally). However, we did not restrict participation to individuals in finance-related occupations, allowing us to recruit a diverse pool of users across domains. This ensured a realistic mix of spreadsheet users while maintaining a baseline familiarity with financial tabular data.
\par{\textbf{Study Protocol}: All study sessions were conducted by one of the authors. At the beginning of each session, the administrator briefed participants on the task they would be performing, followed by demonstrating a tutorial using an example table. In pilot studies, we observed that participants would often focus on the formatting and visual aspects (e.g., font and spacing) and the robustness (e.g., number of formulas) of the replications when evaluating and fixing them. Since neither \userstudysys nor \userstudybaseline currently implement such aspects, the study administrator asked participants to focus solely on ensuring that the table replicates the structure and data of the source table. Participants then worked on each generation sequentially, answering the questionnaire after each table they worked on. In case they missed an error that was crucial to the correctness of the table while fixing the table (e.g., a missing row), the study administrator hinted at this issue to the participant. Participants completed working on generations by both \userstudysys and \userstudybaseline for each table they worked on before moving on to the next table.
}
\par{\textbf{Data Collection:}} We collected both quantitative (questionnaire responses) and qualitative (video and audio recordings of study sessions) data from our study. To gain a deeper understanding of the rationale behind participants' responses, we qualitatively coded the actions they performed while fixing the regenerated tables based on the study recordings. Table \ref{tab:annotation_labels} displays the list of such actions. If many actions were performed repetitively, only one such action was coded (e.g., if participants had to add spaces between words across all row headers of the table, this action was only coded once).

\par{\textbf{Limitations:}} Our study only uses spreadsheets related to financial data, so our results might not generalize to other domains. Additionally, while we aimed to recruit a diverse set of participants, we use a relatively small sample size ($N = 21$), which may not be representative of all spreadsheet users. Future work could explore the usefulness of \sysname in other contexts and with larger sample sizes.
Further, since participants were unfamiliar with the data in the source images and did not have a personal stake in ensuring that the replication was accurate to the source data, this may have influenced their actions and responses. Finally, since both tools in the study were developed by our research team, this may lead to confirmation biases in the analysis. To mitigate this bias, we do not analyse absolute values and only report comparative results. 

\begin{table}[t]  
\centering\scriptsize  
\setlength{\tabcolsep}{4pt}  
\begin{tabular}{@{}lp{7cm}@{}}  
\toprule 
\ \textbf{Code} & \textbf{Description} \\  
\midrule
\multicolumn{2}{c}{\textbf{Action Type}} \\  
\midrule  
Add & Adding values to the replication that were present in the source table but missing from the replication \\  
Modify & Fixing incomplete or incorrect values within individual cells in the replication \\
Delete & Removing values from the replication that were not present in the source table \\ 
    Realign & Shifting cells that were incorrectly aligned in the replication \\
\midrule  
\multicolumn{2}{c}{\textbf{Data Type}} \\  
\midrule  
Column Heading & Heading for a column \\  
Row Heading & Heading for a row \\  
Subheading & Heading not for a particular row or column but for a subsection of the spreadsheet \\   
Note & Values that are not associated with a particular row or column, but describe the data in the spreadsheet (e.g., 'values in thousands') \\  
Data & All non-header, non-note values\\
\bottomrule  
\end{tabular}  
\caption{Actions performed by user study participants to fix table regenerations. Each action consisted of an action type as well as the data type upon which the action was performed}  
\label{tab:annotation_labels}  
\end{table}


%% file: appendix/app_results.tex
\section{Results and Analysis}
\label{sec:app_results}
\begin{table*}[!t]
\centering
\begin{tabular}{llcccccc}
\toprule
\textbf{Prompting Strategy} & \textbf{Model} & \textbf{Cov.} & \textbf{Hall.} & \textbf{TED} & \textbf{E.M.} & \textbf{C.V.M.} & \textbf{Col.V.M.} \\
\midrule
\multirow{6}{*}{\textsc{Base Prompt}} 

& GPT-4 & 0.87 & 0.13 & 0.53 & 28.30 & 53.98 & 45.89\\
& DeepSeek-R1 & 0.77 & 0.18 & 0.41 & 19.80 & 47.05 & 39.42\\
& O3-mini & 0.78 & 0.14 & 0.42 & 3.0 & 50.12 & 45.95\\
& DEEPSEEK-V3-0324 & 0.83 & 0.13 & 0.59	& 31.60 & 58.85 & 50.75\\
& Phi-4 & 0.82 & 0.16 & 0.41 & 15.90 & 45.00 & 34.82\\
& Mistral-Small & 0.87 & 0.12 & 0.50 & 13.80 & 49.95 & 39.08\\
\midrule
\multirow{4}{*}{\textsc{Chain-of-Thought(CoT)}} 
  & GPT-4 & 0.87 & 0.13 & 0.52 & 27.30 & 53.41 & 45.09 \\
  & DEEPSEEK-V3-0324 & 0.89 & 0.12 & 0.59 & 32.00 & 58.71 & 51.11 \\
  & Phi-4 & 0.82 & 0.17 & 0.41 & 15.6 & 44.72 & 34.90\\
  & Mistral-Small & 0.86 & 0.13 & 0.46 & 11.3 & 48.37 & 37.97 \\
\midrule
\multirow{6}{*}{\textsc{Structural Decomposition}} 
  & GPT-4 & 0.94 & 0.11 & 0.65 & 36.80 & 64.75 & 56.90 \\
  & DeepSeek-R1 & 0.96 & 0.10 & 0.74 & 28.7 & 76.06 & 69.20 \\
  & O3-mini & 0.94 & 0.11 & 0.48 & 23.00 & 50.62 & 40.35 \\
  & DEEPSEEK-V3-0324 & 0.93 & 0.11 & 0.65 & 35.80 & 63.65 & 55.90 \\
  & Phi-4 & 0.88 & 0.14 & 0.45 & 13.80 & 47.18 & 36.83 \\
  & Mistral-Small & 0.92 & 0.11 & 0.60 & 29.7 & 60.02 & 49.61 \\
\bottomrule
\end{tabular}
\caption{WikiTables Dataset - Baseline comparison across prompting strategies and models. Metrics: Coverage, Hallucination Rate, Tree Edit Distance(TED), Exact Match(E.M., \%), Cell Value Match(C.V.M., \%), Column Value Match(Col.V.M., \%).}
\label{tab:wikipedia-metrics}
\end{table*}

\begin{table*}
\centering
\begin{tabular}{llcccccc}
\toprule
\textbf{Prompting Strategy} & \textbf{Model} & \textbf{Cov.} & \textbf{Hall.} & \textbf{TED} & \textbf{E.M.} & \textbf{C.V.M.} & \textbf{Col.V.M.} \\
\midrule
\multirow{6}{*}{\textsc{Base Prompt}} 
  & GPT-4 &  0.94 &  0.14 & 	0.07 &  45 &  64.12  & 54.96\\
  & DeepSeek-R1 & 0.92 & 0.17 & 0.13 &  47.3 & 65.69 & 56.87\\
  & O3-mini & 0.92 & 0.16 & 0.17 & 32.7 & 65.54 & 55.12\\
  & DEEPSEEK-V3-0324 & 0.96 & 0.17 & 0.07 & 42.4 & 68.93 & 58.08\\
  & Phi-4 & 0.88 & 0.17 & 0.07 & 34.8 & 55.87 & 46.15 \\
  & Mistral-Small & 0.90 & 0.14 & 0.06 & 35.4 & 60.39 & 51.59 \\
\midrule
\multirow{4}{*}{\textsc{Chain-of-Thought(CoT)}} 
& GPT-4 &  0.96 &  0.17 & 0.108 & 42.8 & 63.57 & 54.69\\
& DEEPSEEK-V3-0324 & 0.97 & 0.15 & 0.105 & 44 & 68.63 & 58.17\\
& Phi-4 & 0.88 & 0.2 & 0.04 &  36.6 & 53.43 & 46.85\\
& Mistral-Small & 0.88 & 0.17 & 0.04 & 33.8 & 58.25 & 50.96\\
\midrule
\multirow{6}{*}{\textsc{Structural Decomposition}} 
  & GPT-4 & 0.97 &	0.15 &	0.22 &	81.6 &	81.63 &	69.45\\
  & DeepSeek-R1 & 0.98 &	0.14 &	0.21 &	72.1 &	86.56 & 73.28\\
  & O3-mini & 0.97 &	0.14 &	0.33 &	68.6 &	64.87 & 54.58\\
  & DEEPSEEK-V3-0324 & 0.97 & 0.14 &	0.17 &	76.8 &	76.56 &	64.34 \\
  & Phi-4 & 0.9 &	0.2 &	0.07 &	63.9 &	60.59 & 51.49 \\
  & Mistral-Small & 0.91 &	0.16 &	0.22 &	60.2 &	70.84 & 59.05 \\
\bottomrule
\end{tabular}
\caption{BrokenCSV Dataset - Baseline comparison across prompting strategies and models. Metrics: Coverage, Hallucination Rate, Tree Edit Distance(TED), Exact Match(E.M., \%), Cell Value Match(C.V.M., \%), Column Value Match(Col.V.M., \%).}
\label{tab:brokencsv-metrics}
\end{table*}

\begin{table*}
\centering
\begin{tabular}{llcccccc}
\toprule
\textbf{Prompting Strategy} & \textbf{Model} & \textbf{Cov.} & \textbf{Hall.} & \textbf{TED} & \textbf{E.M.} & \textbf{C.V.M.} & \textbf{Col.V.M.} \\
\midrule
\multirow{6}{*}{\textsc{Base Prompt}} 
  & GPT-4 & 0.91	& 0.14  & 0.45  &30 &  48.18   & 40.22  \\
  & DeepSeek-R1 & 0.89 &	0.17  &	0.51 & 30.60 &  55.54 & 44.44  \\
  & O3-mini & 0.89 & 0.16 & 0.51 & 22.00 &  55.19  & 46.57  \\
  & DEEPSEEK-V3-0324 & 0.92 & 0.16 & 0.42 & 27.40 & 49.68 & 37.88 \\
 & Phi-4 &	0.85 &	0.2	& 0.36  & 25.80 & 43.95 & 34.38\\
 & Mistral-Small &	0.88 &	0.17 &	0.428 & 27.40 & 46.15 &	36.88\\
\midrule
\multirow{4}{*}{\textsc{Chain-of-Thought(CoT)}} 
& GPT-4 &	0.91 &	0.15 &	0.48 &	29.80& 47.38  &	38.38 \\
&DEEPSEEK-V3-0324 &	0.92 &	0.149 &	0.50 &	32.00 &	54.46  &	43.16 \\
& Phi-4 &	0.84 &	0.19 &	0.34 & 22.60 & 40.46  & 32.49 \\
& Mistral-Small &	0.84 &	0.16 & 0.32 &	20.00 & 35.73  & 28.53 \\
\midrule
\multirow{6}{*}{\textsc{Structural Decomposition}} 

& GPT-4	& 0.93	& 0.14	& 0.67	&  61.60 &	71.76 &	75.41\\
& DeepSeek-R1	& 0.96	& 0.13	& 0.65& 50.20 &	64.51 & 57.98\\
& O3-mini	& 0.96	& 0.12	& 0.79& 41.00 &	53.97 & 49.53\\
& DEEPSEEK-V3-0324	& 0.93	& 0.14	& 0.64& 52.00 &	65.57 &	59.93\\
& Phi-4	& 0.88	& 0.18	& 0.43& 	42.00 &	50.26 & 46.45\\
& Mistral-Small	& 0.92	& 0.14	& 0.68&  40.20 & 57.24 & 49.52\\

\bottomrule
\end{tabular}
\caption{PubtabNet Dataset - Baseline comparison across prompting strategies and models. Metrics: Coverage, Hallucination Rate, Tree Edit Distance(TED), Exact Match(E.M., \%), Cell Value Match(C.V.M., \%), Column Value Match(Col.V.M., \%).}
\label{tab:pubtabnet-metrics}
\end{table*}

\begin{table*}
\centering
\begin{tabular}{llcccccc}
\toprule
\textbf{Prompting Strategy} & \textbf{Model} & \textbf{Cov.} & \textbf{Hall.} & \textbf{TED} & \textbf{E.M.} & \textbf{C.V.M.} & \textbf{Col.V.M.} \\
\midrule
\multirow{6}{*}{\textsc{Base Prompt}} 
&GPT-4	& 0.89& 	0.19 &	0.13 & 8.9 & 22.47	& 14.45\\
&DeepSeek-R1	& 0.84	& 0.19 &	0.28 &  12.4 &	33.17 & 25.31\\
&O3-mini	& 0.90& 	0.14 &	0.36 & 2.2 &	40.76	& 33.09\\
&DeepSeek-V3-0324	& 0.91& 	0.18 &	0.16 & 7.7 &	27.34	& 13.81\\
&Phi-4	& 0.84& 	0.21 &	0.013 & 5.7 &	19.04	& 10.01\\
&Mistral-Small	& 0.87	& 0.18 &	0.13 & 8.2 &	22.91	& 14.66\\
\midrule
\multirow{4}{*}{\textsc{Chain-of-Thought(CoT)}} 
  & GPT-4	& 0.89	& 0.17 & 0.226	& 10.3 & 24.02 &16.72\\
 & DEEPSEEK-V3-0324	& 0.91	& 0.16 & 0.21	& 11.6 & 29.16 & 19.27\\
& Phi-4	& 0.80	& 0.21 & 0.077 & 6.5 & 18.80 & 10.27 \\
& Mistral-Small	& 0.81	& 0.17 & 0.081 & 4.5 & 16.75 & 10.82\\
\midrule
\multirow{6}{*}{\textsc{Structural Decomposition}}
& GPT-4 & 0.91	& 0.15 & 	0.47 & 21.7	& 33.36 & 29.77\\
& DeepSeek-R1 & 0.94 & 0.14	& 0.44 & 10.1 & 38.84 & 32.17\\
& O3-mini & 0.93 & 0.15 & 0.70 & 14.4 & 27.17 & 25.10\\
& DEEPSEEK-V3-0324 & 0.93	& 0.14	& 0.36 & 21.9$\pm$0.026 &	31.69 & 28.93\\
& Phi-4 & 0.82	& 0.21	& 0.15 & 8.9 & 18.92 & 10.75\\
& Mistral-Small & 0.90 & 0.16	& 0.45 & 20 & 34.01	& 28.69\\
\bottomrule
\end{tabular}
\caption{FinTabNet Dataset - Baseline comparison across prompting strategies and models. Metrics: Coverage, Hallucination Rate, Tree Edit Distance(TED), Exact Match(E.M., \%), Cell Value Match(C.V.M., \%), Column Value Match(Col.V.M., \%).}
\label{tab:fintabnet-metrics}
\end{table*}

\begin{figure*}
    \begin{subfigure}{\textwidth}
    \centering
        \includegraphics[width=0.75\textwidth]{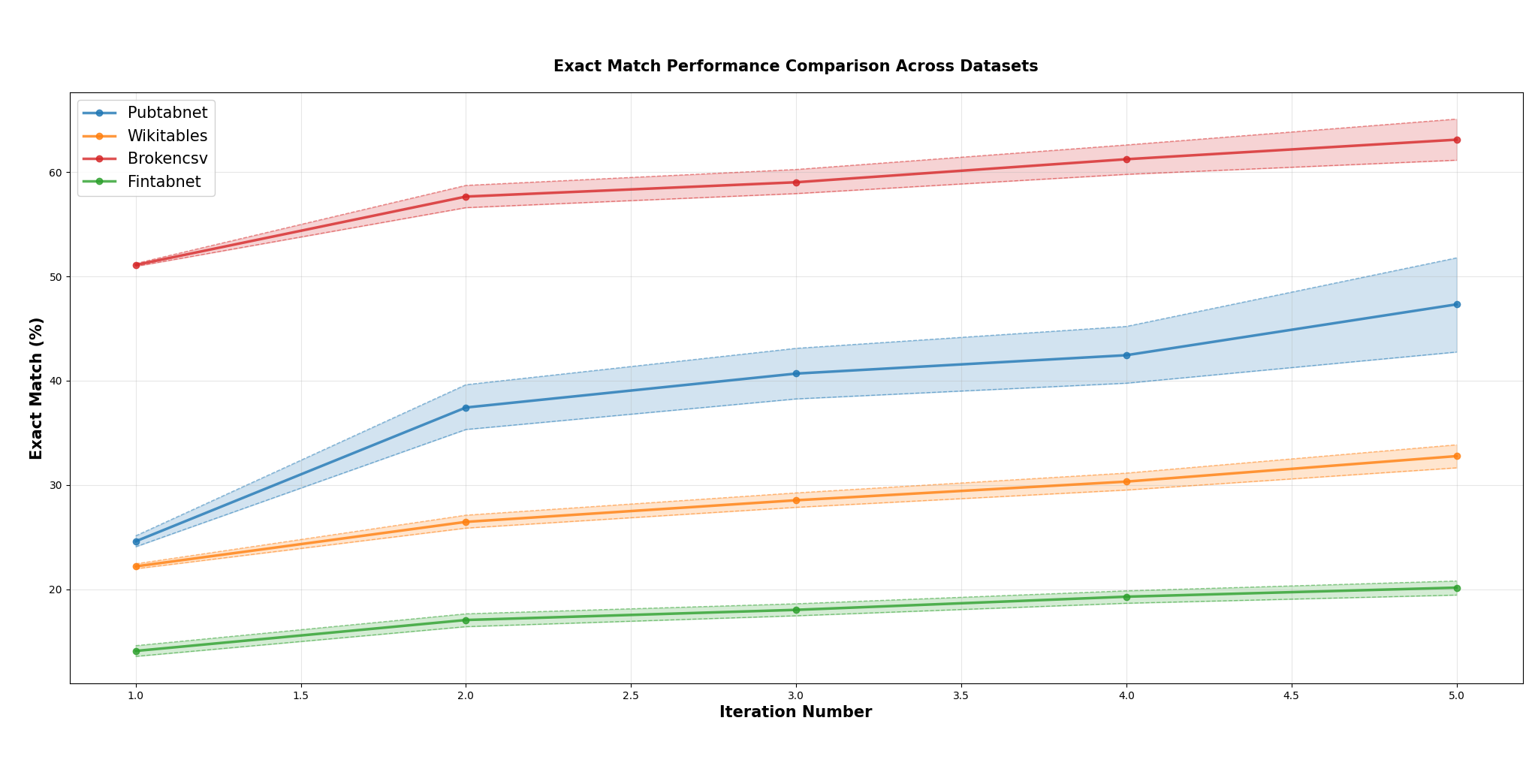}
        \caption{Exact Match}
    \end{subfigure}
    \begin{subfigure}{\textwidth}
    \centering
        \includegraphics[width=0.75\textwidth]{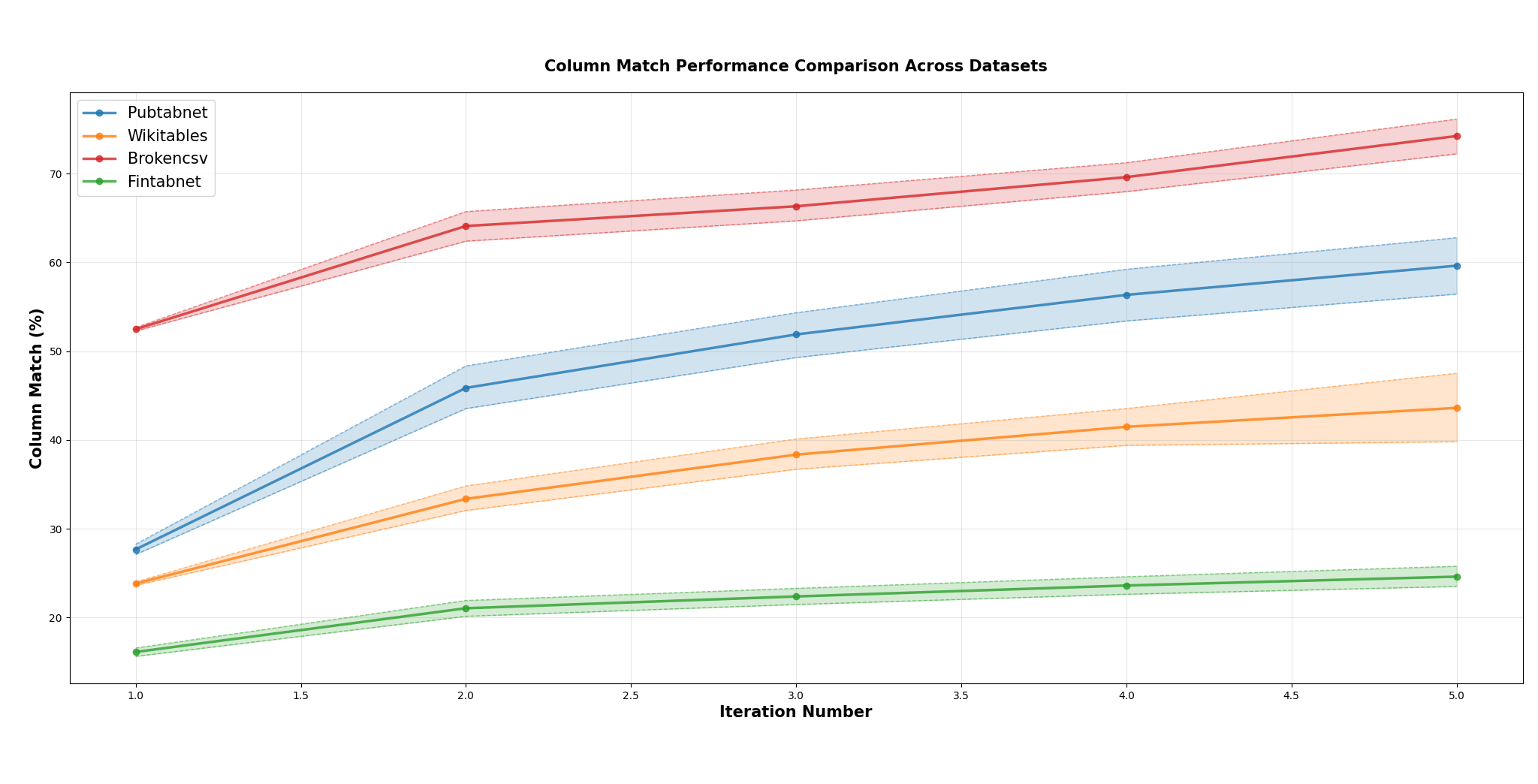}
        \caption{Column Match}
    \end{subfigure}
    \begin{subfigure}{\textwidth}
    \centering
        \includegraphics[width=0.75\textwidth]{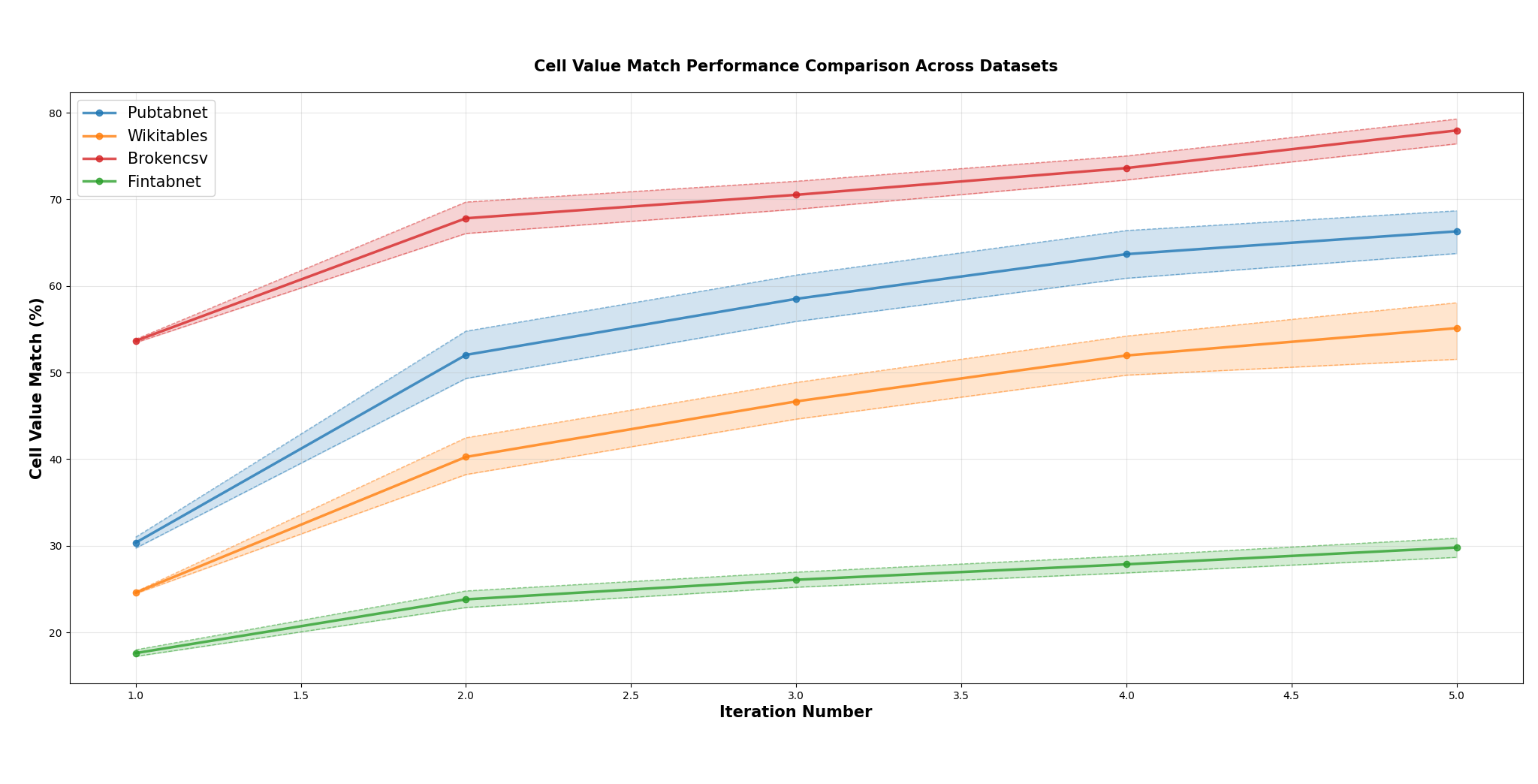}
        \caption{Cell Value Match}
    \end{subfigure}
    \caption{Convergence behavior of the iterative table reconstruction system in \sysname across PubTabNet, WikiTables, BrokenCSV, and FinTabNet. Most gains in Exact Match, Column Match, and Cell Value Match metrics are observed in the first 2–3 iterations, with BrokenCSV and PubTabNet achieving higher final accuracies and FinTabNet saturating early.}
    \label{fig:metric_convergence}
\end{figure*}

\textbf{RQ1}: \textit{To what degree do prompting strategies and model choice influence table-extraction performance when evaluated on datasets of escalating complexity?}\\
\textbf{``Structure first'' beats ``reason more''}: We evaluated the Structural Decomposition prompt (prioritizing schema/layout followed by value filling) across four datasets and six models, comparing it against a baseline prompt and Chain-of-Thought (CoT) prompting. Results (Figure \ref{fig:em_heatmap} demonstrate that the Structural Decomposition approach consistently outperforms both baseline and CoT methods, with improvements of 10–40 EM points (e.g., PubTabNet and BrokenCSV datasets) and 8–15 points in ColVM/CVM. While CoT offers marginal gains over the baseline, the structure-first strategy yields substantially greater enhancements in table extraction accuracy. \\
\textbf{Model Capacity Matters}:
Model capacity plays a clear role in determining table extraction accuracy, with larger models such as GPT-4 and DeepSeek-V3-0324 consistently outperforming smaller models like o3-mini across all prompting strategies.Our results also reveal that Structural Decomposition substantially narrows this performance gap. Under the Base Prompt or COT strategies, smaller models often struggle with layout understanding, yielding low EM scores,for instance, o3-mini achieves only 2.2\% EM on FinTabNet using the Base Prompt. With SD, however, o3-mini’s EM rises to 14.4\%, while GPT-4 improves from 8.9\% to 21.7\% (2.4×), indicating that SD disproportionately benefits lower-capacity models. This trend is consistent across datasets, with small models gaining 15–25 percentage points in EM under SD, effectively reducing their reliance on internal structural reasoning. Nevertheless, higher-capacity models still maintain an advantage, particularly in Cell Value Match (CVM) and Column Match (ColVM) accuracy, and in handling large or complex tables.  \\
\textbf{Table Size Effects and Efficiency Implications:}
Table extraction performance is strongly influenced by the size of the table, regardless of the prompting strategy used. As shown in Figure~\ref{fig:size_em_analysis}, EM scores decline noticeably as the number of rows increases, indicating that deeper tables pose greater challenges for structural reconstruction. Similarly, performance peaks at moderate column widths (approximately 6–8 columns) and deteriorates for wider tables, likely due to increased ambiguity in column alignment and value placement. While Structural Decomposition helps mitigate these effects, larger models exhibit a more gradual degradation in accuracy as table size grows, whereas smaller models tend to plateau early, particularly on wide or deep tables. \\
\textbf{Dataset Complexity and Performance Characteristics}:
Our evaluation reveals interesting relationship between dataset complexity metrics and LLM performance. BrokenCSV demonstrates the highest performance across all models (61.6-76.8 range), followed by PubTabNet (40.2-61.6 range), while FinTabNet exhibits consistently poor performance (8.9-21.9 range) and WikiTables shows moderate but variable results (13.8-36.8 range). This performance hierarchy contradicts traditional complexity assessments, where FinTabNet's sophisticated financial structures and WikiTables' diverse domain coverage would typically be considered more challenging than BrokenCSV's intentionally malformed data. The observed pattern suggests that semantic complexity—characterized by contextual richness and self-descriptive content, is more predictive of LLM performance than structural complexity. BrokenCSV's superior performance can be attributed to its design philosophy of creating self-contained, descriptive entries that remain interpretable despite formatting inconsistencies. Similarly, PubTabNet's scientific tables contain extensive contextual information, including measurement units, methodological descriptions, and domain-specific terminology that facilitate content interpretation without relying on structural cues. In contrast, FinTabNet's financial tables depend heavily on precise positional relationships and abbreviated notation that lose semantic meaning when delimiters are removed, while WikiTables' encyclopedic format often assumes structural organization for disambiguation.\\
\textbf{Iterative Feedback Loop Performance Analysis}: To evaluate the effectiveness of our iterative feedback loop, we analyzed performance trajectories across multiple iterations of Structural Decomposition. As shown in Figure \ref{fig:metric_convergence}, the largest gains in EM, C.V.M., and Col.V.M. occur within the first two to three iterations, with diminishing returns thereafter. Specifically, EM typically increases by 10–20 percentage points between iterations 1 and 2, and by iteration 3 or 4, most tables reach a performance plateau. By iteration 5, improvements are minimal.\\
\begin{figure*}
    \centering
    \includegraphics[width=\linewidth]{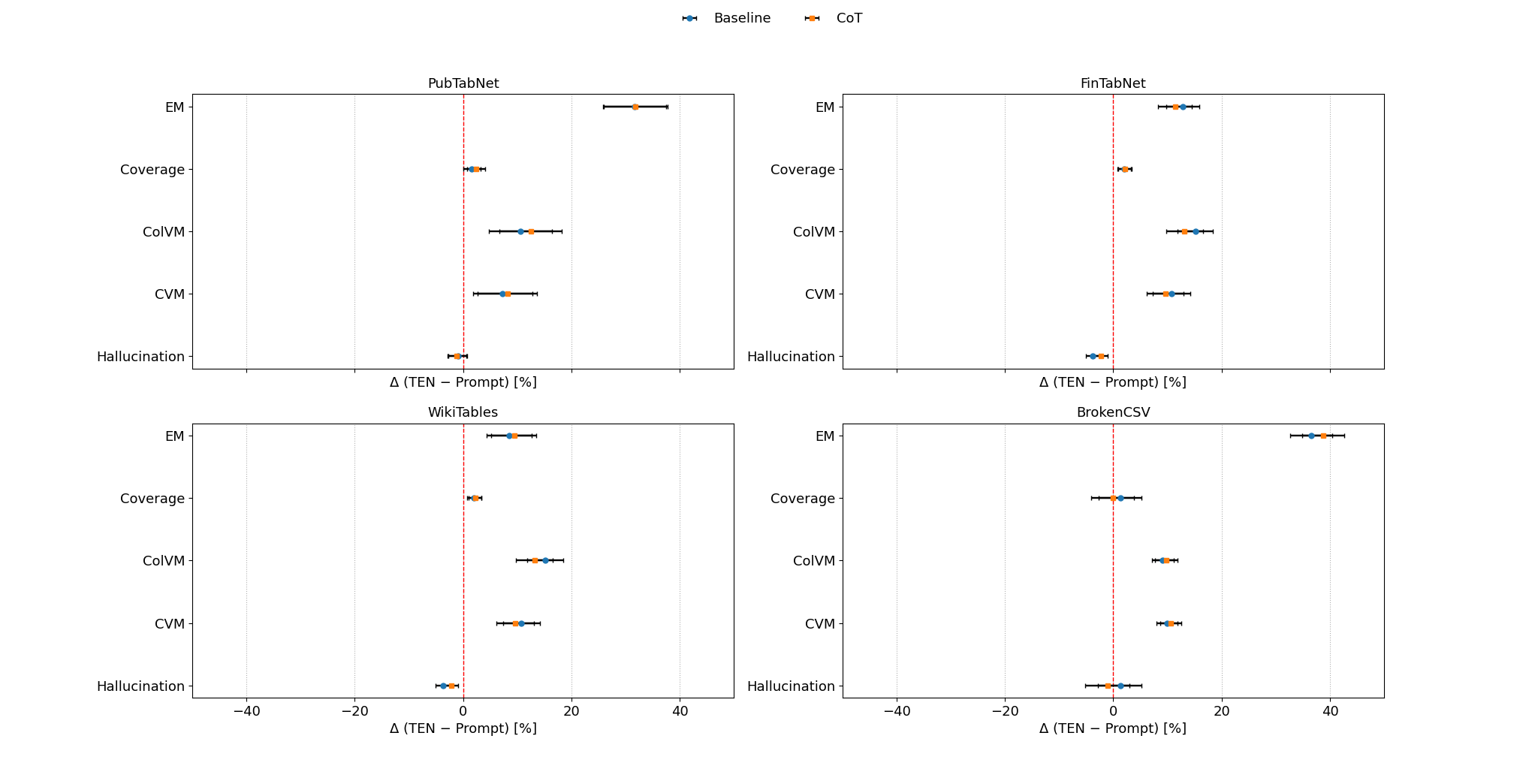}
    \caption{ Absolute delta ($\Delta$) in performance metrics between Structural Decomposition and each prompting strategy (Baseline, CoT), reported per dataset. Horizontal error bars indicate 95\% bootstrap confidence intervals over 10,000 paired samples. Positive values indicate higher performance by \sysname. \sysname consistently improves Exact Match (EM), Column Value Match (ColVM), and Cell Value Match (CVM) across all datasets. Differences in Hallucination and Coverage vary across datasets. Significance is assessed separately in Figure.}
    \label{fig:statistical_result_bootstrapped}
\end{figure*}

\begin{figure*}[t]
    \centering
    \begin{minipage}{0.48\textwidth}
        \centering
        \includegraphics[width=\linewidth]{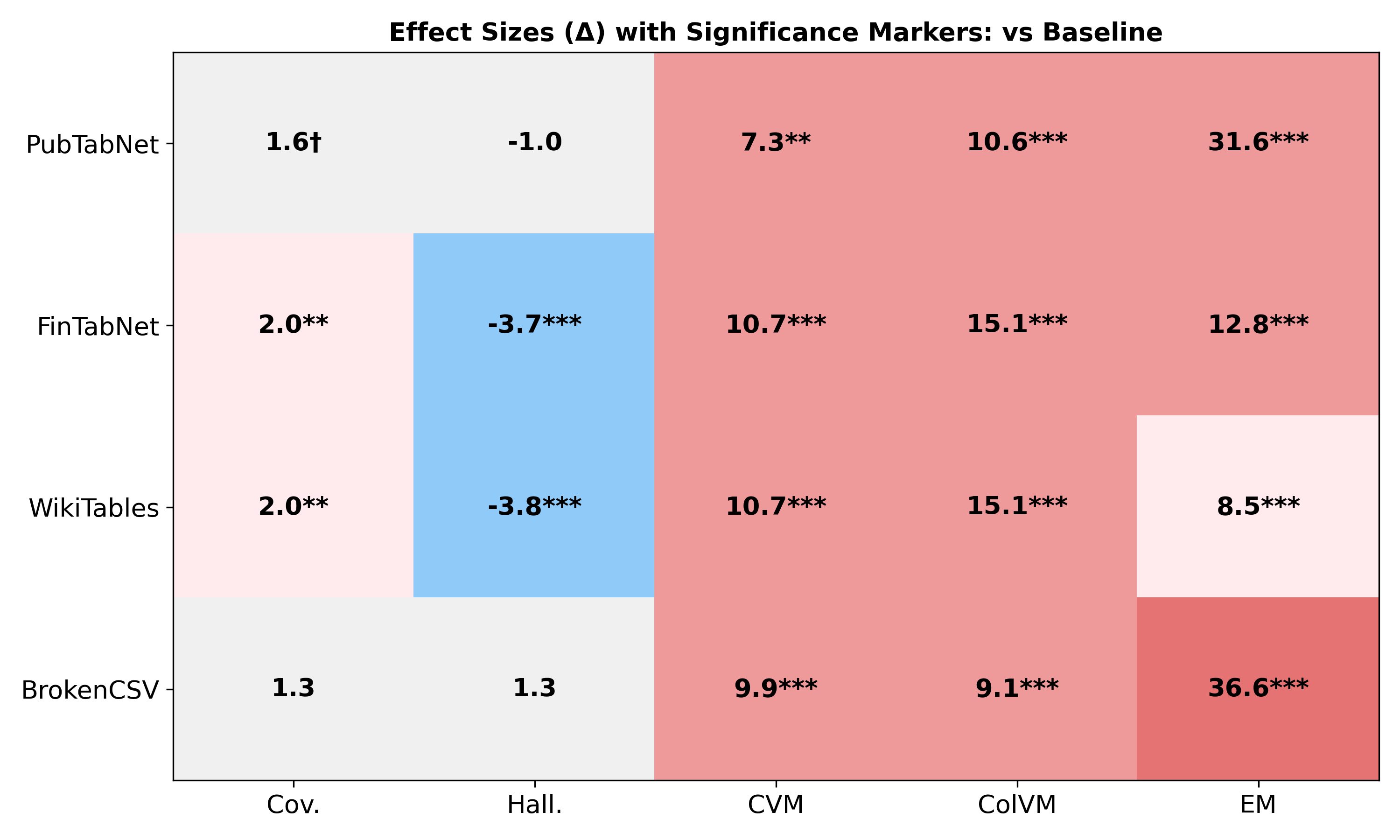}
        \caption{Heatmap showing effect sizes ($\Delta$) and statistical significance for performance differences between Structural Decomposition and the Baseline strategy. Each cell reports the average delta ($\Delta = \text{Structural Decomposition} - \text{Baseline}$) for a given dataset and metric. Color encodes both the magnitude and direction of the effect: blue shades indicate negative effects (TEN outperforms), red shades indicate positive effects (TEN underperforms), and gray indicates non-significant results. Effect size is measured using Cohen's \(d\); significance is marked as: \textbf{***} ($p < 0.001$), \textbf{**} ($p < 0.01$), \textsuperscript{\textdagger} ($p < 0.1$). For EM, Coverage, ColVM, and CVM, higher values are better; for Hallucination, lower is better.}

        \label{fig:heatmap_baseline}
    \end{minipage}\hfill
    \begin{minipage}{0.48\textwidth}
        \centering
        \includegraphics[width=\linewidth]{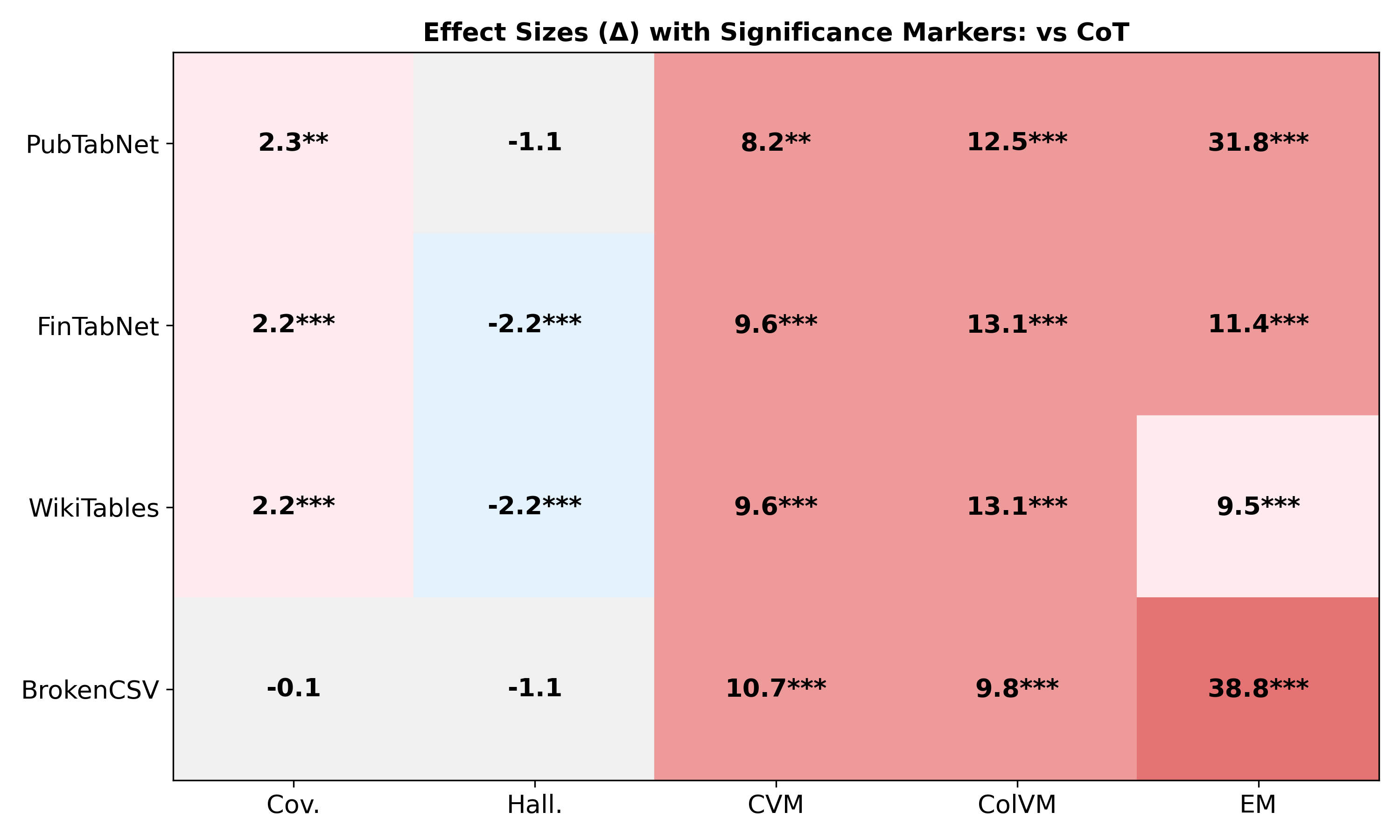}
        \caption{Heatmap showing effect sizes ($\Delta$) and statistical significance for performance differences between Structural Decomposition and the Chain-of-Thought (CoT) prompting strategy. Each cell represents the average delta ($\Delta = \text{Structural Decomposition} - \text{Baseline}$) across datasets and metrics. Color encodes both the magnitude and direction of the effect: blue indicates negative effects (TEN outperforms CoT), red indicates positive effects (TEN underperforms), and gray marks non-significant results. Effect size is measured using Cohen's \(d\); significance is marked as: \textbf{***} ($p < 0.001$), \textbf{**} ($p < 0.01$), \textsuperscript{\textdagger} ($p < 0.1$). For EM, Coverage, ColVM, and CVM, higher values are better; for Hallucination, lower is better.}
        \label{fig:heatmap_cot}
    \end{minipage}
\end{figure*}
\textbf{Robustness \& Statistical Significance}: To assess the reliability of our observed improvements, we conducted  statistical analyses across all datasets and prompting strategies. We used paired bootstrap resampling (B = 10,000 samples) to compute 95\% confidence intervals for the difference ($\delta$) in performance metrics, between Structural Decomposition and each baseline strategy. As shown in Figure \ref{fig:statistical_result_bootstrapped}, the confidence intervals for EM, CVM, and ColVM are consistently positive and do not overlap zero, indicating statistically significant gains for Structural Decomposition across all datasets. In addition, we computed Cohen’s d effect sizes to quantify the magnitude of these improvements and tested their statistical significance (Figures \ref{fig:heatmap_baseline} and \ref{fig:heatmap_cot}). Most effect sizes for EM, CVM, and ColVM are large and significant ($p < 0.001$), while hallucination rates show minor and dataset-dependent differences.

\medskip\noindent
\textbf{RQ2}: How accurately does \sysname reconstruct tables compared to existing baselines?\\

\ignore{ 

\begin{table}[t]
\centering
\scriptsize
\setlength{\tabcolsep}{4pt}
\begin{tabular}{@{}llcccc@{}}
\toprule
\textbf{Method} & \textbf{Metric} & \textbf{PubTabNet} & \textbf{FinTabNet} & \textbf{WikiTables} & \textbf{BrokenCSV} \\
\midrule
\multirow{4}{*}{\sysname(GPT-4o)} 
& TED      & 0.67 & 0.47 & 0.65 & 0.22  \\
& EM        & 61.60 & 21.7 & 36.80 & 81.6  \\
& C.V.M.    & 71 & 33 & 64 & 81  \\
& Col.V.M. & 75 & 29 & 56 & 69  \\

\midrule
\multirow{4}{*}{Revilio} 
& TED       & 0.60 & 0.53 & 0.59 & 0.17 \\
& EM        & 52.34 & 15.65 & 33.43 & 60.96 \\
& C.V.M.    & 66 & 29 & 59 & 76 \\
& Col.V.M.  & 60 & 23 & 51 & 63 \\
\midrule
\multirow{4}{*}{SplitText} 
& TED       & 0.55 & 0.58 & -- & 0.43 \\
& EM        & 26 & 11 & 1.1 & 53.8 \\
& C.V.M.    & 37 & 22 & 1 & 46 \\
& Col.V.M.  & 35 & 19 & 8 & 40 \\
\bottomrule
\end{tabular}
\caption{Comparison of \sysname with existing baselines on table reconstruction accuracy.}
\label{tab:baseline_comparison}
\end{table}
\endignore}

To address this research question, we compare \sysname with Revelio and SplitText across four benchmarks and results are shown in a Table in the main paper. 
\sysname consistently outperforms both Revilio and SplitText on content-fidelity metrics (EM, CVM, ColVM). On PubTabNet, \sysname improves EM by +9.26 pts over Revilio, together with higher CVM (+5 pt) and ColVM (+15 pt). On the more challenging FinTabNet, absolute EM is lower for all methods, but \sysname still yields a +6.05 pt EM gain and higher CVM/ColVM. On WikiTables, \sysname delivers smaller yet consistent improvements (+3.37 EM; +5 CVM; +5 ColVM). The largest gains appear on BrokenCSV, where \sysname attains 81.6 EM (vs. 60.96 for Revilio and 53.8 for SplitText), alongside higher CVM/ColVM.

Structural alignment, measured by TED, shows mixed results. \sysname achieves lower TED on FinTabNet and is substantially better than SplitText on BrokenCSV, but slightly worse than Revilio on PubTabNet and WikiTables. This pattern suggests that \sysname’s structure-first decoding can sometimes normalize header layouts or span handling in ways that improve content agreement (EM/CVM/ColVM) yet diverge modestly from the reference tree, reflecting granularity differences rather than content errors. 

TED varies across datasets because it rewards structural isomorphism and penalizes benign layout normalization. It compares systems to the reference representation rather than semantic correctness. When \sysname consolidates multi-row headers, propagates or flattens spans, removes empty scaffold rows/columns, or relocates footnotes and units, the values are preserved but the parse tree changes. Consequently, TED can worsen even as EM, CVM, and ColVM improve, so it should be interpreted alongside content-fidelity metrics.

\medskip\noindent
\textbf{RQ3}: How useful is \sysname at reconstructing tables from real-world financial documents?\\
To answer this research question, we analysed participant's actions and questionnaire responses in our user study. We compared these actions across tables generated by \sysname (Structural Decomposition), hereafter referred to as \userstudysys, and those generated by \sysname (COT), hereafter referred to as \userstudybaseline.

\begin{figure}
    \centering
    \includegraphics[width=1\linewidth]{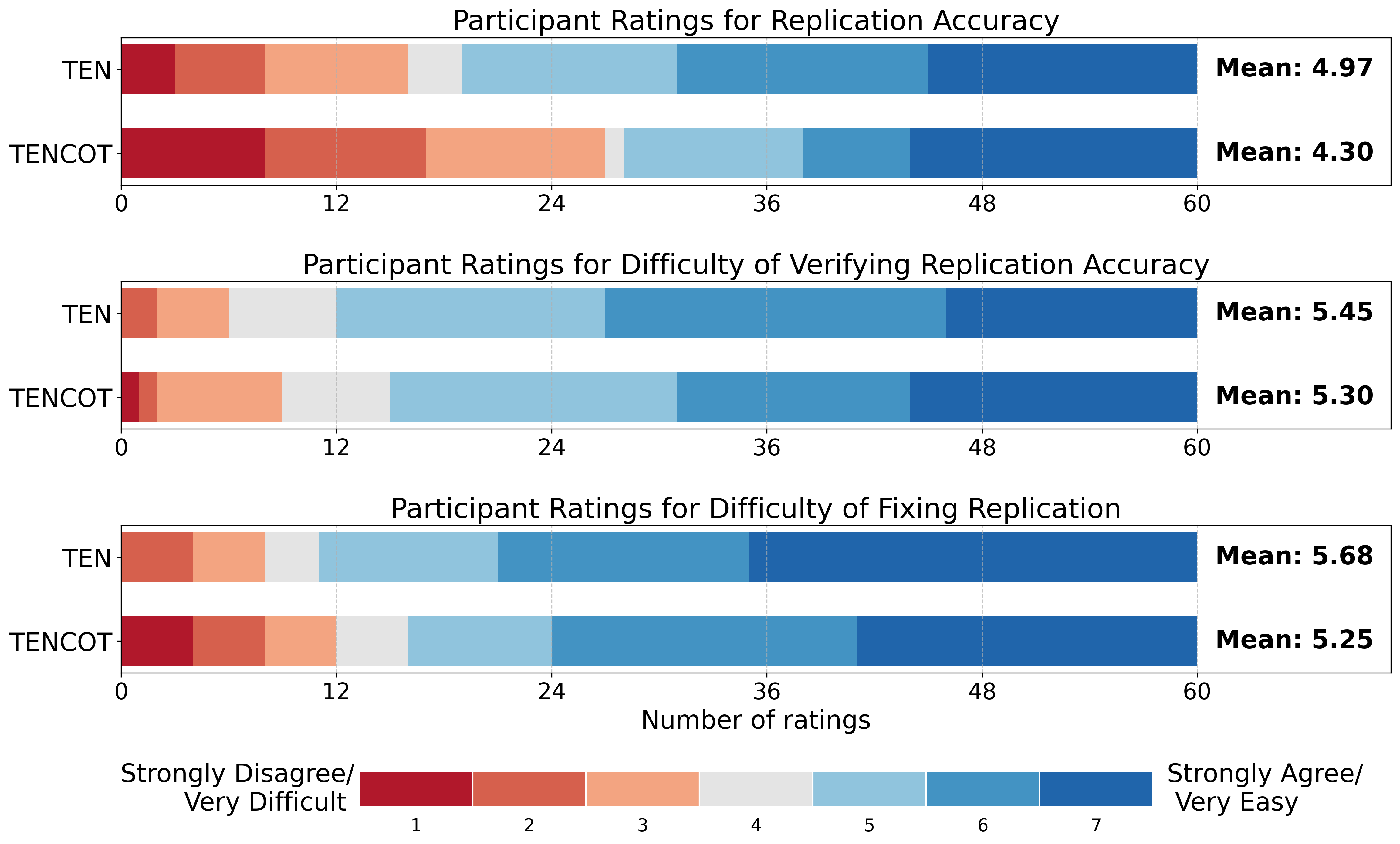}
    \caption{Questionnaire responses in user study}
    \label{fig:annotations}
\end{figure}

\par \textbf{Perceived Accuracy: }
Figure \ref{fig:annotations} summarizes participants' responses to the different questions in the questionnaire provided to them. To statistically test the differences in their responses across the two conditions (\userstudysys and \userstudybaseline), we apply the Wilcoxon signed-rank test, as our data is paired (each participant provided one response for each condition) and ordinal (Likert scale). Further, since this test is non-parametric, it is suited to smaller sample size such as in our study ($N = 20*3 = 60$).
We found that on the Likert scale from 1-7, participants reported that the accuracy of tables generated by \userstudysys (mean = 4.97) was significantly higher than of those generated by \userstudybaseline (mean = 4.30) (Wilcoxon signed-rank test, $z = 284.0, p = .02$) with a moderate effect size ($r = -0.30)$. 
Further, in 7 out the 20 tables used for the study, all three reviewers independently rated \userstudysys to be more accurate than \userstudybaseline, while all participants preferred \userstudybaseline for only 1 table. 

Despite this difference in responses on the tables' accuracy, we did not observe any significant differences in participants' responses on the effort required to verify ($z = 260.0, p = .51$) or fix ($z = 266.0, p = .08$) these tables. Out of the 43 responses in which participants reported a difference in accuracy between tables generated by the two tools, participants did not report a corresponding difference in the effort required to fix these tables in 11 cases. Further, in the 17 responses in which participants did not report any difference in accuracy between the two tools, 
participants reported a difference in the effort required to fix the generations by each tool in 5 cases. This suggests that accuracy alone does not fully reflect user burden and may not always be a sufficient indicator for the usefulness of the replication -- the effort required to locate and fix different types of errors in these replications might play an important role in understanding the overall usability of the replication.

\begin{table*}  
    \centering\scriptsize  
    \setlength{\tabcolsep}{3pt}  
    \begin{tabular}{@{}lcccccc@{}}  
    \toprule  
    \textbf{Action Type} & \textbf{Avg. Difficulty With Action} & \textbf{Avg. Difficulty Without Action} & \textbf{N} & \textbf{$z$} & \textbf{$p$} & \textbf{$r$ (effect size)} \\  
    \midrule  
    Add & 4.71 & 5.66 & 35 & 64.0 & 0.013 & -0.42 \\  
    Remove & 5.50 & 5.14 & 22 & 68.5 & 0.452 & -0.16 \\  
    Whitespace & 4.67 & 6.22 & 9 & 0.0 & 0.026 & -0.74 \\  
    Modify & 4.36 & 5.68 & 25 & 25.0 & 0.008 & -0.53 \\  
    Manual Misaligned & 4.96 & 5.57 & 23 & 33.0 & 0.115 & -0.33 \\  
    \bottomrule  
    \end{tabular}  
    \caption{Wilcoxon signed-rank test results for fix difficulty ratings with/without performing  specific action types. $N$ represents the number of samples used for the particular test. $r$ below }  
    \label{tab:fix_rating_stats}  
\end{table*}  

\par \textbf{Participant Actions:} Based on the codes in Table \ref{tab:annotation_labels}, we coded a total of 385 actions performed by participants (179 with tables generated by \sysname vs 206 with tables generated by \userstudybaseline). To assess the effect of each type of action and type of data (Table \ref{tab:annotation_labels}) on the increase in effort by participants to verify and fix the table, we perform statistical tests on those samples for which participants performed the action for one tool and did not perform the action for the other tool.
Thus, for each test, one sample contains instances where participants performed the action on a particular table and one sample contains instances where participants did not perform the action for the same table (but a different tool). We do not control for which tool participants worked with in these samples to avoid any biases.
While creating such pairs leads to a reduction in our sample size, this ensures that our data is paired, and allows us to continue to use the Wilcoxon signed-rank test to test for this effect. 
The results of these tests across different types of actions and types of data operated upon are reported in the Appendix (Section \ref{sec:wil_stats}). From these tests, we can infer that participants reported that it was significantly more difficult to verify the accuracy of the replication when they were required to modify values or fix misaligned data in the replication. Further, in addition to modifying values and fixing misaligned data, participants reported that it was significantly more difficult to fix replications when adding new data.


Similarly, we found that participants reported that it was significantly more difficult to fix tables when performing operations on subheadings, row headings, and data values, and reported that it  was significantly more difficult to verify tables when performing operations on row headings and data values. This may be due to the fact that these types of data tend to be embedded within the table and thus may be difficult to locate and to operate on.

To analyze the occurence of the different types of actions and data that participants worked on across the two conditions, we performed McNemar's exact test \cite{McNemar1947}. The results of these tests are reported in the Appendix (Section \ref{sec:mcnemar}). We found that the number of times participants added new data to the replication was significantly higher when working on tables generated by \userstudybaseline as compared to those generated by \userstudysys. Further, participants operated on subheadings a significantly higher number of times when working on tables generated by \userstudybaseline. On the other hand, the number of times participants worked on fixing misaligned data was significantly higher when working on tables generated by \userstudysys. Upon performing the Wilcoxon signed-rank test, we found that participants rated the accuracy of replications to be significantly higher when the fixes to these replications did not require adding any data ($N = 35,z = 52.5, p = .002$) with a moderate effect size ($r = -0.53$). However, we did not find significant differences in the accuracy ratings based on fixing misaligned data in the replication ($N = 16, z  = 18.0, p = .051$). \footnote{$N$ is lesser than 60 for these tests as we only consider those tables for which one replication required performing the action, and the other one did not.} This suggests that participants may have reported that the tables generated by \userstudybaseline were less accurate due to missing data in these tables. However, since both adding missing data and fixing misaligned data significantly increased the effort required to fix the replication, each tool had its own drawbacks that increased the effort for users to verify and fix the replications generated by that tool.



Our results thus indicate that these different approaches to reconstructing tables might be useful in different scenarios. When accuracy is critical, applying a stricter approach that prioritizes maximum coverage (possibly at the cost of alignment issues) might ensure that the replication is useful for users. On the other hand, to ensure that the replication is easy for users to verify, we should apply a more flexible approach that ensures that individual values are accurate and correctly aligned.

\medskip\noindent
\textbf{RQ4}: What is the impact of different components of \sysname?

\ignore{ 
\begin{table*}[t]
\centering
\scriptsize
\setlength{\tabcolsep}{4pt}
\begin{tabular}{@{}llcccc@{}}
\toprule
\textbf{Variant} & \textbf{Metric} & \textbf{Wikipedia} & \textbf{BrokenCSV} & \textbf{PubTabNet} & \textbf{FinTabNet} \\
\midrule
\multirow{6}{*}{\sysname(No feedback)} 
& Coverage   & 0.92  & 0.94  & 0.93  & 0.90 \\
& Halluc.    & 0.13  & 0.17  & 0.18  & 0.17 \\
& TED        & 0.71  & 0.36  & 0.76  & 0.50 \\
& EM         & 15.18 & 60.31 & 17.16  & 12.91 \\
& C.V.M.     & 42.31 & 65.34 & 54.98 & 22.96 \\
& Col.V.M.   & 45.49 & 61.01 & 48.12 & 19.26 \\
\midrule
\multirow{6}{*}{\sysname(Symbolic feedback)} 
& Coverage   & 0.94  & 0.96  & 0.93  & 0.91 \\
& Halluc.    & 0.11  & 0.14  & 0.15  & 0.15 \\
& TED        & 0.68  & 0.30  & 0.75  & 0.48 \\
& EM         & 20.36 & 68.93 & 21.28  & 20.34 \\
& C.V.M.     & 55.84 & 68.56 & 59.96 & 25.36 \\
& Col.V.M.   & 57.69 & 63.35 & 53.35 & 21.12 \\
\midrule
\multirow{6}{*}{\sysname(LLM feedback)} 
& Coverage   & 0.94  & 0.96  & 0.94 & 0.92 \\
& Halluc.    & 0.11  & 0.15  & 0.11 & 0.11 \\
& TED        & 0.66  & 0.31  & 0.73 & 0.48  \\
& EM         & 27.93 & 72.45 & 22.83 & 18.69 \\
& C.V.M.     & 59.01 & 74.93  & 55.69 & 29.12 \\
& Col.V.M.   & 54.69 & 70.89  & 52.87 & 24.95 \\
\midrule
\multirow{6}{*}{\sysname(GPT-4o)} 
& Coverage   & 0.94  & 0.97 & 0.93   & 0.91 \\
& Halluc.    & 0.11  & 0.15 & 0.14   & 0.15 \\
& TED        & 0.65  & 0.28 & 0.67   & 0.47 \\
& EM         & 36.80 & 81.6 & 61.60  & 21.73 \\
& C.V.M.     & 64.36 & 81.03 & 71.54 & 33.92 \\
& Col.V.M.   & 59.59 & 69.56 & 75.69 & 29.36 \\
\bottomrule
\end{tabular}
 \caption{Ablation Study for Variants of \sysname, EM, C.V.M., and Col.V.M. are reported as percentages.}
\label{tab:ablationstudy}
\end{table*}
\endignore}

To assess the contributions of \sysname’s feedback mechanisms, we ablated its components and evaluated four variants under identical prompts, decoding, and evaluation setup: (i) No feedback disables all feedback; (ii) Symbolic feedback enables rule-based checks; (iii) LLM feedback uses critique-and-rewrite without symbolic validation; (iv) \sysname(GPT-4o) combines both. The table with these results is in the main text of the paper.

\textbf{Impact of Removing Feedback Components.} Compared to No feedback, Symbolic feedback consistently improves structural and content metrics, with EM gains of +5.18 (Wikipedia) and +8.62 (BrokenCSV). LLM feedback provides even larger gains on simpler or noisier datasets, e.g., +12.75 EM (Wikipedia). However, on complex tables, it is less stable and may underperform Symbolic feedback on alignment metrics.

\textbf{Full System Performance.} The combined \sysname(GPT-4o) achieves the best performance across all metrics. It offers the highest content fidelity (EM, CVM), lowest TED, and near-saturated coverage (0.90–0.97). In some cases, LLM feedback slightly reduces hallucination compared to the full system, but this comes at the cost of structural fidelity.

%% file: appendix/user_study.tex
\section{Statistical Tests for User Study}
\label{sec:stat_tests}

\subsection{Effect of actions on fix and verification difficulty} \label{sec:wil_stats}
Tables \ref{tab:wilc_fix_action} and \ref{tab:wilc_fix_table_elements} report the results of the Wilcoxon signed-rank test applied to determine the effect of performing each type of actions and of operating on each type of data respectively on participants' responses regarding the difficulty of fixing the replication. Similarly, Tables \ref{tab:wilc_verify_action} and \ref{tab:wilc_verify_table_elements} report the same results for participants' responses regarding the difficulty of verifying the replication. Rows in which statistical significance was found are highlighted in bold.

Each test uses only those tables for which one replication required performing the action (or operating on the data type), and the other one did not. Thus, each test has a different sample size ($N$). For each test, we report the test statistic ($z$), p-value ($p$), and the effect size ($r$). While $\left| r \right| < 0.30$ indicates a small effect size, $ 0.30 \leq \left| r \right| < 0.50 $ indicates a moderate effect size and $0.50 \leq \left| r \right|$ indicates a large effect size.

\begin{table}[t]  
    \centering\scriptsize  
    \setlength{\tabcolsep}{3pt}  
    \begin{tabular}{@{}lccccc c@{}}  
        \toprule  
        \multirow{2}{*}{\textbf{Action Type}}   
        & \multicolumn{2}{c}{\textbf{Average Fix Difficulty Rating (1-7)}}   
        & \multirow{2}{*}{\textbf{$N$}}   
        & \multirow{2}{*}{\textbf{$z$}}   
        & \multirow{2}{*}{\textbf{$p$}}   
        & \multirow{2}{*}{\textbf{$r$}} \\  
        \cmidrule(lr){2-3}  
        & \textbf{With Action} & \textbf{Without Action} & & & & \\  
        \midrule  
        \textbf{Add} & \textbf{4.71} & \textbf{5.66} & \textbf{35} & \textbf{64.0} & \textbf{0.013} & \textbf{-0.42} \\  
        Remove & 5.50 & 5.14 & 22 & 68.5 & 0.452 & -0.16 \\  
        \textbf{Modify} & \textbf{4.36} & \textbf{5.68} & \textbf{25} & \textbf{25.0} & \textbf{0.008} & \textbf{-0.53} \\  
\textbf{Realign} & \textbf{4.63} & \textbf{5.81} & \textbf{16} & \textbf{13.5} & \textbf{0.022} & \textbf{-0.57} \\   
        \bottomrule  
    \end{tabular}  
    \caption{Wilcoxon signed-rank test results for participant ratings on effort required to fix replications with/without performing specific action types. Participant ratings were recorded on a scale from 1 (Very Difficult) to 7 (Very Easy).}  
    \label{tab:wilc_fix_action}  
\end{table}

\begin{table}[t]  
    \centering\scriptsize  
    \setlength{\tabcolsep}{3pt}  
    \begin{tabular}{@{}lccccc c@{}}  
        \toprule  
        \multirow{2}{*}{\textbf{Data Type}}     
        & \multicolumn{2}{c}{\textbf{Average Fix Difficulty Rating (1-7)}}     
        & \multirow{2}{*}{\textbf{$N$}}     
        & \multirow{2}{*}{\textbf{$z$}}     
        & \multirow{2}{*}{\textbf{$p$}}     
        & \multirow{2}{*}{\textbf{$r$}} \\  
        \cmidrule(lr){2-3}  
        & \textbf{With Data Type} & \textbf{Without Data Type} & & & & \\  
        \midrule  
        Column Header       & 5.57 & 5.14 & 21 & 31.0  & 0.293 & -0.23 \\  
        \textbf{Row Header} & \textbf{3.76} & \textbf{5.76} & \textbf{21} & \textbf{12.5} & \textbf{0.002} & \textbf{-0.67} \\  
        \textbf{Subheading} & \textbf{4.68} & \textbf{5.77} & \textbf{22} & \textbf{34.5} & \textbf{0.045} & \textbf{-0.43} \\  
        Note               & 5.89 & 5.56 & 9  & 9.0   & 0.380 & -0.29 \\  
        \textbf{Data}      & \textbf{4.17} & \textbf{6.00} & \textbf{18} & \textbf{0.0} & \textbf{0.001} & \textbf{-0.81} \\  
        \bottomrule  
    \end{tabular}  
    \caption{Wilcoxon signed-rank test results for participant ratings on effort required to fix replications with/without operating on specific data types. Participant ratings were recorded on a scale from 1 (Very Difficult) to 7 (Very Easy).}  
    \label{tab:wilc_fix_table_elements}  
\end{table}  

\begin{table}[t]  
    \centering\scriptsize  
    \setlength{\tabcolsep}{3pt}  
    \begin{tabular}{@{}lccccc c@{}}  
        \toprule  
        \multirow{2}{*}{\textbf{Action Type}}     
        & \multicolumn{2}{c}{\textbf{Average Verification Difficulty Rating (1-7)}}     
        & \multirow{2}{*}{\textbf{$N$}}     
        & \multirow{2}{*}{\textbf{$z$}}     
        & \multirow{2}{*}{\textbf{$p$}}     
        & \multirow{2}{*}{\textbf{$r$}} \\  
        \cmidrule(lr){2-3}  
        & \textbf{With Action} & \textbf{Without Action} & & & & \\  
        \midrule  
        Add                & 5.11 & 5.51 & 35 & 70.5 & 0.190 & -0.22 \\  
        Remove             & 5.45 & 5.09 & 22 & 44.0 & 0.354 & -0.20 \\  
        \textbf{Modify}    & \textbf{4.52} & \textbf{5.52} & \textbf{25} & \textbf{28.5} & \textbf{0.012} & \textbf{-0.50} \\  
        \textbf{Realign} & \textbf{4.69} & \textbf{5.56} & \textbf{16} & \textbf{5.0} & \textbf{0.036} & \textbf{-0.52} \\  
        \bottomrule  
    \end{tabular}  
    \caption{Wilcoxon signed-rank test results for participant ratings on verification effort with/without performing specific action types. Participant ratings were recorded on a scale from 1 (Very Difficult) to 7 (Very Easy).}  
    \label{tab:wilc_verify_action}  
\end{table} 

\begin{table}[t]  
    \centering\scriptsize  
    \setlength{\tabcolsep}{3pt}  
    \begin{tabular}{@{}lccccc c@{}}  
        \toprule  
        \multirow{2}{*}{\textbf{Data Type}}     
        & \multicolumn{2}{c}{\textbf{Average Verification Difficulty Rating (1-7)}}     
        & \multirow{2}{*}{\textbf{$N$}}     
        & \multirow{2}{*}{\textbf{$z$}}     
        & \multirow{2}{*}{\textbf{$p$}}     
        & \multirow{2}{*}{\textbf{$r$}} \\  
        \cmidrule(lr){2-3}  
        & \textbf{With Data Type} & \textbf{Without Data Type} & & & & \\  
        \midrule  
        Column Header       & 5.38 & 5.10 & 21 & 33.0  & 0.373 & -0.19 \\  
        \textbf{Row Header} & \textbf{4.29} & \textbf{5.57} & \textbf{21} & \textbf{27.0} & \textbf{0.010} & \textbf{-0.56} \\  
        Subheading          & 5.36 & 5.73 & 22 & 32.5  & 0.357 & -0.20 \\  
        Note                & 5.44 & 5.44 & 9  & 10.5  & 1.000 & 0.00  \\  
        \textbf{Data}       & \textbf{4.28} & \textbf{5.44} & \textbf{18} & \textbf{10.0} & \textbf{0.012} & \textbf{-0.59} \\  
        \bottomrule  
    \end{tabular}  
    \caption{Wilcoxon signed-rank test results for participant ratings on verification effort with/without operating on specific data types. Participant ratings were recorded on a scale from 1 (Very Difficult) to 7 (Very Easy).}  
    \label{tab:wilc_verify_table_elements}  
\end{table}

\subsection{Occurrence of actions across \userstudysys and \userstudybaseline} 
\label{sec:mcnemar}
\begin{table}
    \centering\scriptsize  
    \setlength{\tabcolsep}{3pt}  
    \begin{tabular}{@{}lccccc c@{}}  
        \toprule  
        \multirow{2}{*}{\textbf{Action}}     
        & \multicolumn{2}{c}{\textbf{Number of occurrences in fixes}}     
        & \multirow{2}{*}{\textbf{$p$}}     
        & \multirow{2}{*}{\textbf{Odds Ratio}} \\  
        \cmidrule(lr){2-3}  
        & \textbf{\userstudysys} & \textbf{\userstudybaseline} & & \\  
        \midrule   
        \textbf{Add}       & \textbf{17} & \textbf{42} & \textbf{$<$0.001} & \textbf{6.00} \\  
        Remove    & 19  & 21 & 0.832 & 1.20 \\  
        Modify    & 26 & 21 & 0.405 & 1.56 \\
                \textbf{Realign} & \textbf{40} & \textbf{28} & \textbf{0.004} & \textbf{7.00} \\ 
        \bottomrule  
    \end{tabular}  
    \caption{McNemar exact test results for occurrence of each action.}
    \label{tab:mcn_actions}  
\end{table}   
\begin{table}
    \centering\scriptsize  
    \setlength{\tabcolsep}{3pt}  
    \begin{tabular}{@{}lccccc c@{}}  
        \toprule  
        \multirow{2}{*}{\textbf{Data Type}}     
        & \multicolumn{2}{c}{\textbf{Number of occurrences in fixes}}     
        & \multirow{2}{*}{\textbf{$p$}}     
        & \multirow{2}{*}{\textbf{Odds Ratio}} \\  
        \cmidrule(lr){2-3}  
        & \textbf{TEN} & \textbf{Baseline} & & \\  
        \midrule   
        Column Header    & 39 & 34 & 0.383 & 1.62 \\  
        Row Header       & 18 & 23 & 0.383 & 1.62 \\  
        \textbf{Subheading}       & \textbf{7}  & \textbf{27} & \textbf{$<$0.001} & \textbf{21.00} \\  
        Note             & 20 & 21 & 1.000 & 1.25 \\
        Data             & 18 & 26 & 0.096 & 2.60 \\ 
        \bottomrule  
    \end{tabular}  
    \caption{McNemar exact test results for occurrence of each data type}
    \label{tab:mcn_data}  
\end{table}  
Tables \ref{tab:mcn_actions} and \ref{tab:mcn_data} report the results of applying the McNemar exact test \cite{McNemar1947} across \userstudysys and \userstudybaseline to analyze the differences in the occurrence of each type of action and each data type in fixing the replications generated by these tools. We use the exact test as we have smaller sample sizes -- the total number of pairs in our sample is $N = 60$.
We report the p-values ($p$) and the odds ratio (OR) for each test, along with the number of occurrences for each action/data type for each tool. We calculate odds ratio as the maximum of $b/c$ and $c/b$ for the  2 $\times$ 2 contingency table:
$$ 
\begin{pmatrix}  
a & b \\  
c & d  
\end{pmatrix}  
$$

\section{User Study Questionnaire}
\label{sec:questionnaire}
Following are the questions participants in our user study answered after fixing each replication. We use the Single Effort Question (SEQ) \cite{seq} to gather perceived effort to fix and verify the replication, and record responses in 7-point Likert scales.

\begin{itemize}
    \item Please rate your agreement with the following statement: The copy accurately replicates the structure of the original table.\\  
    \textit{Scale: Strongly Disagree, Disagree, Somewhat Disagree, Neutral, Somewhat Agree, Agree, Strongly Agree}  

    \item Overall, how difficult or easy was it to verify the accuracy of the copy?\\  
    \textit{Scale: Very Easy, Easy, Somewhat Easy, Neutral, Somewhat Difficult, Difficult, Very Difficult}  

    \item Overall, how difficult or easy was it to fix the copy?\\  
    \textit{Scale: Very Easy, Easy, Somewhat Easy, Neutral, Somewhat Difficult, Difficult, Very Difficult}

\end{itemize}

%% file: appendix/StructuralDecompositionExample.tex
\newpage
\section{Structural Decomposition Prompting Applied to PDF-Copied Table}
\label{app:structural-decomposition}
\par
\begin{figure}[htbp]
    \centering
    \includegraphics[width=0.8\linewidth]{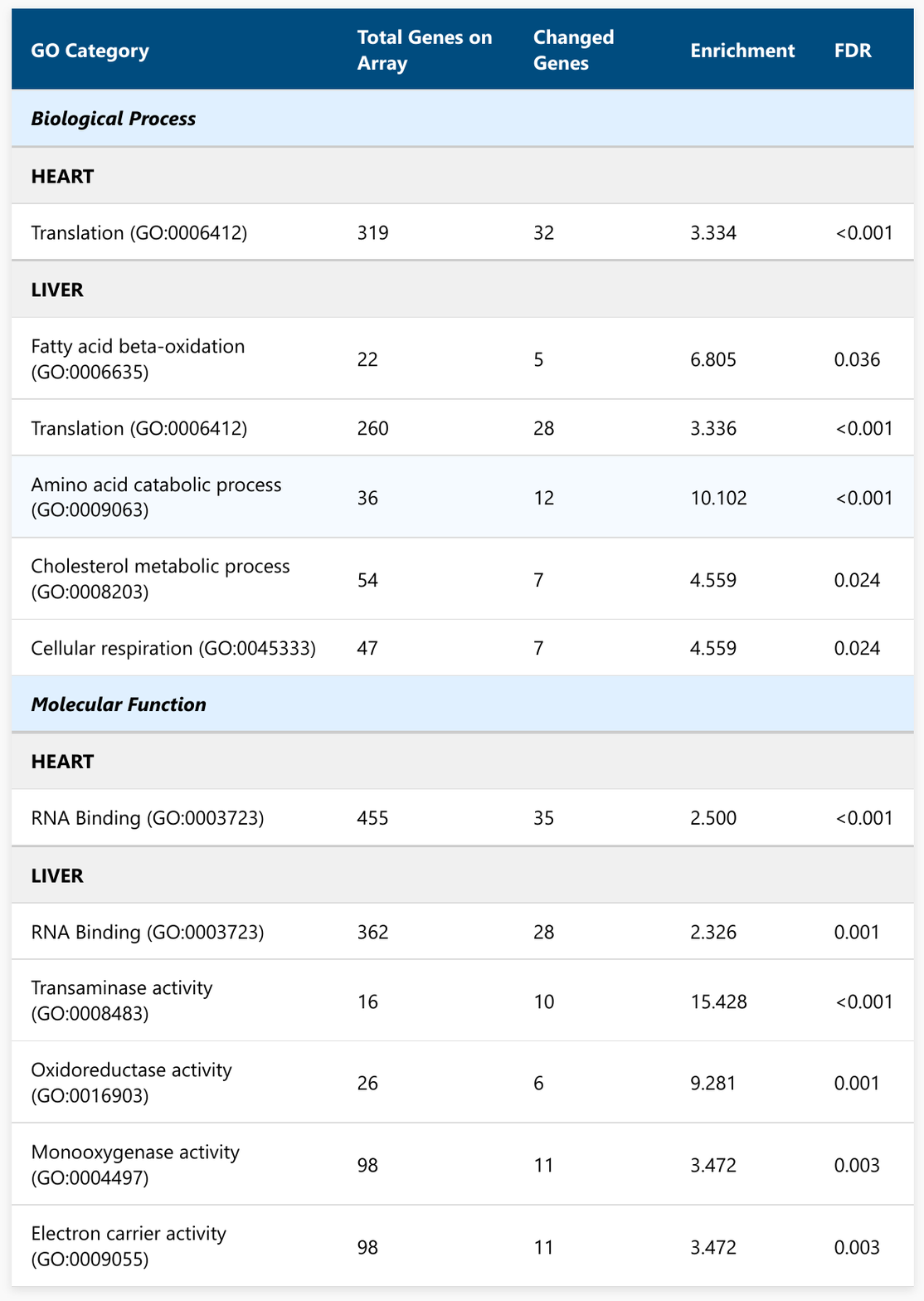}
    \caption{Original structured GO annotation table taken from PubTabNet dataset. It contains multiple biological categories (Biological Process, Molecular Function) with tissue-specific subgroups (HEART, LIVER) and associated enrichment statistics.}
    \label{fig:gocategorytable}
\end{figure}

 Figure~\ref{fig:gocategorytable} displays an example of a semantically rich table containing biological annotations. When such tables are copy-pasted from a PDF, they often lose structural cues such as headers, groupings, and cell boundaries, resulting in a flattened and hard-to-interpret text block (see Listing~\ref{lst:ten-input-example}). This degraded input poses a challenge for table reconstruction systems.

\par \sysname, addresses this challenge by explicitly reconstructing the hierarchical structure of such flattened inputs. It identifies distinct categories and subgroups, organizes rows under appropriate headers, and outputs clean, semantically organized HTML tables (see Listing~\ref{lst:json-output}).

\begin{listing}
\begin{minted}[fontsize=\footnotesize, breaklines=true, breakanywhere=true, linenos=false]{text}
GO Category Total Genes on
Array
Changed
Genes Enrichment FDR
Biological Process
HEART
Translation (GO:0006412) 319 32 3.334 <0.001
LIVER
Fatty acid beta-oxidation
(GO:0006635) 22 5 6.805 0.036
Translation (GO:0006412) 260 28 3.336 <0.001
Amino acid catabolic process
(GO:0009063) 36 12 10.102 <0.001
Cholesterol metabolic process
(GO:0008203) 54 7 4.559 0.024
Cellular respiration (GO:0045333) 47 7 4.559 0.024
Molecular Function
HEART
RNA Binding (GO:0003723) 455 35 2.500 <0.001
LIVER
RNA Binding (GO:0003723) 362 28 2.326 0.001
Transaminase activity
(GO:0008483) 16 10 15.428 <0.001
Oxidoreductase activity
(GO:0016903) 26 6 9.281 0.001
Monooxygenase activity
(GO:0004497) 98 11 3.472 0.003
Electron carrier activity
(GO:0009055) 98 11 3.472 0.003
\end{minted}
\caption{Input provided to \sysname for table explicitization.}
\label{lst:ten-input-example}
\end{listing}

\begin{listing}
\caption{Tables decomposed by \sysname from structurally ambiguous input. Each table corresponds to a semantic category (e.g., Biological Process, Molecular Function) and includes GO annotations with enrichment statistics.}
\label{lst:json-output}
\begin{minted}[fontsize=\scriptsize, breaklines=true,breakanywhere=true, ]{json}
{
  "tables": [
    {
      "starting_token": "Biological Process",
      "html_output": "<table><thead><tr><th>GO Category</th><th>Total Genes on Array</th><th>Changed Genes</th><th>Enrichment</th><th>FDR</th></tr></thead><tbody><tr><td>Biological Process</td></tr><tr><td>HEART Translation (GO:0006412)</td><td>319</td><td>32</td><td>3.334</td><td>&lt;0.001</td></tr><tr><td>LIVER Fatty acid beta-oxidation (GO:0006635)</td><td>22</td><td>5</td><td>6.805</td><td>0.036</td></tr><tr><td>Translation (GO:0006412)</td><td>260</td><td>28</td><td>3.336</td><td>&lt;0.001</td></tr><tr><td>Amino acid catabolic process (GO:0009063)</td><td>36</td><td>12</td><td>10.102</td><td>&lt;0.001</td></tr><tr><td>Cholesterol metabolic process (GO:0008203)</td><td>54</td><td>7</td><td>4.559</td><td>0.024</td></tr><tr><td>Cellular respiration (GO:0045333)</td><td>47</td><td>7</td><td>4.559</td><td>0.024</td></tr></tbody></table>"
    },
    {
      "starting_token": "Molecular Function",
      "html_output": "<table><thead><tr><th>GO Category</th><th>Total Genes on Array</th><th>Changed Genes</th><th>Enrichment</th><th>FDR</th></tr></thead><tbody><tr><td>Molecular Function</td></tr><tr><td>HEART RNA Binding (GO:0003723)</td><td>455</td><td>35</td><td>2.500</td><td>&lt;0.001</td></tr><tr><td>LIVER RNA Binding (GO:0003723)</td><td>362</td><td>28</td><td>2.326</td><td>0.001</td></tr><tr><td>Transaminase activity (GO:0008483)</td><td>16</td><td>10</td><td>15.428</td><td>&lt;0.001</td></tr><tr><td>Oxidoreductase activity (GO:0016903)</td><td>26</td><td>6</td><td>9.281</td><td>0.001</td></tr><tr><td>Monooxygenase activity (GO:0004497)</td><td>98</td><td>11</td><td>3.472</td><td>0.003</td></tr><tr><td>Electron carrier activity (GO:0009055)</td><td>98</td><td>11</td><td>3.472</td><td>0.003</td></tr></tbody></table>"
    }
  ]
}
\end{minted}
\end{listing}

%% file: appendix/Prompt_Details.tex
\newpage
\onecolumn
\section{Prompts for Table Explicitization}
\subsection{Prompt Used For Structural Decomposition.}
\label{app:structural-decomposition_prompt}
\begin{promptrefbox}[box:ten_table_generation]{Prompt for Extracting Table}
\textbf{Role:} You are an expert in interpreting diverse table formats. Your task is to generate structured tables from unstructured text. The primary objective is to \textbf{generalize} the process of table extraction and representation.

\textbf{Instructions:}
\begin{enumerate}[label=(\arabic*)]
    \item Review the input text for the presence of tabular data.
    \item Identify the most appropriate row delimiter that separates the input into distinct rows. These delimiters can be system-specific (e.g., \verb|\r\n|, \verb|<br>|). If no such delimiter suffices, propose a custom regular expression-based delimiter and report it under \texttt{\$row\_delimiter\$}.
    \item Identify row headers that signify the start of sub-tables or logical sections. Row headers are typically distinct lines (e.g., titles, section markers) that separate groups of rows into meaningful partitions.
     \item If row headers exist, split the table into multiple logical partitions at these row headers. Return these row headers under \texttt{\$starting\_token\$} as partition markers.
    \item If no row headers or no multiple tables exist, treat the input as a single table. Set \texttt{\$starting\_token\$} to \texttt{null}.
    \item For each partitioned table or the single table, generate a corresponding HTML \verb|<table>| representation with column headers.
    \item Infer the number of columns by analyzing all rows. For rows with fewer columns, pad them with empty cells (\verb|<td></td>| or \verb|<th></th>|) to maintain uniformity.
    \item If a cell should span multiple rows or columns (e.g., merged cells), preserve it as-is. Do not split such cells to force a rectangular shape. Instead, pad the surrounding rows/columns appropriately to retain a rectangular layout.
    \item Wrap all column headers inside \verb|<thead>| and all data rows (including row headers, if any) inside \verb|<tbody>|.
    \item Do not remove, skip, or modify any content from the input text. Do not add any additional content or annotations. Do not correct spelling, formatting, or whitespace in the input.
    \item Return the final output as structured JSON, encapsulated in a code block, using the following format:
\end{enumerate}

\begin{lstlisting}[language=json]
{
  "tables": [
    {
      "$starting_token$": "<identified_row_header_1>",
      "$html_output$": "<html_representation_of_table_1>"
    },
    ... 
  ],
  "$row_delimiter$": "<row_delimiter_used>"
}
\end{lstlisting}

\textbf{Input:} \texttt{{\{\{Input Text\}\}}}
\end{promptrefbox}
\subsection{Prompt Used For Critique Generation.}
\label{app:critique_generation}
\begin{promptrefbox}[box:ten_critique_prompt]{Prompt used for Critique LLM}
Below is a table that is constructed from a noisy text.  
\textbf{Role:} You are an expert in fixing messy tabular data extracted from unstructured text.

\textbf{Instructions:}
Your task is to critically assess whether the \textbf{table is correct or messy} using your understanding of what a clean table should look like.

If you find actual structural or formatting issues, explain what they are and how to fix them.  
If the table structure is clear, consistent, and logical, do not suggest unnecessary changes. Prioritize semantic and visual clarity over strict rule adherence.

\textbf{Do not:}
\begin{enumerate}[label=(\arabic*)]
    \item Add new columns or data
    \item Delete any existing columns or data
    \item Hallucinate new entries or remove original values
    \item Add indentations or extra spaces in any cells
    \item Correct spelling, reformat dates, or fill empty cells with placeholders like “N/A”
\end{enumerate}

\textbf{Do not change Semantics. Only the positions of content may be changed to fix structural issues.}

\medskip
\textbf{Your critique:}
\end{promptrefbox}

\subsection{Prompt Used for Table Regeneration.}
\label{app:regeneration_prompt}
\begin{promptrefbox}[box:ten_table_regeneration]{Prompt for Table Regeneration Based on Structural Critique}
\textbf{Role:} You are a table regeneration agent.
\textbf{Instructions:}
You will be given a previously generated table along with feedback highlighting structural or formatting issues.

Your task is to regenerate the table in \textbf{HTML format}, correcting issues such as:
\begin{itemize}[noitemsep,topsep=0pt]
    \item Merged cells
    \item Row alignment
    \item Header misplacement
\end{itemize} 

\textbf{Do Not:}
\begin{itemize}[noitemsep,topsep=0pt]
    \item Add new rows or columns.
    \item Hallucinate any values.
\end{itemize}

You may split or reassign existing values across rows/columns to improve structure.

\medskip
\textbf{Return the regenerated table in the following structured JSON format.:}

\begin{lstlisting}[language=json]
{
  "tables": [
    {
      "html_output": "<html_table_with_<thead>_and_<tbody>>"
    },
    ...
  ]
}
\end{lstlisting}

\textbf{Critique:}  
{\textless{}Insert critique here\textgreater{}}

\textbf{Original Table:}  
{\textless{}Insert table rows here\textgreater{}}
\end{promptrefbox}

\subsection{Prompts used for Baseline}
Table extraction prompt is shown Box in \ref{box:general_table_extraction}. Critique and table regeneration prompt are same as in Box \ref{box:ten_critique_prompt} and Box \ref{box:ten_table_regeneration}.

\begin{promptrefbox}[box:general_table_extraction]{Prompt for Generalized Table Extraction}
\textbf{Role:} \textbf{Role:} You are an expert in interpreting diverse table formats. Your task is to generate structured tables from unstructured text. The primary objective is to \textbf{generalize} the process of table extraction and representation.

\textbf{Instructions:}
1. Examine the following unstructured text to identify any tabular data.\\
2. Convert the identified table(s) into HTML, ensuring that:  
   \begin{itemize}
     \item Column headers (table headers) are wrapped within \verb|<thead></thead>|.
     \item All data rows are wrapped within \verb|<tbody></tbody>|.
   \end{itemize}
3. You must return the table in structured JSON format inside code blocks as shown below.

\textbf{Input:}  
\begin{verbatim}
{{Input Text}}
\end{verbatim}

\textbf{Output:}
\begin{lstlisting}[language=json]
{
  "tables": [
    {
      "html_output": "<html_table_with_<thead>_and_<tbody>>"
    },
    ...
  ]}
\end{lstlisting}
\end{promptrefbox}
\twocolumn

%% file: appendix/Qualitative_Analysis_of_Ablation.tex
\newpage
\section{Ablation Qualitative Analysis}

\subsection{Neurosymbolic Feedback Effectively Removes Hallucinated Columns and Restores Table Schema}
\label{app:qual_analysis_1}
\begin{figure}
    \centering
    \begin{minipage}[t]{\linewidth}
        \centering
        \includegraphics[width=\linewidth]{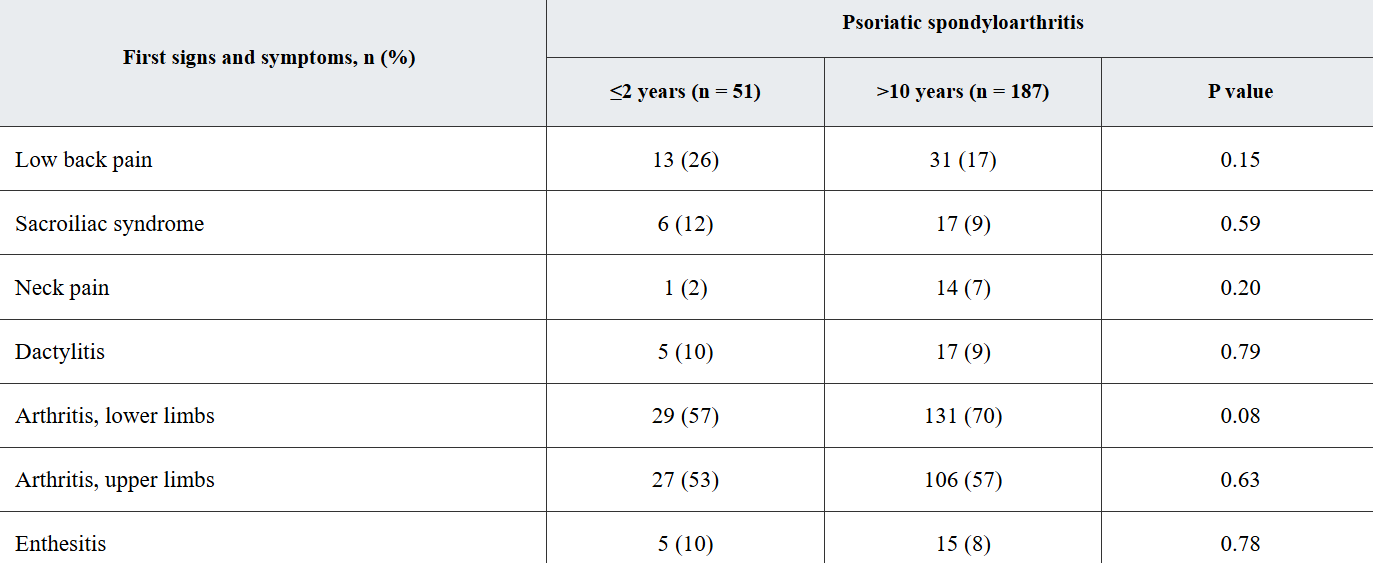}
        \caption{Ground truth}
        \label{fig:gt}
    \end{minipage}
\end{figure}

\begin{figure}[t]
    \includegraphics[width=\linewidth]{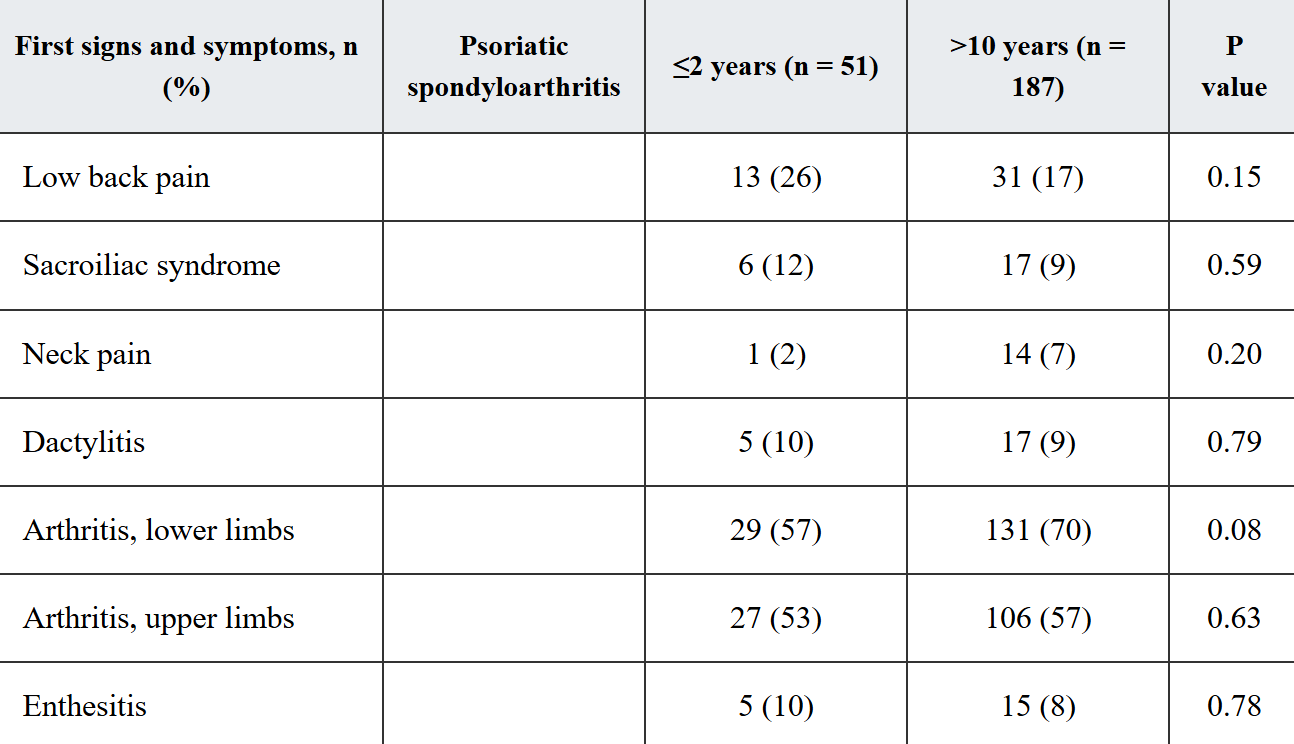}
    \caption{Initial table (hallucinated column ``Psoriatic spondyloarthritis'')}
    \label{fig:init_hallucinated_table}
\end{figure}

\begin{figure}
    \includegraphics[width=\linewidth]{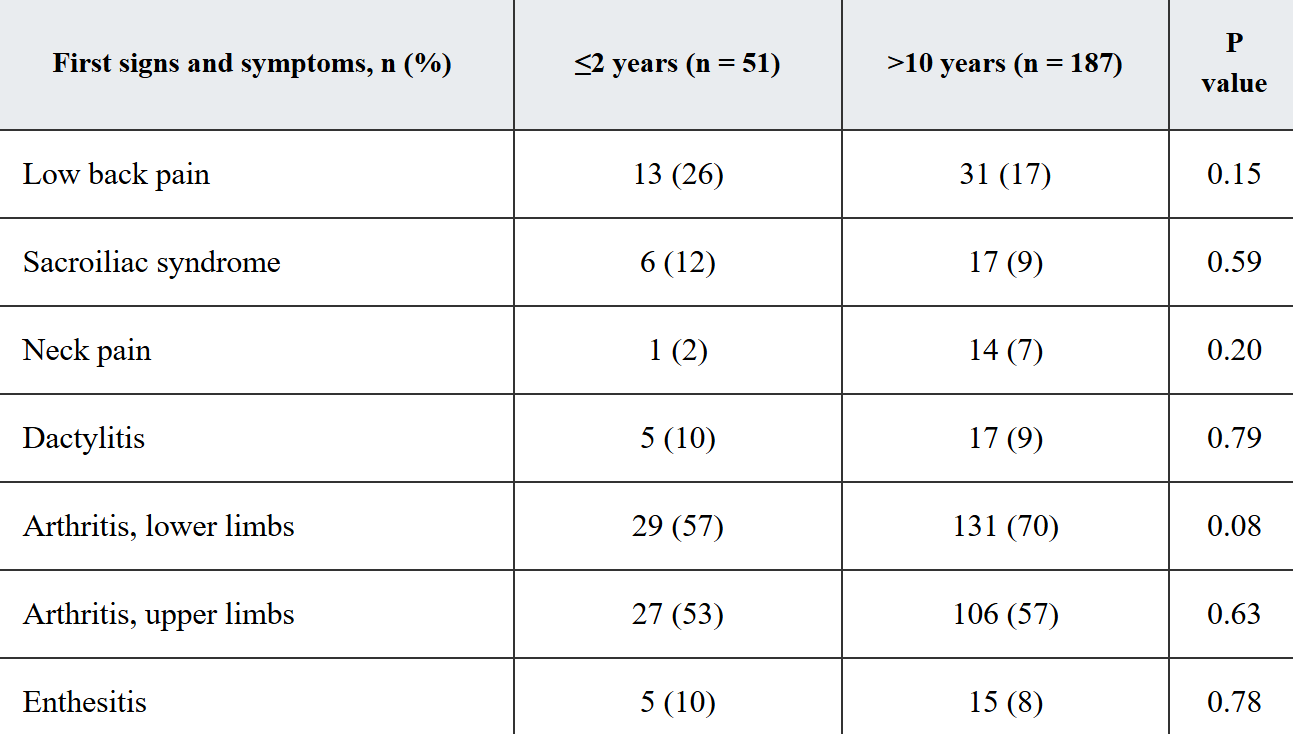}
    \caption{Regenerated table (schema fixed)}
    \label{fig:regenerated_fixed_table}
\end{figure}

In the initial output, the LLM incorrectly interpreted the label "Psoriatic spondyloarthritis" as a data column header (Figure \ref{fig:init_hallucinated_table}), causing the insertion of an empty and unnecessary second column. This misinterpretation also flattened what was originally intended as a hierarchical two-row group header structure (Figure \ref{fig:gt}). The critique from the intermediate feedback stage correctly identified the header misalignment and flagged the redundant empty column, explicitly suggesting that "Psoriatic spondyloarthritis" should serve as a group label rather than a column header (Figure \ref{fig:regenerated_fixed_table}).

Guided by this detailed critique (Listing \ref{lst:qual_feedback_critique}), the NeuroSymbolic loop regenerated the table by removing the hallucinated column and correctly positioning "Psoriatic spondyloarthritis" as a group-level descriptor. 

\lstset{
    basicstyle=\ttfamily\tiny, 
    breaklines=true, 
    frame=single, 
    backgroundcolor=\color{gray!10}, 
    rulecolor=\color{black}, 
    keywordstyle=\color{blue}, 
    showspaces=false, 
    showstringspaces=false, 
    tabsize=1, 
    caption={Intermediate Feedback for Table Structure in Ablation Qualitative Analysis}, 
    label={lst:table_critique}, 
}

\begin{lstlisting}[
caption={Critique Feedback Identifying Structural Issues in the Initial Table},
label={lst:qual_feedback_critique}
]
The table extracted from the noisy text has several structural issues that need to be addressed for clarity and consistency. Here's a breakdown of the issues and how to fix them:
1. Column Headers: The first row of the table contains column headers, but the header for the second column ("Psoriatic spondyloarthritis") is not aligned with the data. It seems to be a category rather than a header.  The actual headers should be:
   - "First signs and symptoms, n (%)"
   - "$\leq$2 years (n = 51)"
   - "$\ge$10 years (n = 187)"
   - "P value"
2. Empty Cells in the Second Column: The second column is empty for all rows, which suggests that the "Psoriatic spondyloarthritis" label might be misplaced. This column should be removed or the header should be integrated into the table in a way that makes sense, such as a title or a category label above the table.
3. Row Structure: Each row should consistently represent a symptom and its corresponding data across the time categories and p-value. The current structure is correct in terms of data placement, but the empty second column creates confusion.
4. Signature Mismatches: The rule-based signals indicate type inconsistencies, but these are primarily due to the empty second column and the misalignment of headers. Once the column header issue is resolved, these mismatches should be addressed.
To fix the table:
- Remove the second column entirely or reposition the "Psoriatic spondyloarthritis" label as a title above the table.
- Ensure the headers are clearly defined as:
  - "First signs and symptoms"
  - "$\leq$2 years (n = 51)"
  - "$\ge$10 years (n = 187)"
  - "P value"
- Maintain the current data structure for each symptom row, as it is consistent and logical once the header issue is resolved.
By addressing these issues, the table will be clearer and consistent without altering the original data content.
\end{lstlisting}


\subsection{Comparing LLM-Only and NeuroSymbolic Feedback}
\label{app:qual_analysis_2}


\begin{figure*}[t]
    \centering
    \begin{subfigure}[t]{0.48\linewidth}
        \centering
        \includegraphics[width=\linewidth]{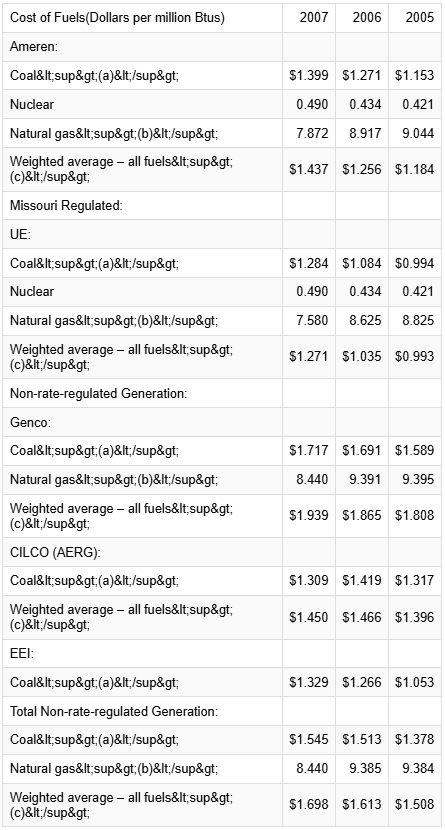}
        \caption{Ground Truth}
        \label{fig:gt_fuel}
    \end{subfigure}
    \hfill
    \begin{subfigure}[t]{0.48\linewidth}
        \centering
        \includegraphics[width=\linewidth]{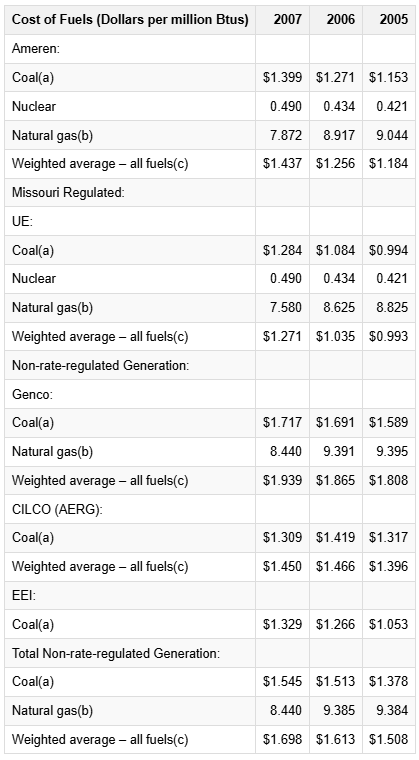}
        \caption{NeuroSymbolic (hierarchy preserved)}
        \label{fig:neurosymbolic_feedback}
    \end{subfigure}
    \caption{Comparison of outputs with NeuroSymbolic feedback vs LLM-only feedback.}
    \label{fig:neurosym_corrected}
\end{figure*}
\begin{figure}[t]
    \centering
    \includegraphics[width=\linewidth]{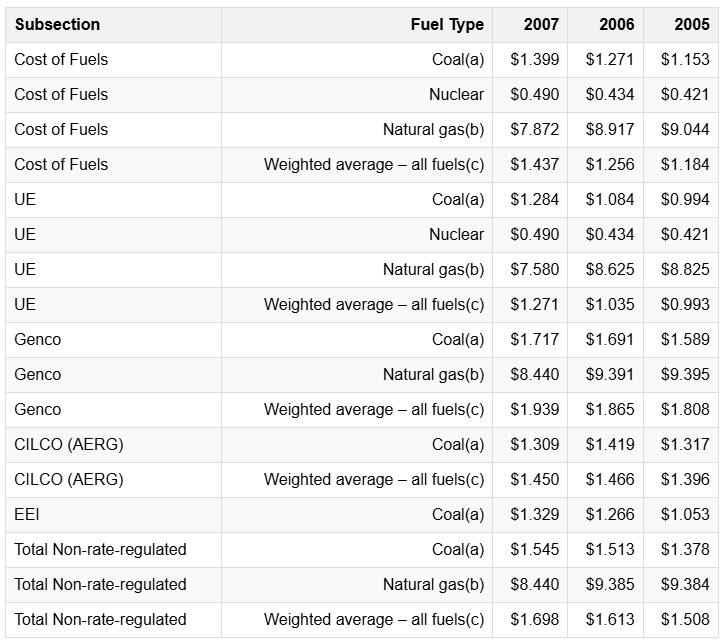}
    \caption{LLM-only (flattened structure)}
    \label{fig:llm_flattened}
\end{figure}

In this example, the table contains hierarchical subsection labels (e.g., "Ameren", "UE", "Genco") interleaved with fuel cost data across three years (2007–2005). The initial table output failed to properly distinguish these group headers from data rows, leading to a flat and misaligned structure that impaired readability and accuracy (Figure~\ref{fig:gt_fuel}).
\begin{lstlisting}[
caption={LLM-Only Feedback Flattened the Hierarchy and Suggested Column Reorganization},
label={lst:llm_only_feedback}
]
Issues Identified:
1. Inconsistent Entity Structure in Rows: The table alternates between headers (e.g., "UE:", "Genco:", "CILCO (AERG):", etc.) and data rows without clear separation or alignment. These headers are not properly distinguished from the data rows, making it difficult to interpret the table structure.
2. Misalignment of Headers and Data: The headers (e.g., "UE:", "Genco:", etc.) are placed in the same column as the data, which disrupts the logical flow of the table. Headers should ideally be separated or visually distinct from the data rows.
3. Inconsistent Formatting of Values: Some values include a dollar sign (e.g., "$1.399"), while others do not (e.g., "0.490"). This inconsistency can confuse readers and should be standardized.
4. Missing Column Headers for Subsections: Subsections like "UE:", "Genco:", "CILCO (AERG):", etc., do not have their own column headers, making it unclear what these sections represent. This lack of clarity affects the readability of the table.
5. Ambiguity in Weighted Average Rows: Rows labeled "Weighted average - all fuels(c)" appear multiple times under different subsections, but it is unclear whether these averages are specific to the subsection or represent a global average. This ambiguity should be clarified.
6. Rule-Based Signals: The flagged rule-based signals highlight potential type inconsistencies in various cells. While these signals are not always accurate, they point to areas where the table's structure may be inconsistent or unclear.
Proposed Fixes:
To address these issues, the table should be reorganized to ensure structural clarity while preserving the original content and formatting. Below are the steps to fix the table:
1. Separate Headers from Data Rows: Move subsection headers (e.g., "UE:", "Genco:", etc.) into a dedicated column to distinguish them from the data rows.
2. Standardize Formatting of Values: Ensure all monetary values consistently include a dollar sign (e.g., "$0.490" instead of "0.490").
3. Clarify Subsection Context: Add a column for subsection names (e.g., "UE", "Genco", etc.) to make it clear which subsection each row belongs to.
4. Align Data Rows Under Correct Headers: Ensure that all data rows are properly aligned under the correct year columns (2007, 2006, 2005).
\end{lstlisting}

The LLM-only feedback attempted to regularize this structure by converting subsection labels into a Subsection column and fuel types into a Fuel Type column (Figure~\ref{fig:llm_flattened}). While this improved visual uniformity, it fundamentally altered the table's schema—flattening the hierarchy, duplicating fuel type rows, and disconnecting values from their group context. As a result, this version exhibited lower structural fidelity, despite being more readable.
\begin{lstlisting}[
caption={NeuroSymbolic Feedback Identified Header Misalignment and Preserved Hierarchical Grouping},
label={lst:ten_feedback}
]
Critique: The table extracted from the noisy text has several structural issues that need to be addressed for clarity and consistency. Here are the identified issues and suggestions for fixing them:
1. Entity Structure Consistency: The rule-based signals correctly identify rows where the entity structure is inconsistent. The table mixes headers and data rows without clear separation, which can confuse the reader. For example, "Ameren:" and "Missouri Regulated:" are headers that should be clearly distinguished from the data rows.
2. Header and Subheader Alignment: The headers such as "Ameren:", "Missouri Regulated:", "UE:", "Non-rate-regulated Generation:", "Genco:", "CILCO (AERG):", and "EEI:" should be aligned properly to indicate they are categories or subcategories. This can be achieved by placing them in a separate column or using indentation to visually separate them from the data rows.
3. Column Consistency: The table should maintain consistent column headers across all rows. The main header "Cost of Fuels(Dollars per million Btus) | 2007 | 2006 | 2005" should be clearly separated from the data rows, and each data row should align under these headers.
4. Visual Clarity: The table lacks visual clarity due to the mixing of headers and data. To improve this, headers should be bolded or separated by lines to distinguish them from the data rows.
5. Grouping of Related Data: The data related to each entity (e.g., Ameren, UE, Genco) should be grouped together to maintain logical flow. This can be done by ensuring that all rows related to a specific entity are contiguous and clearly marked.
6. Signature Mismatch: The signals indicate type inconsistencies, likely due to the mixing of headers and data. Ensuring that headers are not treated as data rows will resolve these mismatches.
To fix these issues, the table should be restructured to clearly separate headers from data rows, maintain consistent column alignment, and visually group related data. This will enhance semantic and visual clarity without altering the original text content.
\end{lstlisting}

In contrast, the NeuroSymbolic feedback correctly preserved group headers as distinct non-data rows, maintained value alignment under year columns, and reduced hallucination (Figure~\ref{fig:neurosym_corrected}). The regenerated table was semantically faithful to the original intent and achieved better alignment with the ground truth, even though minor differences in formatting remained.

%% file: appendix/FinReCon60DatasetDetails.tex
\newpage
\onecolumn
\section{Qualitative Analysis of \sysname's Output}
\label{sec:FinReCon60DatasetDetails}

\par
\begin{figure}
    \centering
    \includegraphics[width=0.8\linewidth]{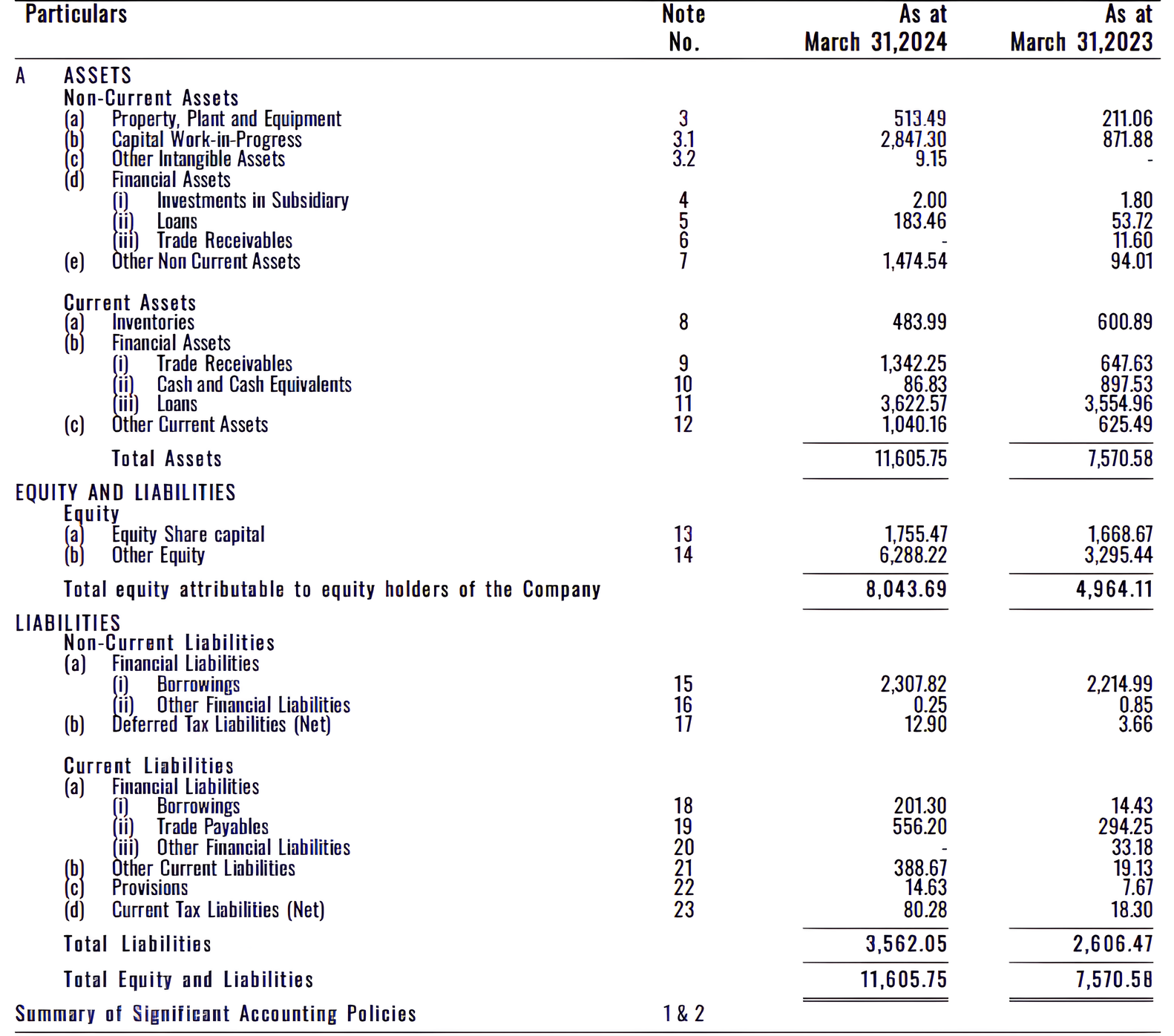}
    \caption{Example of a real-world financial table from an annual report (Mercury EV – Tech Ltd., 2024) from FinRecon20 dataset. This balance sheet exhibits several structural challenges relevant to table explicitization, including hierarchical row groupings, multi-line labels, and alignment-dependent value placement.}
    \label{fig:example_1_png}
\end{figure}

 This example demonstrates the table extraction and explicitization capability of \sysname using a real-world financial balance sheet from Mercury EV-Tech Ltd. (Annual Report, 2024), taken from the FinReCon20 dataset. Figure \ref{fig:example_1_png} illustrates the original balance sheet from the financial document, exhibiting several structural complexities such as hierarchical row groupings (Assets and Liabilities), multi-line textual labels (e.g., “Property, Plant and Equipment”), and numerical values whose meaning is dependent on alignment and indentation.
\subsection{Example 1}

\begin{lstlisting}[language=,caption={ Input text to \sysname: a structurally ambiguous table representation resulting from manual copy-pasting from a financial report. This serves as the source for table explicitization.},label={lst:finrecon20_input}]
  Particulars Note As at As at\r\nNo. March 31,2024 March 31,2023\r\nA ASSETS\r\nNon-Current Assets\r\n(a) Property, Plant and Equipment 3 513.49 211.06\r\n(b) Capital Work-in-Progress 3.1 2,847.30 871.88\r\n(c) Other Intangible Assets 3.2 9.15 -\r\n(d) Financial Assets\r\n(i) Investments in Subsidiary 4 2.00 1.80\r\n(ii) Loans 5 183.46 53.72\r\n(iii) Trade Receivables 6 - 11.60\r\n(e) Other Non Current Assets 7 1,474.54 94.01\r\nCurrent Assets\r\n(a) Inventories 8 483.99 600.89\r\n(b) Financial Assets\r\n(i) Trade Receivables 9 1,342.25 647.63\r\n(ii) Cash and Cash Equivalents 10 86.83 897.53\r\n(iii) Loans 11 3,622.57 3,554.96\r\n(c) Other Current Assets 12 1,040.16 625.49\r\nTotal Assets 11,605.75 7,570.58\r\nEQUITY AND LIABILITIES\r\nEquity\r\n(a) Equity Share capital 13 1,755.47 1,668.67\r\n(b) Other Equity 14 6,288.22 3,295.44\r\nTotal equity attributable to equity holders of the Company 8,043.69 4,964.11\r\nLIABILITIES\r\nNon-Current Liabilities\r\n(a) Financial Liabilities\r\n(i) Borrowings 15 2,307.82 2,214.99\r\n(ii) Other Financial Liabilities 16 0.25 0.85\r\n(b) Deferred Tax Liabilities (Net) 17 12.90 3.66\r\nCurrent Liabilities\r\n(a) Financial Liabilities\r\n(i) Borrowings 18 201.30 14.43\r\n(ii) Trade Payables 19 556.20 294.25\r\n(iii) Other Financial Liabilities 20 - 33.18\r\n(b) Other Current Liabilities 21 388.67 19.13\r\n(c) Provisions 22 14.63 7.67\r\n(d) Current Tax Liabilities (Net) 23 80.28 18.30\r\nTotal Liabilities 3,562.05 2,606.47\r\nTotal Equity and Liabilities 11,605.75 7,570.58\r\nSummary of Significant Accounting Policies 1 & 2
\end{lstlisting}

\par
\begin{figure}
\centering
    \includegraphics[width=0.8\linewidth]{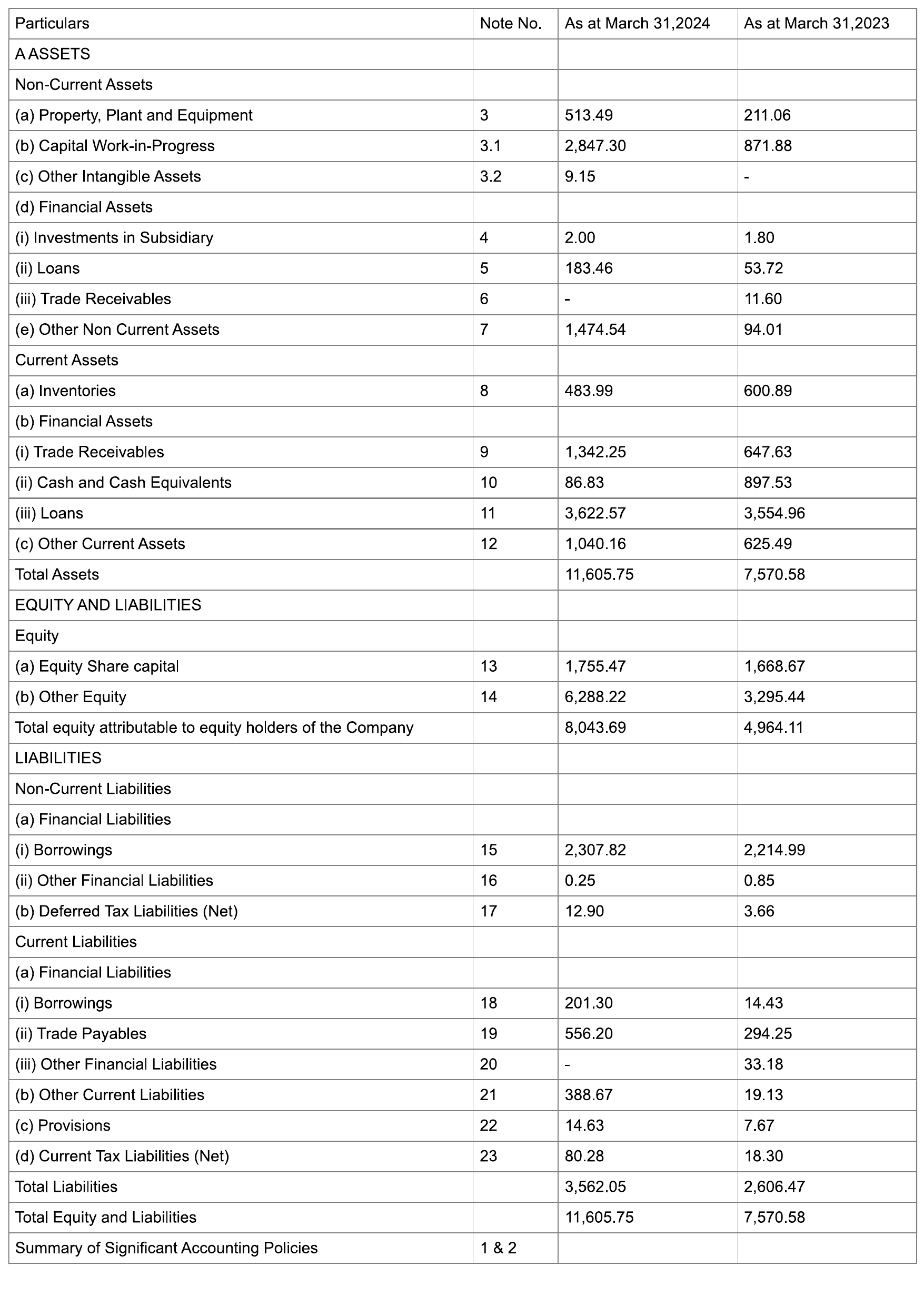}
    \caption{ Structured balance sheet generated by the \sysname from unstructured copy-paste input. \sysname accurately recovers multi-level headings, financial hierarchies, and aligned numerical values, closely matching the original document layout.}
    \label{fig:ten_output_1}
\end{figure}

 Listing \ref{lst:finrecon20_input} provides the raw, structurally ambiguous input text that results from a typical manual copy-paste operation from the PDF document. Notice that the hierarchical relationships and structural semantics of the table are lost due to flattening, inconsistent spacing, and disrupted formatting. Such ambiguities pose significant challenges for traditional extraction methods relying solely on visual layout cues.
Figure \ref{fig:ten_output_1} presents the output produced by \sysname. The system successfully reconstructs the table structure, accurately identifying and restoring:
\begin{itemize}
\item Multi-level hierarchical relationships (e.g., clearly distinguishing between Non-Current and Current Assets and Liabilities).
\item Complex, multi-line row labels, correctly segmented and aligned.
\item Numerical value placements matching the original structural semantics of the balance sheet.
\end{itemize}

This demonstrates the effectiveness of \sysname in recovering structured representations from highly ambiguous, flattened textual inputs, significantly reducing manual table-reconstruction effort for end-users.

\subsection{Example 2}
\par
\begin{figure}
    \centering
    \includegraphics[width=\linewidth]{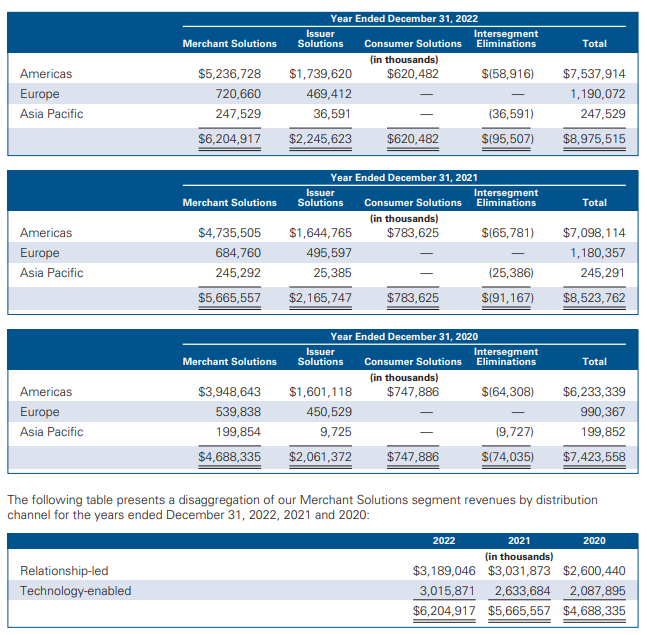}
    \caption{ Ground truth table of segment-wise revenues by geography and product line, extracted from a corporate annual report. The table spans three years (2020–2022) and contains five financial dimensions: Merchant Solutions, Issuer Solutions, Consumer Solutions, Intersegment Eliminations, and Totals. Complexities include repeated multi-level headers, hierarchical geographies, and negative values, all of which challenge structure recovery in copy-pasted representations.}
    \label{fig:example_3_png}
\end{figure}

 This example showcases the reconstruction performance of \sysname on a challenging, multi-segment revenue breakdown table from a corporate annual report. The ground-truth table (Figure \ref{fig:example_3_png}) captures revenues segmented geographically (Americas, Europe, Asia Pacific) and by business solutions (Merchant, Issuer, Consumer Solutions) across three consecutive financial years (2020–2022). 
\par
\begin{figure}
\centering
    \includegraphics[width=\linewidth]{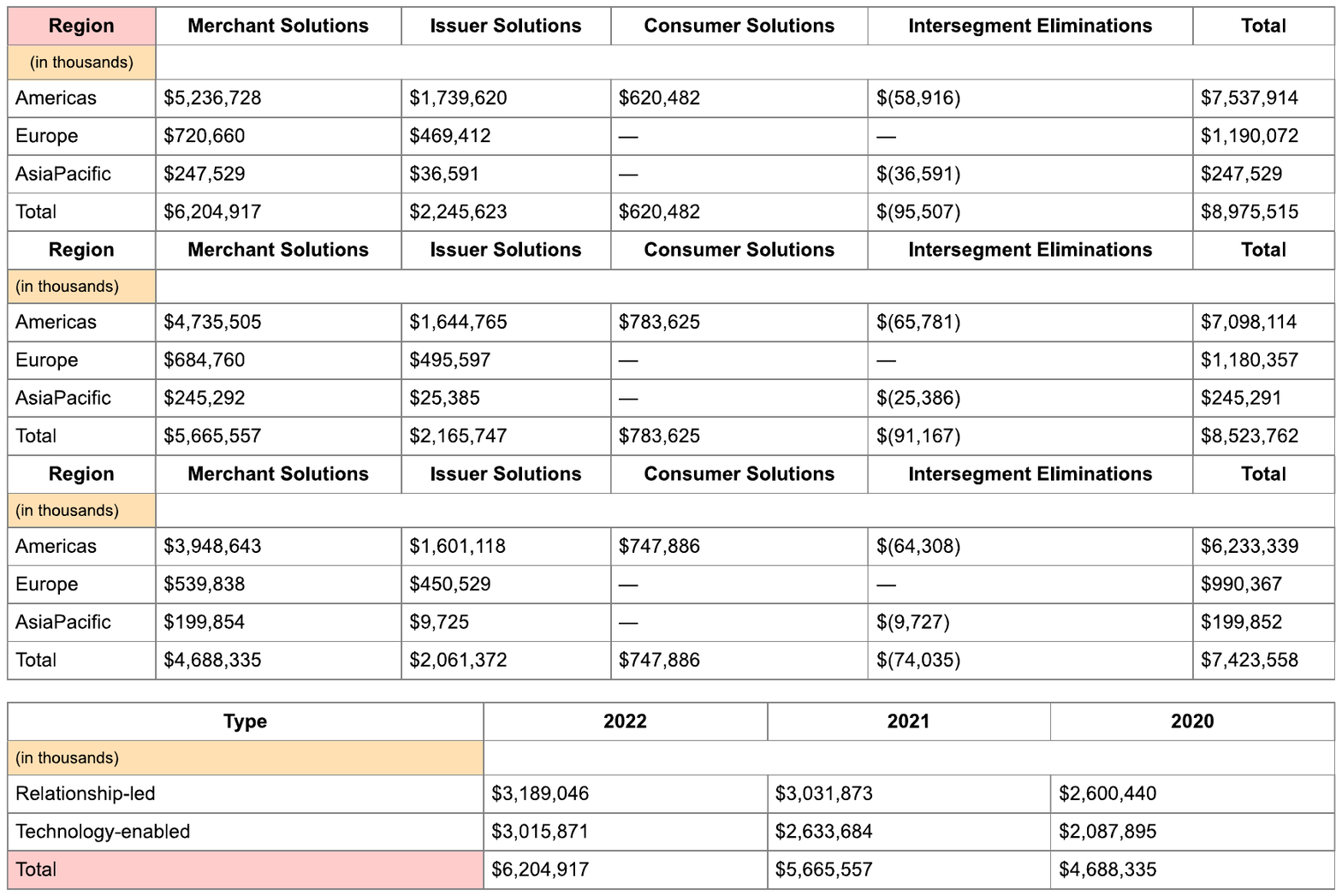}
    \caption{Table generated by \sysname. \sysname successfully preserved the core hierarchical structure with three distinct temporal sections and maintained geographic segmentation across most business segments. \sysname is also able to preserve numerical data across the majority of cells and successful separation of the distribution channel breakdown table as a distinct section. However, there are also some text-based reconstruction issues: (1) hallucinated text entries (highlighted in red) where \sysname introduced content not present in the original  document, and (2) misalignment issues (highlighted in light orange) where data values from the source were incorrectly positioned, resulting in values being associated with wrong column headers or row categories. }
\label{fig:example_3_pdf}
\end{figure}

 Listing \ref{example_listing_3} illustrates the unstructured, copy-pasted input provided to \sysname. The flattened textual representation lacks clear delineation of headers, rows, and hierarchical relationships, creating ambiguity for structural recovery.
\begin{lstlisting}[language=,caption={Unstructured input from multi-segment revenue breakdown table, used as input to \sysname}, label=example_listing_3]
YearEndedDecember31,2022\r\n MerchantSolutions\r\n Issuer\r\n Solutions ConsumerSolutions\r\n Intersegment\r\n Eliminations Total\r\n (inthousands)\r\n Americas $5,236,728 $1,739,620 $620,482 $(58,916) $7,537,914\r\n Europe 720,660 469,412 - - 1,190,072\r\n AsiaPacific 247,529 36,591 - (36,591) 247,529\r\n $6,204,917 $2,245,623 $620,482 $(95,507) $8,975,515\r\n YearEndedDecember31,2021\r\n MerchantSolutions\r\n Issuer\r\n Solutions ConsumerSolutions\r\n Intersegment\r\n Eliminations Total\r\n (inthousands)\r\n Americas $4,735,505 $1,644,765 $783,625 $(65,781) $7,098,114\r\n Europe 684,760 495,597 - - 1,180,357\r\n AsiaPacific 245,292 25,385 - (25,386) 245,291\r\n $5,665,557 $2,165,747 $783,625 $(91,167) $8,523,762\r\n YearEndedDecember31,2020\r\n MerchantSolutions\r\n Issuer\r\n Solutions ConsumerSolutions\r\n Intersegment\r\n Eliminations Total\r\n (inthousands)\r\n Americas $3,948,643 $1,601,118 $747,886 $(64,308) $6,233,339\r\n Europe 539,838 450,529 - - 990,367\r\n AsiaPacific 199,854 9,725 - (9,727) 199,852\r\n $4,688,335 $2,061,372 $747,886 $(74,035) $7,423,558\r\n
\end{lstlisting}
The resulting structured table generated by \sysname is presented in Figure \ref{fig:example_3_pdf}. Notably, \sysname successfully:
\begin{itemize}
\item Preserves the core temporal structure, clearly segmenting data by year.
\item Maintains geographic and business segment distinctions, correctly recovering the hierarchical table layout.
\item Accurately recovers and aligns numerical values in the majority of cases.
\item Appropriately isolates and separates the distribution channel breakdown, maintaining its distinction from the main revenue segment.
\end{itemize}

However, two primary reconstruction issues emerge:

\begin{enumerate}
\item \textbf{Hallucinated text entries (highlighted in red):} \sysname introduces textual content not present in the source input, reflecting challenges in fully resolving ambiguous textual labels.
\item \textbf{Misalignment of numerical values (highlighted in orange):} Certain numerical data entries are incorrectly positioned, resulting in mismatches with their respective column headers or row categories.
\end{enumerate}

\subsection{Example 3}
\begin{figure}
    \centering
    \includegraphics[width=\linewidth]{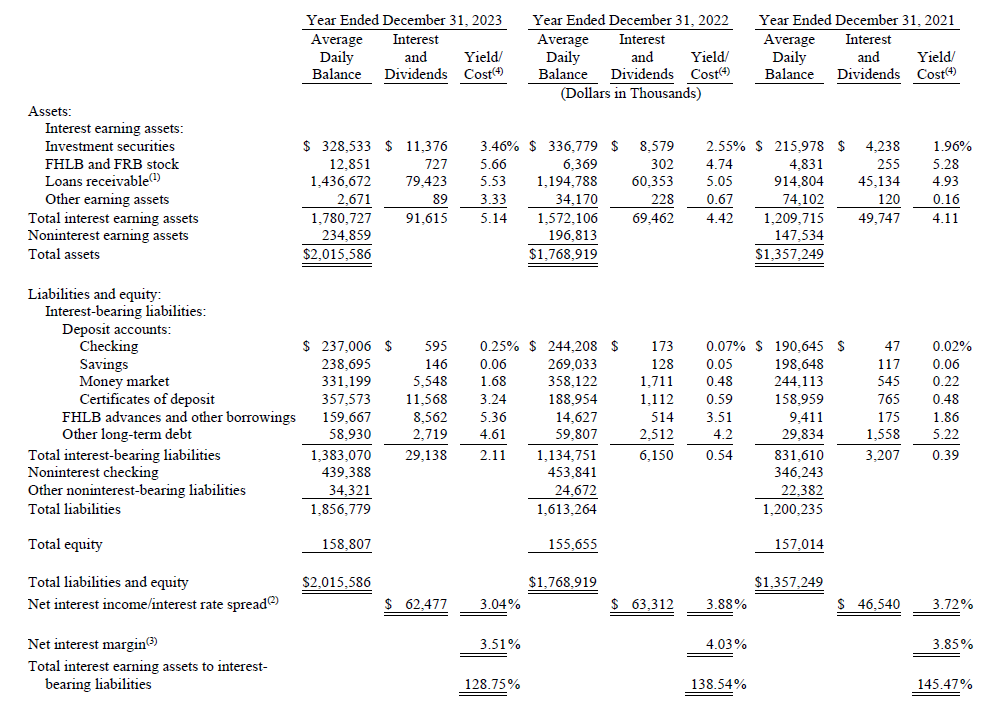}
    \caption{The table presents a multi-year financial performance report, including average balances, interest earned, and yields across assets and liabilities. The original table has multi-level column headers and repeated metrics across years.}
    \label{fig:example_2_png}
\end{figure}

This example illustrates the performance of \sysname on reconstructing a structurally intricate multi-year financial performance report extracted from a corporate financial statement. The original table (Figure \ref{fig:example_2_png}) captures detailed financial metrics, including average balances, interest earned, and yield rates across multiple asset and liability classes spanning three consecutive years (2021–2023). Notable complexities in this table include:

\begin{itemize}
\item Multi-level column headers, with repeated labels for each year (e.g.,``Average Balance'', ``Interest'', ``Yield'').
\item Nested row labels categorizing financial elements such as assets, liabilities, and equity.
\item Numerical values in varying formats, including percentages, currency, and absolute figures.
\end{itemize}

Listing \ref{example_listing_2} shows the raw, structurally ambiguous textual input provided to \sysname. This flattened textual representation features challenges such as inconsistent whitespace, merged and misaligned columns, and disrupted multi-line headers, representing typical copy-paste noise encountered in practice.

\begin{figure}
\centering
    \includegraphics[width=0.8\linewidth]{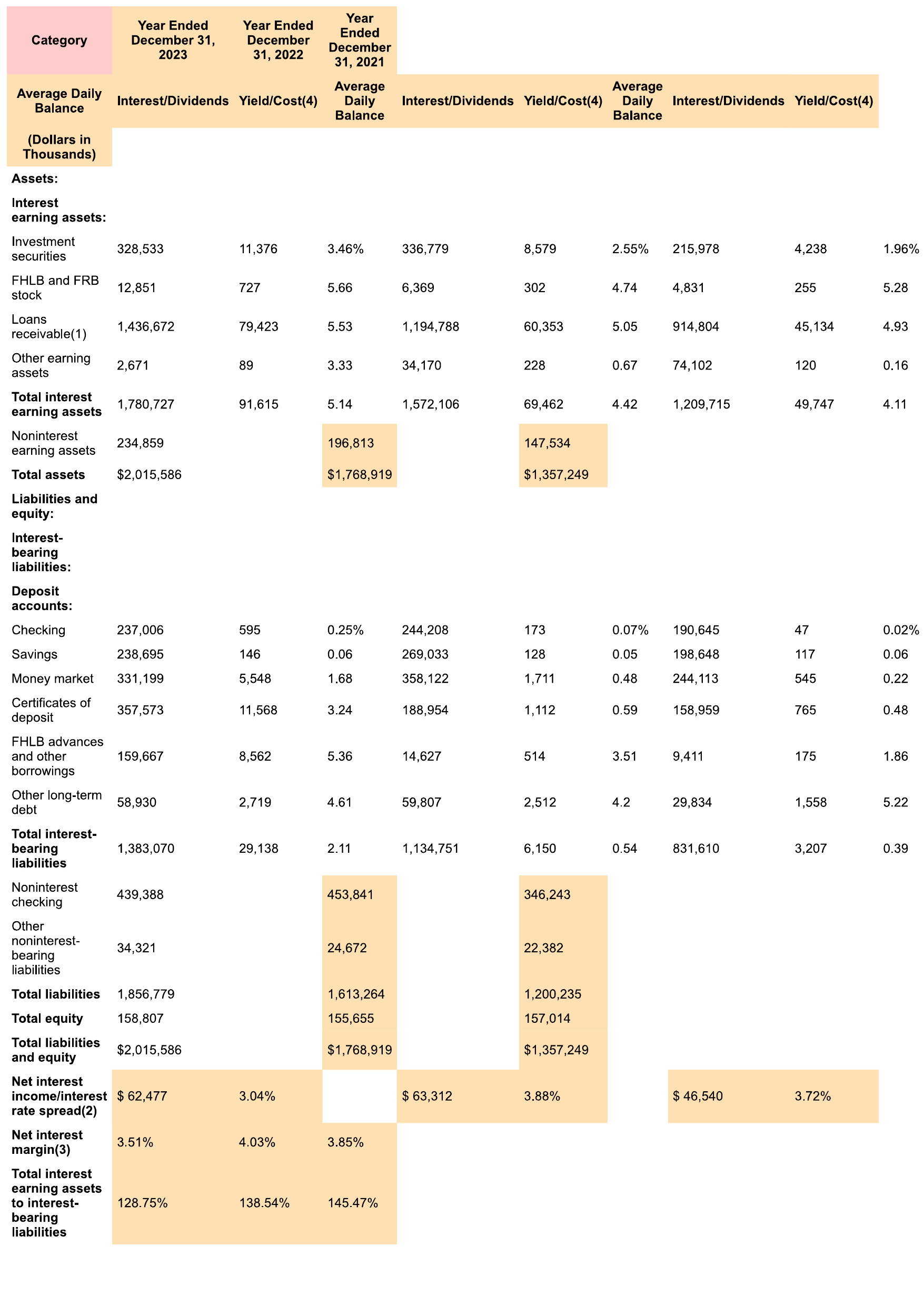}
    \caption{Table generated by \sysname. Despite the structural intricacy of the table, featuring nested header rows, label repetitions across columns, and varying numerical formats, \sysname is able to successfully reconstruct the overall layout and preserve the majority of header hierarchies and cell alignments with high fidelity. Cells shaded in light red represent \textit{hallucinated values}, which were not present in the ground truth but were incorrectly introduced and yellow cells indicate \textit{misaligned values}, where content was correctly extracted but inserted into an incorrect row or column, typically due to confusion between repeated sub-headers. These errors highlight current limitations of \sysname in aligning hierarchical headers and preserving structural context, especially when operating solely on text extracted via copy-paste. In such scenarios, visual layout cues such as merged cells, indentation, and spacing are lost, making it difficult for language models to infer correct associations between headers and values.}
    \label{fig:example_2_pdf}
\end{figure}

The structured table generated by \sysname (Figure \ref{fig:example_2_pdf}) demonstrates significant capability in:

\begin{itemize}
\item Accurately reconstructing the overall table layout, successfully segmenting yearly data and maintaining clear distinctions between asset and liability categories.
\item Preserving hierarchical header structures and correctly associating most numerical values with their corresponding financial dimensions.
\end{itemize}

However, this example also highlights specific reconstruction errors:

\begin{enumerate}
\item \textbf{Hallucinated numerical values (highlighted in red):} \sysname incorrectly introduces numeric data absent from the original source, indicating difficulties in interpreting ambiguous textual inputs.
\item \textbf{Misaligned numerical values (highlighted in yellow):} Although the data values themselves are correctly extracted from the source, they are occasionally misplaced into incorrect cells due to confusion between repeated sub-headers or loss of spatial alignment cues inherent in the original visual representation.
\end{enumerate}

These observations underscore existing limitations of \sysname when reconstructing structurally complex tables relying exclusively on noisy textual inputs without visual formatting cues.

\begin{lstlisting}[language=,caption={ Input text to \sysname: a structurally ambiguous table representation resulting from manual copy-pasting from a financial report. This unstructured text contains noise such as misaligned columns, merged headings, and inconsistent whitespace, and serves as the input for table explicitization.},label=example_listing_2]
Year Ended December 31, 2023 Year Ended December 31, 2022 Year Ended December 31, 2021   \r\n    Average Interest   Average Interest  Average Interest \r\n    Daily and Yield/ Daily and Yield/ Daily and Yield/   \r\n    Balance Dividends Cost(4) Balance Dividends Cost(4) Balance Dividends Cost(4)   \r\n    (Dollars in Thousands)   \r\nAssets:  \r\nInterest earning assets:  \r\nInvestment securities   $ 328,533 $ 11,376 3.46%   $ 336,779    $ 8,579 2.55%   $ 215,978    $ 4,238 1.96% \r\nFHLB and FRB stock 12,851  727 5.66 6,369 302 4.74 4,831 255 5.28  \r\nLoans receivable(1) 1,436,672  79,423 5.53 1,194,788 60,353 5.05 914,804 45,134 4.93  \r\nOther earning assets 2,671  89 3.33 34,170 228 0.67 74,102 120 0.16  \r\nTotal interest earning assets 1,780,727  91,615 5.14 1,572,106 69,462 4.42 1,209,715 49,747 4.11  \r\nNoninterest earning assets 234,859 196,813    147,534 \r\nTotal assets   $2,015,586    $1,768,919  $1,357,249 \r\n   \r\nLiabilities and equity:  \r\nInterest-bearing liabilities:  \r\nDeposit accounts:  \r\nChecking   $ 237,006 $ 595 0.25%   $ 244,208    $ 173 0.07%   $ 190,645    $ 47 0.02% \r\nSavings 238,695  146 0.06 269,033 128 0.05 198,648 117 0.06  \r\nMoney market 331,199  5,548 1.68 358,122 1,711 0.48 244,113 545 0.22  \r\nCertificates of deposit 357,573  11,568 3.24 188,954 1,112 0.59 158,959 765 0.48  \r\nFHLB advances and other borrowings 159,667  8,562 5.36 14,627 514 3.51 9,411 175 1.86  \r\nOther long-term debt 58,930  2,719 4.61 59,807 2,512 4.2 29,834 1,558 5.22  \r\nTotal interest-bearing liabilities 1,383,070  29,138 2.11 1,134,751 6,150 0.54 831,610 3,207 0.39  \r\nNoninterest checking 439,388 453,841    346,243 \r\nOther noninterest-bearing liabilities 34,321 24,672    22,382 \r\nTotal liabilities 1,856,779 1,613,264    1,200,235 \r\n   \r\nTotal equity 158,807 155,655    157,014 \r\n   \r\nTotal liabilities and equity   $2,015,586    $1,768,919  $1,357,249 \r\nNet interest income/interest rate spread(2) $ 62,477 3.04% $ 63,312 3.88% $ 46,540 3.72% \r\n   \r\nNet interest margin(3) 3.51%   4.03%    3.85% \r\nTotal interest earning assets to interest\r\nbearing liabilities 128.75%   138.54%    145.47%
\end{lstlisting}

